\documentclass{article} %
\usepackage{collas2022_conference,times}
\usepackage{amsfonts}       %
\usepackage{nicefrac}       %
\usepackage{microtype}      %
\usepackage{xcolor}         %

\usepackage{authblk}
\usepackage{longtable}
\usepackage{tabu}
\usepackage{listings}
\usepackage{amsthm}
\usepackage{amsmath}
\usepackage{dirtytalk}
\usepackage{scalerel}
\usepackage{color, colortbl}

\usepackage{subfiles}
\usepackage{colortbl}     
\usepackage{wrapfig,lipsum}
\usepackage{subcaption}

\usepackage{microtype}

\usepackage[utf8]{inputenc} %
\usepackage[T1]{fontenc}    %
\usepackage{hyperref}       %
\usepackage{url}            %
\usepackage{booktabs}       %
\usepackage{graphicx}

\newcommand\blfootnote[1]{%
  \begingroup
  \renewcommand\thefootnote{}\footnote{#1}%
  \addtocounter{footnote}{-1}%
  \endgroup
}
 
\definecolor{Gray}{gray}{0.95} 

\newcommand{\Message}[1]{\textcolor{black}{#1}}

\usepackage{amsmath,amsfonts,bm}

\def\eqref#1{equation~\ref{#1}}

\def\1{\bm{1}}

\DeclareMathAlphabet{\mathsfit}{\encodingdefault}{\sfdefault}{m}{sl}
\SetMathAlphabet{\mathsfit}{bold}{\encodingdefault}{\sfdefault}{bx}{n}

\usepackage{hyperref}
\hypersetup{   
    colorlinks=true,
    linkcolor=red, 
    filecolor=magenta,      
    urlcolor=blue,
    citecolor=purple,   
    pdftitle={Overleaf Example},
    pdfpagemode=FullScreen,
    }
  
\title{Continual Learning with Foundation Models: An Empirical Study of Latent Replay}
\author{%
    Oleksiy Ostapenko$^{123}$ Timothee Lesort$^{12}$ Pau Rodr\'iguez$^3$  Md Rifat Arefin$^{12}$ \linebreak Arthur Douillard$^{46}$ Irina Rish$^{127}$ Laurent Charlin$^{157}$
    \\ 
    \small$^1$Mila - Quebec AI Institute, $^2$Université de Montréal, $^3$ServiceNow, $^4$ Heuritech  $^5$HEC Montréal, \\ $^6$Sorbonne University, $^7$Canada CIFAR AI Chair  \\
}

\collasfinalcopy %

\begin{document}

\maketitle

\begin{abstract}
Rapid development of large-scale pre-training has resulted in foundation models that can act as effective feature extractors on a variety of downstream tasks and domains. %
Motivated by this, we study the efficacy of pre-trained vision models as a foundation for downstream continual learning (CL) scenarios.
Our goal is twofold. First, we want to understand the compute-accuracy trade-off between CL in the raw-data space and in the latent space of pre-trained encoders. Second, we investigate how the characteristics of the encoder, the pre-training algorithm and data, as well as of the resulting latent space affect CL performance.
For this, we compare the efficacy of various pre-trained models in large-scale benchmarking scenarios with a vanilla replay setting applied in the latent and in the raw-data space.
Notably, this study %
shows how transfer, forgetting, task similarity and learning are dependent on the input data characteristics and not necessarily on the CL algorithms. First, we show that under some circumstances reasonable CL performance can readily be achieved with a non-parametric classifier at negligible compute. We then show how models pre-trained on broader data result in better performance for various replay sizes. We explain this with representational similarity and transfer properties of these representations. Finally, we show the effectiveness of self-supervised (SSL) pre-training for downstream domains that are out-of-distribution as compared to the pre-training domain. We point out and validate  several research directions that can further increase the efficacy of latent CL including representation ensembling. %
The diverse set of datasets used in this study can serve as a compute-efficient playground for further CL research. Codebase is available under \url{https://github.com/oleksost/latent_CL}. \blfootnote{Correspondence to: oleksiy.ostapenko@t-online.de}
\end{abstract}

\section{Introduction}      
The goal of continual learning (CL) is to design machine learning (ML) algorithms that can learn  tasks presented in the form of a non-stationary stream. 
The relevance of CL for practical ML applications is usually justified with a reduced computational cost as compared to offline retraining on all data seen so far. While in a hypothetical infinite compute regime CL can be trivially solved through offline retraining, in practice buffers of small numbers of samples per task are used -- a strategy known as experience replay (ER).
This strategy is known to be very versatile and applicable in large set of scenarios~\citep{Kalifou19,diethe2018continual,lesort2019regularization,prabhu2020gdumb,Traore19DisCoRL,Belouadah2018DeeSIL,wu2019large,Hou_2019_CVPR,lesort2020continual,douillard2020podnet} in comparison with other families of CL methods~\citep{Lopez-Paz17,mendez2021lifelong}. 
In this work we largely focus on the setting of class incremental learning and ER as the strategy of choice.

Recent years have witnessed development of large scale models pre-trained on broad data -- the so called foundation models. These models are intended to be universal feature extractors that can produce useful features for a large number of downstream tasks %
~\citep{bommasani2021opportunities}.
With the increasing popularity of such models as a foundation for downstream tasks, several recent works have explored their continual fine-tuning~\citep{mehta2022an,wu2022pretrained,ramasesh2022effect}. At the same time, an emerging trend in computer vision (CV) and natural language processing (NLP) shows that such models can be used even without expensive fine-tuning of the feature encoder~\citep{devlin2018bert,brown2020language} or in zero-shot manner~\citep{sanh2021multitask,radford2021learning,zhai2022lit}. 

In this work, motivated by this trend, we explore how \textit{frozen} foundation models can be effective for CL. This strategy is justified by its increased compute efficiency, potential to mitigate data privacy concerns~\citep{desai2021continual}, better sample efficiency, as well as potential reduction of CL to training a non-parametric models that do not forget by design.

For this, we conduct an extensive analysis of CL using the ER strategy applied in the latent space of pretrained models (latent ER) using 26 different large-scale models pre-trained on a variety of datasets, architectures and with different pre-training algorithms. We use 5 different raw data streams with varying degree of relatedness to the pre-training domain. Encoding these streams with the pre-trained encoders resulted in a set of 130 encoded CL streams, each being effectively a distinct task sequence with distinct intrinsic properties such as task complexity, similarity, relatedness, etc.. This allows us to study CL as a function of data and not, as traditionally done, as a function of CL algorithm. 

Our main contributions can be summarized as following. (1) We contrast latent ER with the traditional strategy of fine-tuning the whole model and replaying raw data (end2end ER). The comparison is done in terms of the compute-accuracy trade-off and we find that, while being almost two orders of magnitude more computationally expensive, end2end ER leads to substantial accuracy improvement only on streams that are outside of the pre-training domain (OOD streams). At the same time, on all tested streams we have found an encoder that lead to final CL accuracy better or comparable to the best fine-tuned model. (2) We present a data driven empirical investigation of latent CL with various datasets and describe how characteristics, such as transfer and interference, usually associated to CL algorithms' performance, can be heavily dependent on the properties of the data. (3) The analysis and the conclusions that we draw from the experiments help us to better understand the link between the characteristics of the encoded data distribution, properties of the pre-training procedure and the behaviour of downstream CL algorithms trained on it. We further detail the importance of these contributions in Appendix ~\ref{ap:motivation_latent}.

\section{Related Works}
\textbf{Pretrained models.} Transfer learning is one of the most successful paradigms in deep learning~\citep{yosinski2014transferable}. Most computer vision systems start from ImageNet pre-trained weights~\citep{ren2015faster,chen2017deeplab,carion2020end}. A recent papers on natural language processing (NLP) indicate that SSL of large transformer models~\citep{devlin2018bert,reddy2021dall,douillard2021dytox} on vast amounts of pre-training data results in models that, without further training, generalize well to new tasks from few examples~\citep{brown2020language} and perform well in zero-shot learning tasks~\citep{sanh2021multitask}. In the field of computer vision, CNN and transformer-based models are starting to match the performance of supervised models on ImageNet~\citep{caron2021emerging,tomasev2022pushing}. In this work, we investigate whether such pre-trained representations can be leveraged to achieve CL from frozen features.

\textbf{Pre-training in CL} can be either considered (a) as a post-deployment learning of the first task of the CL stream (\emph{internal pre-training}) or (b) as a prior to deployment training on another dataset for which forgetting is not considered (\emph{external pre-training}). In (a), ~\citet{castro2018end} discarded the first task performance considered as ``pre-training.'' ~\citet{gallardo2021self}  remarked that the larger (\textit{w.r.t.} number of classes) the initial task was, the more class-agnostic the feature extractor became, further exacerbated with SSL pre-training, and consequently leading to less forgetting. In (b), pre-training in the classical sense on a large external dataset (\textit{e.g.} ImageNet \citep{deng2009imagenet}), except specific applications (\textit{e.g.} segmentation \citep{cermelli2020modeling}, meta CL \citep{caccia2021special,caccia2020online,javed2019meta}), is rarely used in CL. \citet{Hayes18MemoryEfficient,lesort2021continual,chrysakis2020online,banayeeanzade2021generative} used a ResNet pretrained on ImageNet and then transferred a sequence of tasks. The two formers kept the pretrained feature extractor frozen during the continual training. More recently, \cite{mehta2022an} remarked that pre-training, for both NLP and CV, done on a diverse dataset significantly reduced forgetting in downstream CL tasks. \citet{wu2022pretrained} investigated large NLP models (both encoders-decoders and decoders based) and concluded on the existence of robust early layers thanks to the SSL pre-training but remarked that middle and late layers still heavily suffered from forgetting. Finally, echoing \citet{gallardo2021self}, \citet{hu2022how} showed that SSL was also important for external pre-training.

\textbf{Latent replay} methods propose to store or generate activations from  intermediate layer(s) of a neural network, which are then replayed at the same intermediate layers to prevent forgetting when new tasks are learned during the CL process. While originally motivated through the lance of biological plausibility, this approach have shown some success. 
This method has also a lot of potential for various applications that we detail in appendix section \ref{ap:motivation_latent}.
Intuitively, latent replay methods assume that low level representations can be better shared across tasks and hence require less adaptation. To this end low layers are
either regularized~\citep{liu2020generative}, trained at a slower pace~\citep{pellegrini2020latent}, or, as.  in~\citep{van2020brain,hayes2020remind}, frozen. Freezing the low layers significantly reduces the computational cost of CL through reducing the amount of trainable parameters. These methods rely on the assumption that feature extractor can produce meaningful features and should be preferred in applications with limited compute. %

\textbf{Other existing CL approaches} can be roughly clustered into regularization based~\citep{kirkpatrick2017overcoming,Aljundi17}, dynamic architectures based methods~\citep{yoon2017lifelong,schwarz2018progress}, which include modular methods~\citep{mendez2021lifelong,veniat2021efficient,ostapenko2021continual}, and constrained optimization based methods~\citep{Lopez-Paz17,chaudhry2018efficient}.

\section{Methodology}
\label{sec:methods}          
We first introduce the different pre-training models and datasets and discuss the latent replay strategy. We then introduce the class and task similarity measures used as well as a distinction we make between in- and out-of-distribution streams.

\subsection{Setting and problem statement} 
\textbf{Encoders.} We consider a total of up to 32 encoders (most experiments used 26) with either  ResNet~\citep{He_2020_CVPR} or Vision Transformer (ViT)~\citep{dosovitskiy2020image} architectures. These are pre-trained using different objectives, and amount of data, in order to assess how these properties affect CL performance. See model details in App.~\ref{sec:models}. 

\textbf{Classifiers.} We use a simple multi-layer perceptron (MLP) and non-parametric models as classifiers. The MLP classifier is endowed with 1 hidden layer of 1024 units. The 2 non-parametric metric-based classifiers are the nearest mean classifier (NMC) and streaming linear discriminant Analysis (SLDA)~\citep{PangIncremental2005}, both do not require replay. NMC makes predictions by selecting the class corresponding to the nearest class prototype, where prototypes are the means of all samples in the class. SLDA represents each learned class using running means by class and features covariance matrix. It makes predictions by pondering distance with running means and covariance values (details in App.~\ref{ap:sec:metric_based_compute}).

\textbf{Datasets.} We consider six different datasets for   downstream tasks. CUB-200~\citep{WelinderEtal2010} which contains 6K images of birds distributed in 200 different categories. Cars196~\citep{KrauseStarkDengFei-Fei_3DRR2013}, which is formed 16,185 images of 196 classes of cars. CIFAR-100~\citep{krizhevsky2009learning} contains 60K images distributed in 100 categories. DTD~\citep{cimpoi14describing}, a texture dataset, consisting of 5640 images, organized in 47  categories. The Oxford Pet dataset~\citep{parkhi2012cats}, which is a 37 category pet image dataset with roughly 200 images per class.  FGCVCAircraft~\citep{maji13fine-grained}, which contains 10,200 images of aircrafts, with 100 images for each of the 102 classes.

\textbf{Dataset Encoding and Scenarios.} For each dataset, we encode all its examples using each of the pre-trained models. We do this once, as a pre-processing step before applying CL algorithms. For ResNet-based models, we extract the embeddings of the last layer, after global average pooling. For transformer-based models we extract the \texttt{[CLS]} or \texttt{[EOS]} token, as it is common procedure~\citep{dosovitskiy2020image,radford2021learning}. For models that use batchnorm~\citep{ioffe2015batch}, we use the pre-trained statistics and we do not update them during feature extraction. Models from  \texttt{timm}\footnote{\url{https://github.com/rwightman/pytorch-image-models}} include their own data-preprocessing functions. For other models, such as dino~\citep{caron2021emerging}, we resize images to 224$\times$224 and standardize them with ImageNet statistics (except for the CIFAR100 dataset, which is resized to 100x100). 

The encoding pipeline is provided by \textit{continuum}~\citep{douillard2021continuum}, including the formation of CL streams. 
We used those datasets to create class-incremental scenarios, i.e. a new task brings new classes \citep{van2019three,lesort2021understanding}. %
For most experiments, we evaluate models on a single dataset, which  results from splitting a dataset into five different class groups (tasks) and learning each task in a sequence.  We also consider a CL \textit{multi-dataset stream}, which introduces each of the six different datasets in a sequential manner, resulting in six different tasks.

\textbf{Metrics and variables of interest.} In Tab.~\ref{tab:variables}, we report the various metrics and their description. We also explore some variables of interest to better understand downstream performance: \textit{pre-training data}, \textit{models architecture}, \textit{number of parameters}, \textit{pre-training FLOPS}, and \textit{latent dimension size}.

\subsection{Approach: Latent Replay}
\label{sec:latent_replay}
To fairly compare models with different replay buffer sizes we use a replay buffer sampling strategy that ensures that each class (new and replayed) is sampled with the same probability during learning, guaranteeing a learning setting close to iid at each task. This amounts to oversampling/under-sampling the replay buffer samples if the number of samples per-class in the new task is larger/smaller than the per-class replay buffer. %
Even if not very sample efficient, this strategy avoids imbalanced datasets problems and enables a normalized comparison. It is important to note that in this replay strategy the more past classes were learned, the higher the per-epoch computation cost of the current task will be.

\begin{table*}[t]         
\caption{A summary of CL metrics of interest.}
\label{tab:variables}
\resizebox{\linewidth}{!}{
\begin{tabular}{@{}lp{5cm}p{12cm}@{}}
\toprule   
\textbf{Name} & \textbf{Description} & \textbf{Comments} \\ \midrule 

\rowcolor{Gray}
$A_{\scaleto{CL}{3pt}}^{\scaleto{MLP}{4pt}}$ & \small CL Accuracy & {\small Accuracy on all tasks seen so far. Unless stated otherwise, measured at the end of the stream.} \\ %

$A_{\scaleto{CL-reinit}{4pt}}^{\scaleto{MLP}{4pt}}$ & \small $A_{CL}^{MLP}$ with re-initialization & {\small Accuracy on tasks seen so far when weights are reinitialized at the beginning of each new task.} \\

\rowcolor{Gray}
$A_{\scaleto{CL}{3pt}}^{\scaleto{NMC}{4pt}}$ & \small Nearest Mean Classifier Accuracy & {\small This accuracy is computed by a nearest mean classifier with Euclidean distance. }\\% (also called nearest mean centroid classifier). It uses only the mean of the training data for classification. }\\

$A_{\scaleto{CL}{3pt}}^{\scaleto{SLDA}{4pt}}$ & \small Streaming LDA Accuracy & {\small CL accuracy of the SLDA classifier \citep{PangIncremental2005}.} \\% It takes account of both mean and variance of training data for classification.} \\

\rowcolor{Gray}
$A_{i,t}$ & \small Single task accuracy (acc. task) & {\small Accuracy of task $i$ measured at time $t$.} \\ %

$A_{task-cl}$ & \small Mean task accuracy (acc. task cl) & {\small Average of single task accs.: $A_{task-cl} = \frac{1}{T} \sum_{i=0}^{T-1} A_{i,i}$. Measured for different ER buffer sizes.} \\

\rowcolor{Gray}
$A_{task-FS}$ & \small Mean task Accuracy Few-Shot (FSH) & {\small Like $A_{task-cl}$, but trained with 2 data points per class, no ER buffer and re-initialization.} \\
 
$A_{iid}$ & \small IID Accuracy (upper bound) & {\small Accuracy realized on a dataset in an IID setting (no CL).} \\
  
\rowcolor{Gray}     
$A_{task-iid}$ & \small Mean task acc. IID (acc task iid) & \small Similar to $A_{task-cl}$, but without replay, i.e. avv. acc. when each task is learned separately. \\ %

\textit{\small Interference total} & $A_{task-iid} - A_{iid}$ & {\small Estimates the adverse effect of learning tasks together versus one.} \\

\rowcolor{Gray}     
\textit{\small Interference} & \small $A_{task-iid} - A_{task-cl}$ & {\small Estimates the adverse effect of learning tasks with a given ER buffer versus one by one.} \\
  
\textit{\small Transfer} & \small $A_{CL}^{MLP} - A_{cl-reinit}^{MLP}$ & \small Estimates the amount of knowledge transferred through time/tasks in the weights. \\

\rowcolor{Gray}  
\textit{\small $F_{CL}^{MLP}$} & \small Forgetting / Relative Forgetting & {\small Forgetting is $A_{task-cl} - A_{cl}^{mlp}$. Relative forgetting is $F_{R_{T-1}} = \frac{1}{T} \sum_{t=0}^{T-1}  \frac{A_{task_{t:t}}-A_{task_{t:T-1}}}{A_{task_{t:t}}}$.} \\ \bottomrule
\end{tabular}}
\end{table*}
\subsection{Task and Classes Similarity}
\label{sub:intro_task_sim}
Several recent works have investigated the role of task similarity for catastrophic forgetting \citep{nguyen2019toward,ramasesh2021anatomy,lee2021continual}.
In this work we use task similarity measured in the latent space of large pretrained models as a tool to better understand the CL performance of latent ER models. Moreover, it provides a supplementary justification to the ID / OOD distinction that we introduce in the next section.
We use two different similarity measures. (1) We extend the idea of subspace overlap, recently introduced by~\citet{ramasesh2021anatomy} to longer task sequences (details in appendix \ref{ap:sec:similarity}). Importantly, similarity measures can not be used for algorithmic decision making at CL time when they assume access to future tasks' data (as is the case for subspace overlap). (2) We measure inter-class similarity, which can be computed prior to learning a new task, hence this is an actionable measurement that can be used by a CL algorithm. We detail how both  similarity metrics are calculated in Appendix~\ref{ap:sec:similarity}.

\subsection{In-Distribution (ID) vs Out-of-Distribution (OOD)}
\label{sub:in_out}                                                                          
We study the CL in two regimes. Each regime is characterized by the relationship between the source domain, i.e. pre-training data, and the target downstream tasks' domain: (1) the in-distribution (ID) and (2) out-of-distribution (OOD) regimes. %
In our setting, the ID datasets are CIFAR100 and CUB200. ID dataset contain classes from the pre-training data (e.g. CIFAR100 and 52 classes of birds are in the ImageNet~\cite{wen2022can}), while OOD dataset have different classes.  %
The datasets that we consider OOD are FGCVAicraft and Cars196. These datasets contain samples of cars and airplanes, respectively. Notably, while the super-classes ``car'' and ``airplane'' are present in the pre-training datasets such as ImageNet1K and ImageNet21k, the level of supervision provided by these pre-training datasets (e.g. cars vs. everything else) is not sufficient to result in the ability to distinguish fine-grained  classes of cars (e.g. ``Acura TL Sedan 2012'' vs. ``Acura TSX Sedan 2012'') or airplanes (we confirm this in Section~\ref{sec:experiments}).  %
Note, that we do not know exactly what data CLIP models are trained with. Hence, we do not include CLIP models in the ID versus OOD analysis. %
The exceptionally high performance of CLIP models on Cars196 dataset (see Fig.~\ref{ap:fig:mlp_slda_nmc_per_stream_per_model}) and low inter-class similarity (see Fig.~\ref{ap:fig:prototype_similarity}) indicate that classes from this dataset are likely to be part of the CLIP's data.
    
\begin{figure}[!tbp]
\begin{minipage}[b]{0.49\textwidth}

    \begin{subfigure}{0.49\linewidth}
    \centering
    \includegraphics[width=1\linewidth]{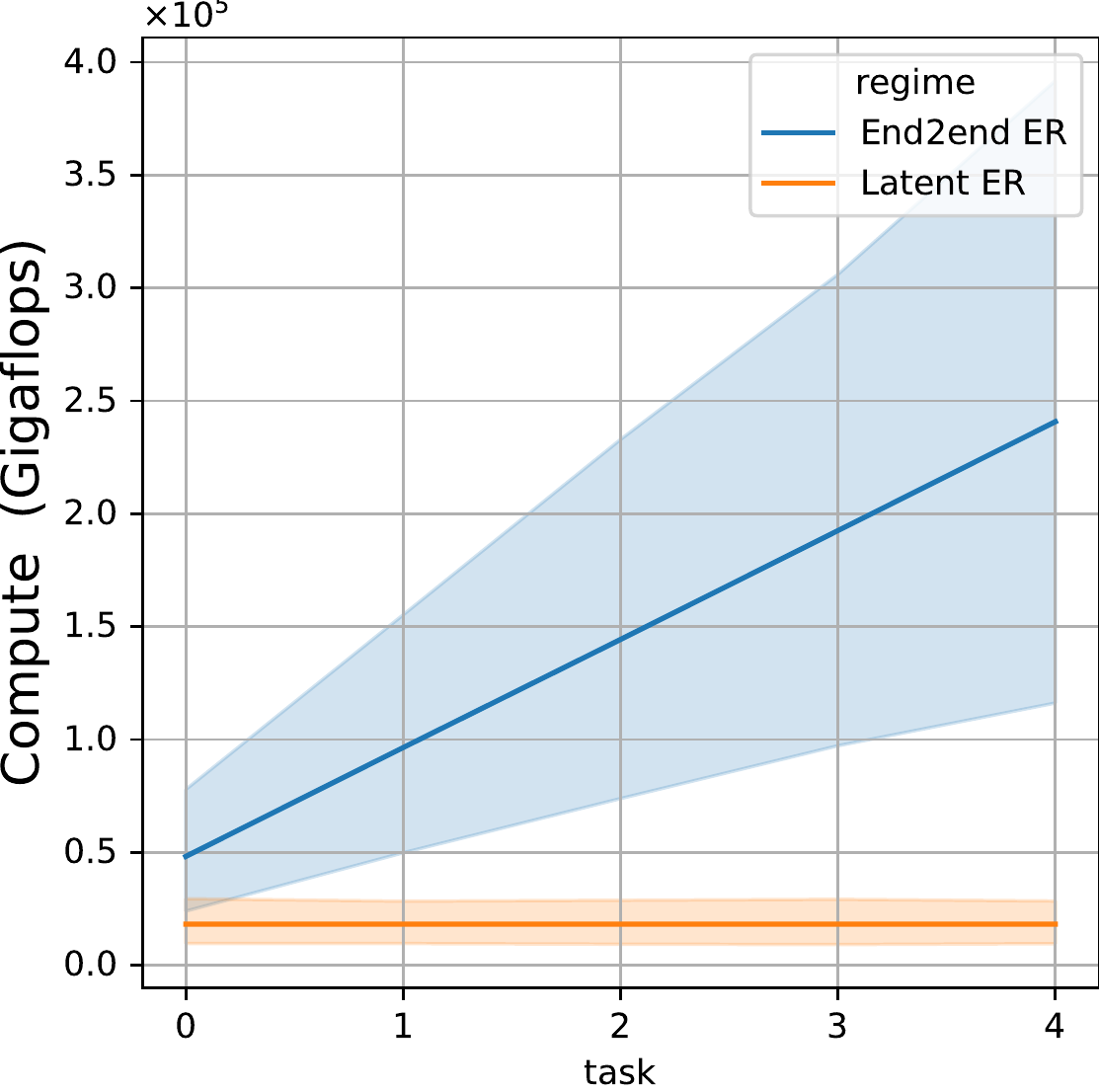}
    \caption{Stream level growth}
    \end{subfigure} 
    \begin{subfigure}{0.49\linewidth}
        \centering
        \includegraphics[width=1\linewidth]{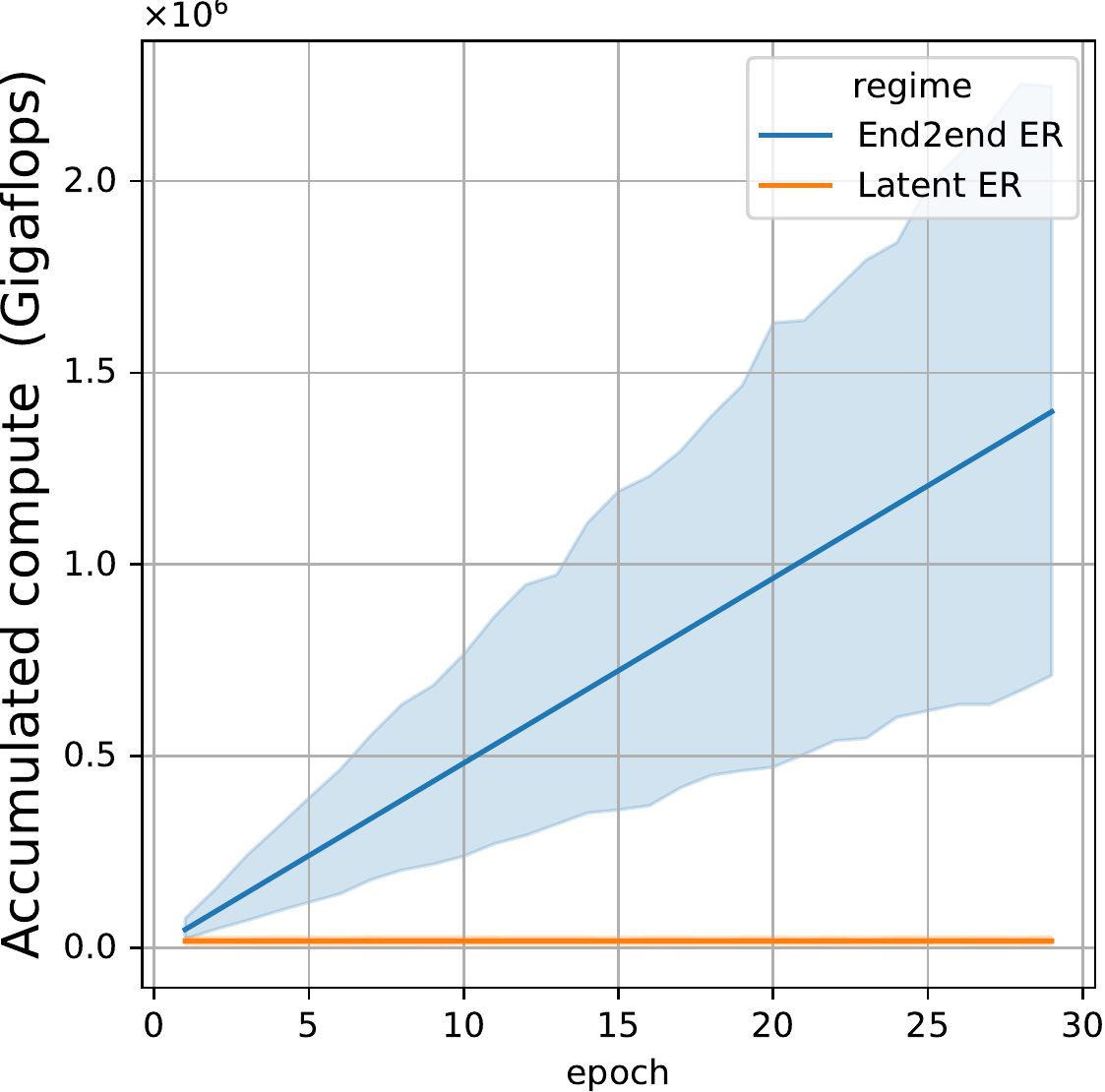}
        \caption{Task level growth}
    \end{subfigure}
    \caption{  \textbf{Compute growth estimation} 
    (a) compute per epoch for each new task (not accumulated) throughout the tasks of CIFAR100/5 for latent ER (including the encoding compute) and end2end ER.
    (b) Compute growth within the first task of CIFAR100/5 for latent ER and end2end ER regimes (accumulated). In both plots we average over encoders and task orderings (5). In the compute estimation we approximate the cost of a backward pass as twice the cost of the forward pass. Latent ER compute growth (orange line) is dominated by end2end ER's compute to an extend that it does not appear to grow at all.} 
    \label{fig:computation}
    \end{minipage}
    \hfill
    \begin{minipage}[b]{0.49\textwidth}
            \includegraphics[width=1.\textwidth]{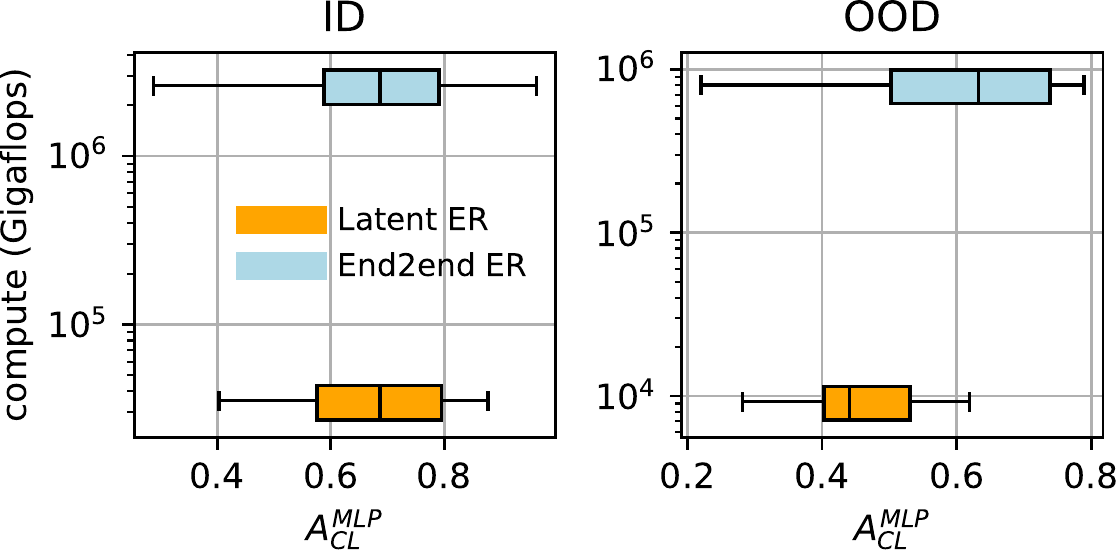}
    \caption{ 
    \textbf{Compute-accuracy trade-off} of end2end ER vs. latent ER  strategies. We perform end2end fine-tuning with end2end ER for 7 different feature encoders on 4    streams with 3 different ER buffer sizes (10,30,50) and with a fixed task order. Here we compare resulting accuracy (x-axis) and average compute (per category, y-axis) on both ID (left) and OOD (right) streams. A detailed per model view is in Fig.~\ref{ap:fig:compute-acc}. \Message{On average, end2end ER strategy results in a substantial performance improvement on OOD streams at the expense of almost 2 orders of magnitude more compute. As expected, no improvement is observed for ID streams.}}
    \label{fig:compute-acc}
    \end{minipage}
\end{figure}

An interesting insight from in-distribution (ID) and out-of-distribution (OOD) is in the Appendix, Fig.~\ref{ap:fig:prototype_similarity}. We observe that ID classes have lower inter-class similarity than OOD classes. We hypothesize that this is due to the fact that OOD datasets are fine-grained and the pre-trained encoders are mapping all samples around the same class prototype.

\section{Experimental Results}
\label{sec:experiments}
In this section we discuss our main findings. We first compare latent experience replay (ER) and end2end ER by reporting how they trade-off computational cost and accuracy. We then analyse latent ER strategies in terms of the relationship between pre-training and  downstream-task data. %
We finalize with an encoder-level analysis.

\subsection{Computational cost comparison: latent ER vs. End2end ER}
\label{sec:computation}
\paragraph{Latent ER scales better in terms of compute.}
The most obvious advantage of latent ER is the reduced compute and memory cost as compared to the conventional end2end ER. In Fig.~\ref{fig:computation} we compare the computational cost of latent ER with the end2end ER strategies. More specifically, we look at the growth pattern of the computational cost at a stream level (as new tasks are introduced) and at a task level (over epochs). In both cases computational cost of the end2end ER grows at a much larger rate making latent ER compute appear constant.
Importantly, latent ER only needs to perform one forward path per sample through the (large) encoder in order to produce the latent representation of tasks, i.e. encode the data. This process constitutes a much larger portion of the overall computation cost than the subsequent CL phase on the encoded data (see Fig.~\ref{ap:fig:compute_encoding_vs_CL} for comparison). Once encoded, the model can perform several epochs on the same encoded data.
On the other hand, the end2end ER strategy has to complete a forward and a backward pass through the entire model for each sample. This leads to a much larger per epoch computational cost of end2end ER  strategy as compared to latent ER, resulting in a steep increase of accumulated compute over the course of learning of a task as shown in Fig.~\ref{fig:computation}b. As for the stream-level growth pattern shown in Fig.~\ref{fig:computation}a, when encountered with a new task, the learner has to learn from the new examples and the examples from the replay buffer. This results in a increasing per epoch cost as new tasks are introduced. As discussed in Section~\ref{sec:latent_replay}, we use the oversampling and under-sampling strategies to avoid the class imbalance problems cause by new tasks having much mode samples per class than tasks in the replay buffer. It is important to mention that more elaborate replay strategies \citep{aljundi2019online,aljundi2019gradient,bagus2021investigation} can be used. These are likely to have equal effect on the scaling properties for both latent and end2end ER\footnote{A method that improves the efficacy of e.g. sample selection in end2end regime can likely be equally effective in latent ER.}. While exploring such strategies is out of this study's scope, we believe that the extent to which latent ER is more compute efficient than end2end ER should be similar for other ER strategies.

\paragraph{Compute-accuracy trade-off.}  
The computational advantage of latent ER can be only realized if the resulting accuracy is comparable to the accuracy of end2end ER. Here we investigate this compute-accuracy trade-off. For this we fine-tuned 7 encoders (both ViT and ResNet architectures, supervised and self-supervised pre-training, details in Appendix~\ref{ap:compute_acc_trade}) on 4 streams. As we show in App. Fig.~\ref{ap:fig:compute-acc}, for each stream we find an encoder for latent ER that results in a better or comparable performance to the performance of the best end2end ER model tested. Thereby latent ER has a substantially smaller computational cost. In Figure~\ref{fig:compute-acc} we give a consolidated view of the compute-accuracy trade-off for both ID and OOD streams. In the ID streams end2end ER even leads to a slight decrease in the CL performance on average as compared to latent ER. We observe that end2end ER results in a considerable performance improvement on the OOD streams on average, at the expense of almost two orders of magnitude more compute. %

\begin{figure}[!tbp]
\begin{minipage}[b]{0.49\textwidth}
    \centering
    \begin{subfigure}{0.49\linewidth}
    \includegraphics[width=0.9\linewidth]{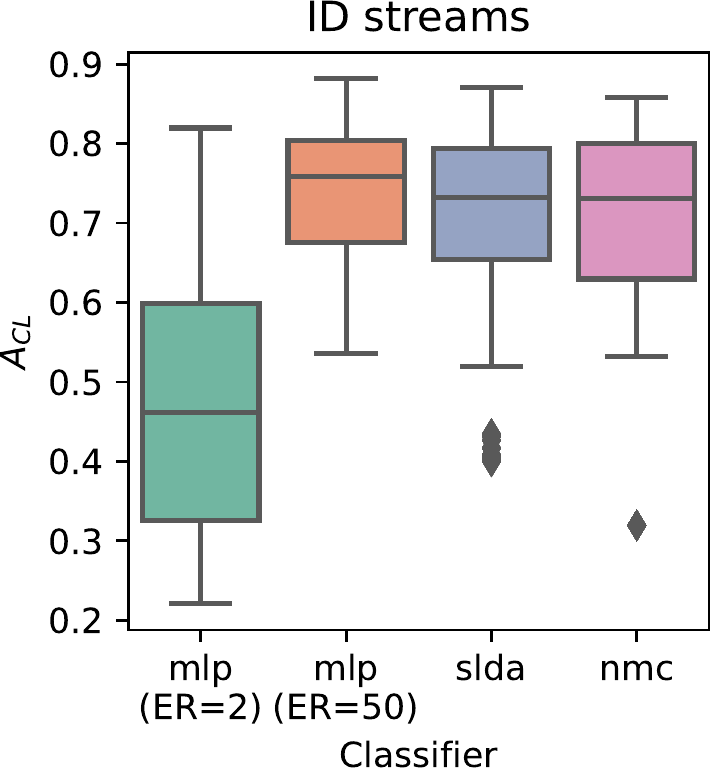}
    \caption{ID streams}
    \end{subfigure}
    \begin{subfigure}{0.49\linewidth}
    \includegraphics[width=0.9\linewidth]{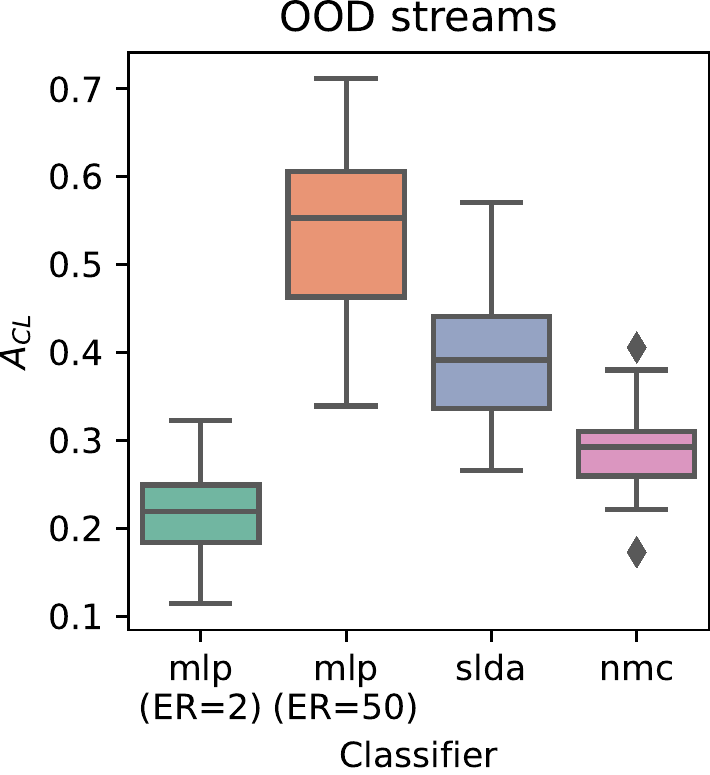}
    \caption{OOD streams}
    \end{subfigure}
    \caption{\textbf{ ID vs. OOD} comparison of the metric based classifiers NMC and SLDA, and replay based MLP in high and low ER buffer size regimes. \Message{For ID streams the CL accuracy of metric based classifiers is close to the accuracy of replay based MLP in the large ER buffer regime.}}
    \label{fig:regime_metric}
\end{minipage}
\hfill 
\begin{minipage}[b]{0.49\textwidth}

    \begin{subfigure}[b]{0.45\linewidth} 
    \includegraphics[width=1\linewidth]{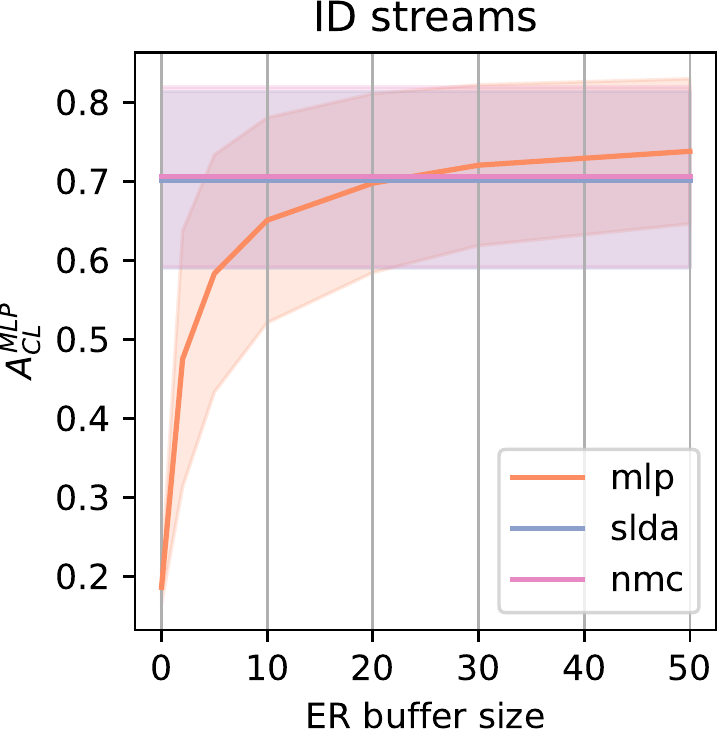}
    \caption{ID streams}
    \end{subfigure}
    \begin{subfigure}[b]{0.45\linewidth} 
    \includegraphics[width=1\linewidth]{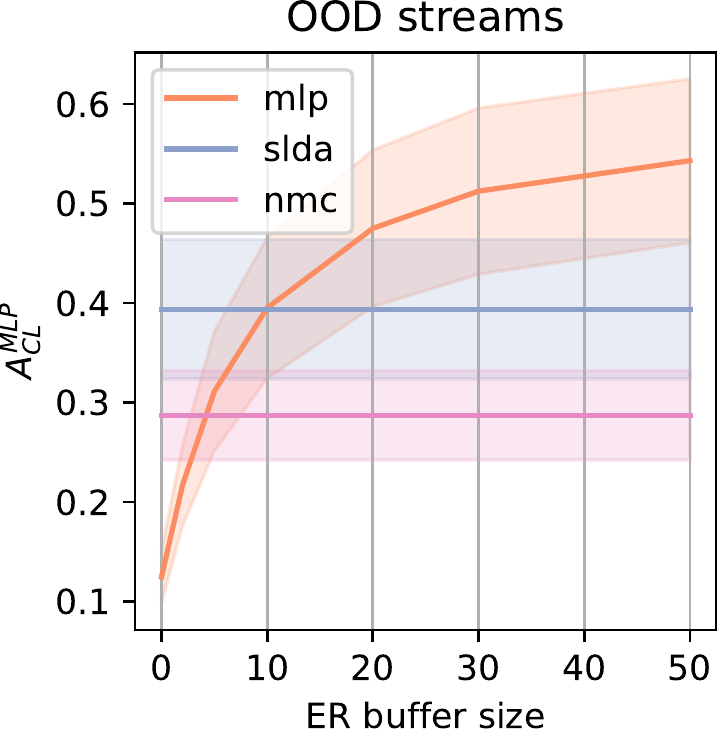}
    \caption{OOD streams}
    \end{subfigure}
    \caption{\textbf{ Comparison of metric based accuracy (NMC/SLDA) with the latent ER based solution.} \Message{ This Fig. enables the estimation of the effective replay buffer size of metric-based methods in ID and OOD datasets. Effective buffer size corresponds to the point at which latent ER methods reach the accuracy of SLDA and NMC. It also shows that increasing the buffer size saturates faster the accuracy in ID datasets than on OOD datasets.}}
    \label{fig:data_efficiency}
\end{minipage}
\end{figure}

\paragraph{Metric based classifiers are compute-efficient and accurate on the ID streams.}
In Fig.~\ref{fig:regime_metric} we compare the performance of the metric based NMC and SLDA classifiers and latent ER based  classifiers on both ID and OOD streams. We observe that on the ID streams metric based classifiers achieve superior performance than latent ER ones with small ER size (2 samples per class) and are close to the performance of MLP with large ER buffer (50 samples per class). We can quantify the efficacy of metric based classifiers in terms of effective replay buffer size. As visualized in Fig.~\ref{fig:data_efficiency}, effective replay size is the buffer size required by latent ER based methods to reach the accuracy of SLDA and NMC. For latent ER based MLP performance gains beyond this point seem to be very marginal on the ID streams. This highlights the efficacy of the metric based classifiers on the ID streams. This can be attributed to the phenomenon discussed by~\cite{papyan2020prevalence} and~\cite{galanti2021role}, who point out that supervised pre-training can lead to features collapsing to the mean feature vector of the same classes facilitating the applicability of linear and metric based classifiers.
\begin{wrapfigure}[23]{r}{0.4\textwidth}   
    \centering                              
    \includegraphics[width=0.82\linewidth]{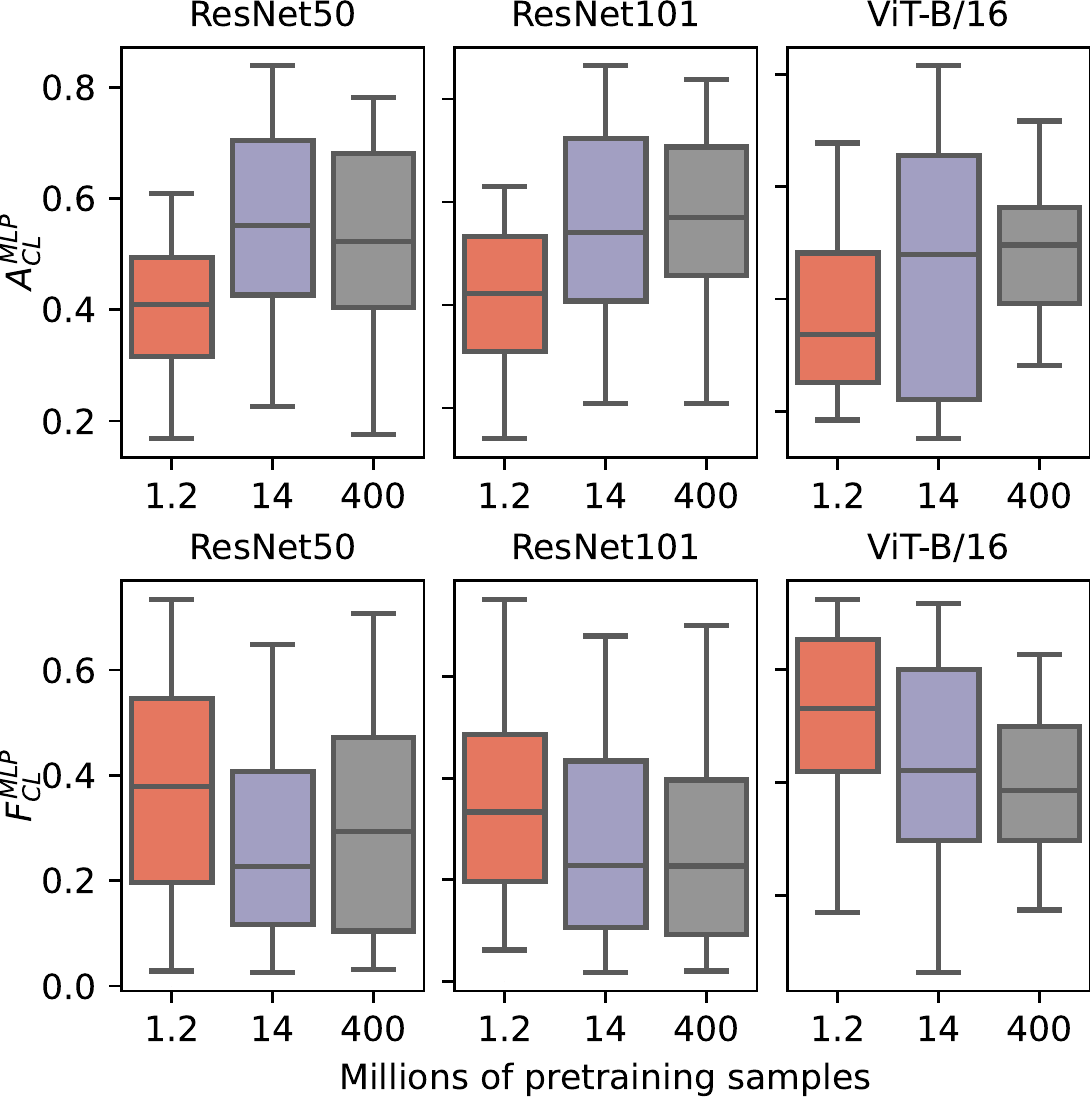}       
    \caption{\textbf{Accuracy} ($\uparrow$) and \textbf{forgetting} ($\downarrow$) for three different architecture families (resnet50, resnet101 and ViT)    pretrained on different datasets (6 replay buffer sizes, 4 streams and 5 task orders). Here, ResNetV2~\citep{szegedy2017inception} architecture is used for ImageNet21K (14M) dataset, which is slightly different from the standard ResNet architecture. \Message{More pre-training data increases CL performance.}}
    \label{fig:architectures}
    \end{wrapfigure}
All in all, it can be concluded that given a pre-trained encoder that produces ``good'' representations, CL can be tackled with a non-parametric classifier. Such models are very cheap computationally (see discussion in Appendix~\ref{ap:sec:metric_based_compute}) as they only require single path through the data and no replay. Currently available foundation models provide good representations for some data streams (ID streams) resulting in performance comparable to the more powerful MLP. Additionally, the efficacy of such strategy was demonstrated recently by \cite{banayeeanzade2021generative}, who first meta-pretrained an encoder and subsequently used a non-parametric classifier to continually learn unseen classes sampled from the same underlying data distribution as the pre-training tasks. While meta-pretraining can be prohibitively expensive on large-scale~\citep{ji2020convergence}, we believe that development of more universal feature encoders, i.e. foundation models pre-trained on broader data distributions, as well as domain-specific pre-training in cases where CL domain is know a priori, will grow the chances of downstream data to be in-distribution easing the CL application. %

\subsection{Data analysis: pre-training and downstream}
\label{sub:data_analysis}
\paragraph{Broader pre-training improves latent CL.}               
Intuitively, pre-training on a broader and more diverse set of examples should improve the universality of the representations produced by a feature extractor~\citep{kaplan2020scaling,yuan2021florence}. We confirm this intuition in Fig.~\ref{fig:architectures}, where we depict forgetting and CL accuracy of different architectural families after pre-training on datasets of different sizes. As discussed by ~\citet{ridnik2021imagenet} and ~\citet{radford2021learning} for the datasets used in this study the number of samples can be used as an approximation for the diversity of the pre-training data. As shown in Fig.~\ref{fig:architectures}, pre-training on the CLIP dataset (400M examples) tends to produce better results on both large architectural families (ResNet101 and ViT). Interestingly, CLIP models \footnote{Models trained on CLIP data (400M samples) with CLIP algorithm} are slightly outperformed by ImageNet21K (14M) pre-trained models in case of ResNet50 architecture. We hypothesize that this is due to limited capacity of the ResNet50 models that is unable to accommodate the diversity of 400M samples. 
  
\begin{figure}[!tbp]
\begin{minipage}[b]{0.42\textwidth}
\centering
\includegraphics[width=0.59\linewidth]{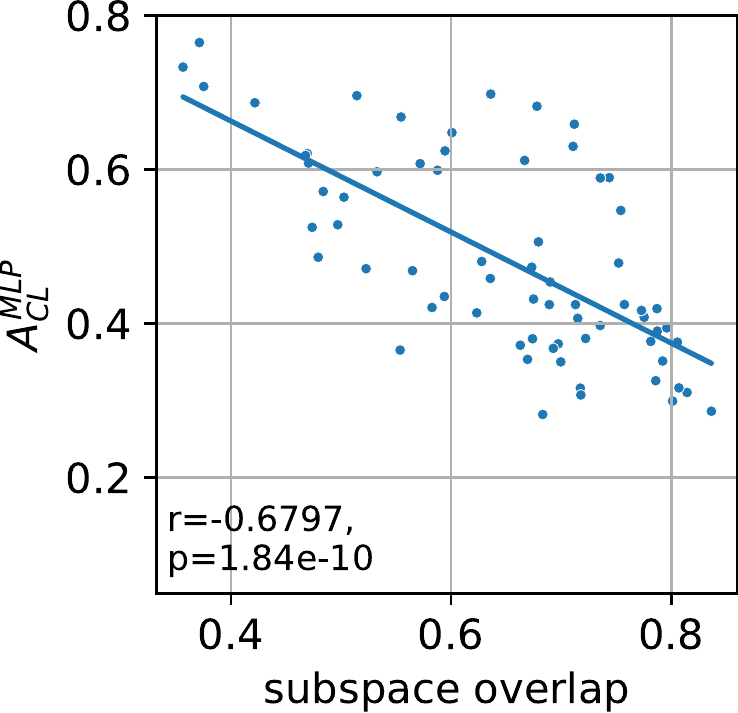}   
\caption{\textbf{Average Subspace Overlap} vs Acc. CL, each point represents a pre-trained model (per stream view in Fig.~\ref{ap:fig:subspace_sim_vs_CL_acc}). We also print the Pearson correlation $r$ and $p$-values for completeness. \Message{Lower overlap leads to better CL final accuracy.}}
\label{fig:subspace_sim_vs_CL_F}
\end{minipage}
\hfill
\begin{minipage}[b]{0.55\textwidth}                 
    \includegraphics[width=1\linewidth]{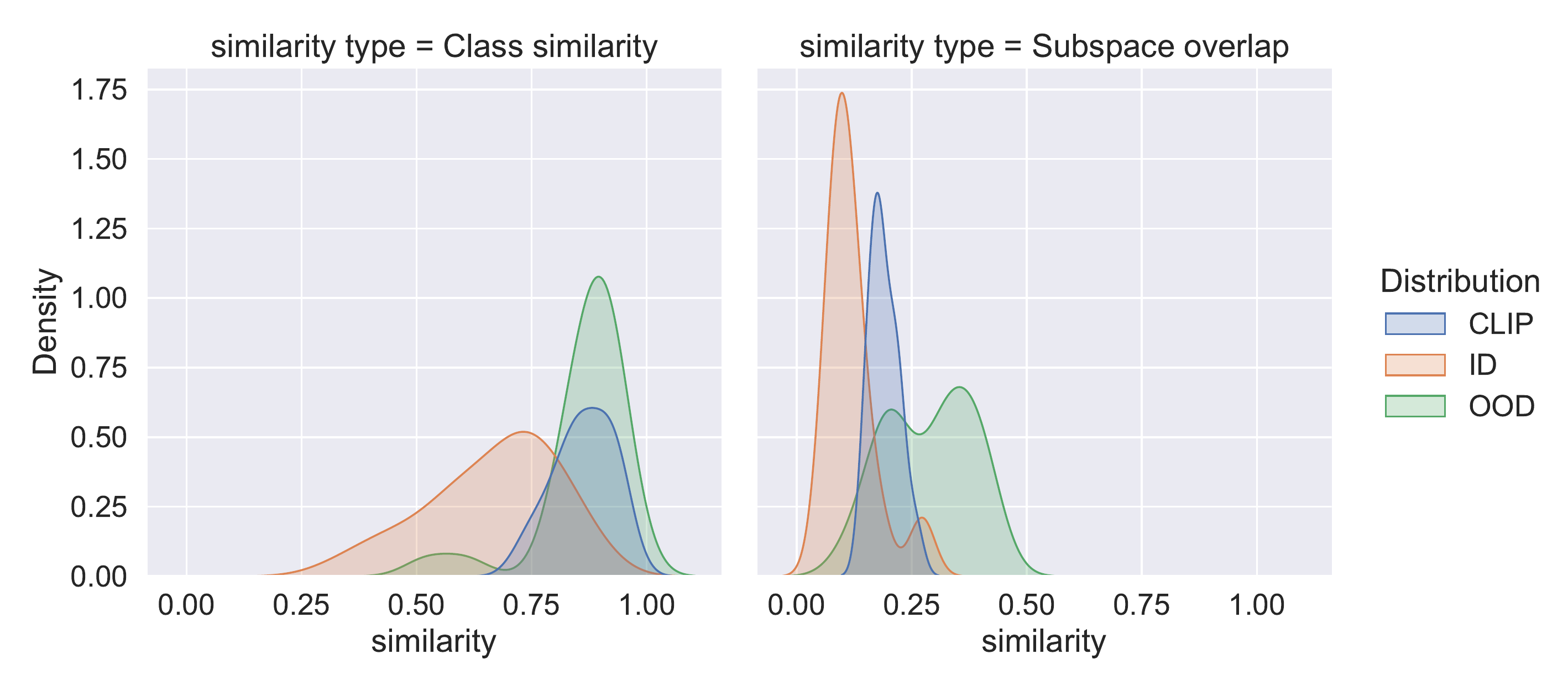}
    \caption{ \textbf{Average similarity density for all our models}. (left) with prototypes class cosine similarity. (right) with subspace overlap similarity. We plot separately CLIP's class similarities since they are not used in the ID vs OOD axis of study. \Message{OOD tasks exhibit larger class cos-similarity  and more subspace-overlap.}}
    \label{fig:sim_density}
\end{minipage}
\end{figure}

\paragraph{Similar tasks are harder to learn continually.}
In Fig.~\ref{fig:sim_density}, we show that class representations in the ID data streams tend to be less similar to each other %
than class representations from the fine-grained OOD streams (for both similarity metrics).  
In the cosine similarity measurement, lower similarity is equivalent to more orthogonal class-representations.
First, this indicates that ID streams are easier to solve since orthogonal representations are easier to discriminate. This is reflected in the overall higher classification accuracy achieved on the ID streams as shown in Fig.~\ref{fig:regime_metric}.  Second, as discussed by~\citet{ramasesh2020anatomy}, orthogonal representations lead to less forgetting due to less interference. We demonstrate this in Fig.~\ref{fig:subspace_sim_vs_CL_F}, where we plot the subset overlap similarity against $A_{MLP}^{CL}$ (see Sec.~\ref{sub:intro_task_sim} for details). We can observe a clear negative correlation between CL accuracy and subspace overlap (see Fig.~\ref{fig:subspace_sim_vs_F_acc} for similar plot but with forgetting and Fig.~\ref{fig:subspace_sim_vc_interference} with interference), i.e. higher subspace overlap correlates with lower CL performance.

\begin{figure}[!t]
    \begin{minipage}[b]{0.49\textwidth}
    \centering
    \includegraphics[width=\linewidth]{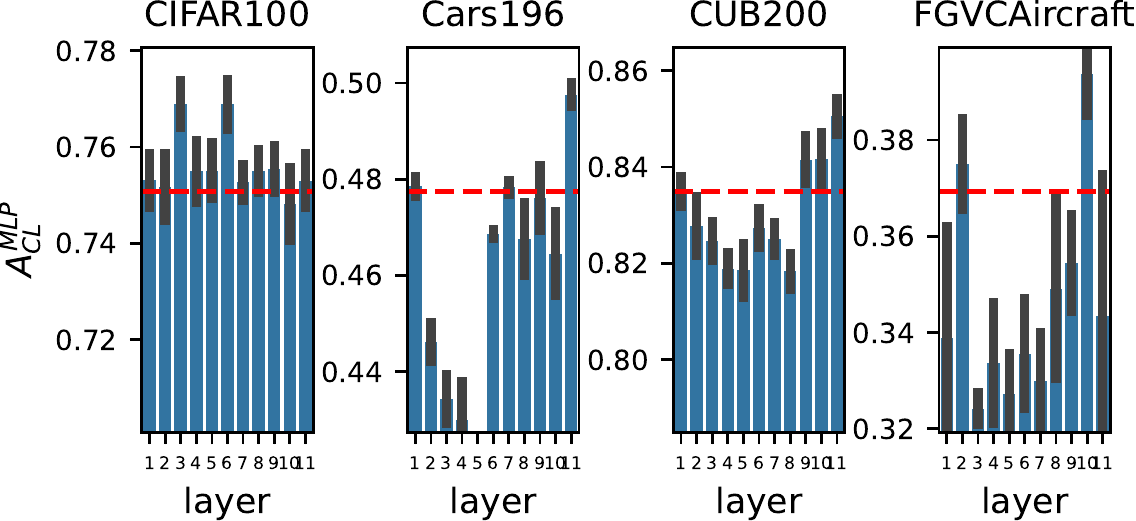}
    \caption{Performance improvement achieved through ensembling deeper and final representations vs. only using final representations (red dotted line). We use ViT-B/16 encoder and ER size of 20 and 5 task orderings. \Message{A slight improvement in CL performance can be achieved by adding deeper representations to the input.} %
    }
    \label{fig:deeper_hidden_layers_latent_ER}
    
    \end{minipage}
    \hfill
    \begin{minipage}[b]{0.49\textwidth}
    \centering
        \includegraphics[width=0.47\linewidth]{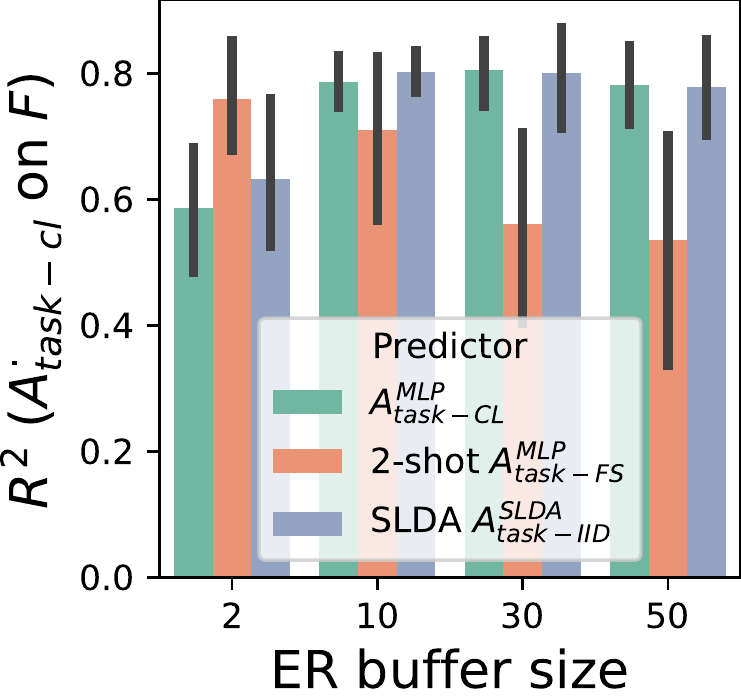}
        \hspace{5pt}
        \includegraphics[width=0.47\linewidth]{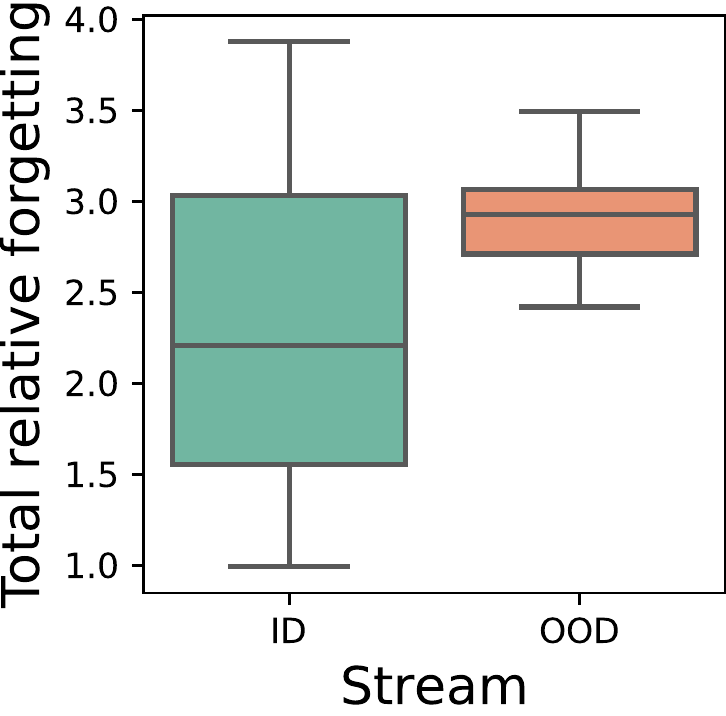}
    \caption{\textbf{Forgetting analysis.}(Left) Coefficient of determination $R^2$ ($\uparrow$) of regression fitted to predict forgetting from $A_{tasks-IID}^{MLP}$, $A_{tasks-FS}^{MLP}$ and $A_{tasks-IID}^{SLDA}$. \Message{Few-shot performance can better explain forgetting in low replay buffer regime.} (Right) Total relative forgetting for ID and OOD streams. \Message{There is more forgetting in OOD regime.}\protect\footnotemark }
    \label{fig:rss_mean_task_acc_on_forgetting}
    \end{minipage}
\end{figure}

\paragraph{Simpler tasks are easier to retain.} 
In order to better understand the forgetting dynamics, we fit a linear regression to predict forgetting using different variables of interest as predictors, namely the per task accuracy of an MLP, $A_{tasks-iid}^{MLP}$, per task accuracy of SLDA, $A_{tasks-iid}^{SLDA}$, as well as the per task few-shot accuracy $A_{tasks-FS}^{MLP}$. Both $A_{tasks-iid}^{SLDA}$ and $A_{tasks-FS}^{MLP}$ describe the notion of tasks simplicity from two different perspectives: a simple task in the few-shot sense is one for which few examples provide enough information to a non-linear classifier for solving it (few examples + complex model); on the other hand, a task can be simple if a simpler model (SLDA) can fit it well with many samples (many examples + simple model). We plot the $R^2$ coefficient of the resulting regressions for different replay buffer sizes in Fig.~\ref{fig:rss_mean_task_acc_on_forgetting}. We observe that few-shot and SLDA accuracy can explain forgetting better (higher $R^2$) than $A_{tasks-iid}^{MLP}$ in the low replay buffer regime -- i.e. in this regime two models can have a similar ability to learn the downstream tasks ($A_{tasks-iid}^{MLP}$), but very different forgetting behaviour. For example, in Fig.~\ref{ap:fig:acc_task_cl_F} we observe that ViT-B/32 and dino\_vitb8 on CIFAR100/5 (ER=2) differ by over 20\% points in forgetting but achieve a comparable accuracy on each of the downstream tasks, i.e. have similar $A_{tasks-iid}^{MLP}$. Somewhat unsurprisingly, we conclude that tasks that are simple in the few-shot sense are easier to retain, i.e. need less replay samples. As expected, for larger replay buffers mean task accuracy becomes better at explaining forgetting than the few-shot accuracy. %

\paragraph{Adding representations from deeper layers can slightly improve performance.} 
\label{sec:head2toe}
In their recent work, \cite{evci2022head2toe} (Head2Toe) showed that ensembling representations from the intermediate and the output layers of a feature extractor tends to improve performance of a linear probe on the OOD data. Inspired by their results, we are interested whether such strategy can lead to better CL results. In Fig.~\ref{fig:deeper_hidden_layers_latent_ER} we show the improvement achieved through ensembling the final output with each of the intermediate layers over using the final layer only (with the ViT-B/16 encoder). Similar to \cite{evci2022head2toe}, we show that there always exists a hidden layer ensembling which always leads to a slight improvement of the CL accuracy (e.g. 2.5\%-points on FGVCAircraft). Additionally, we observe a higher variance in the result on OOD streams (Cars196, FGVCAircraft). The result presented here is intended to serve as a proof-of-concept and we did not implement exactly the Head2Toe solution, which also uses L2 lasso regularization for feature selection. %

\begin{figure}  
    \begin{minipage}[b]{0.59\textwidth}
        \centering    
        \includegraphics[width=1\linewidth]{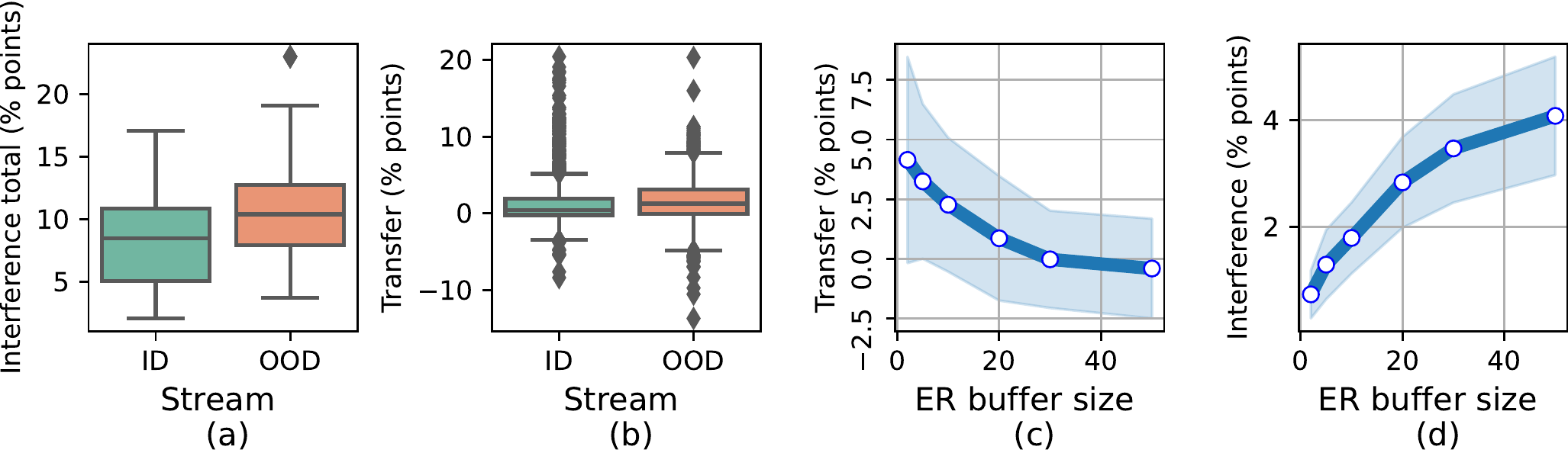}
        \caption{\textbf{Interference and transfer.} \textbf{(a)} Interference total  ($\downarrow$) in ID and OOD streams. \textbf{(b)} Forward transfer ($\uparrow$) in ID and OOD streams.\protect\footnotemark \textbf{(c)} Forward transfer ($\uparrow$) per ER size. \textbf{(d)} Interference ($\downarrow$) per replay buffer size. %
        \Message{Transfer is larger for small ER sizes and in OOD regime. Interference is bigger in OOD regime and larger ER buffer sizes.} }
        \label{fig:transfer_and_interference}
    \end{minipage}
    \hfill
    \begin{minipage}[b]{0.37\textwidth}
        \centering                            
        \includegraphics[width=0.9\linewidth]{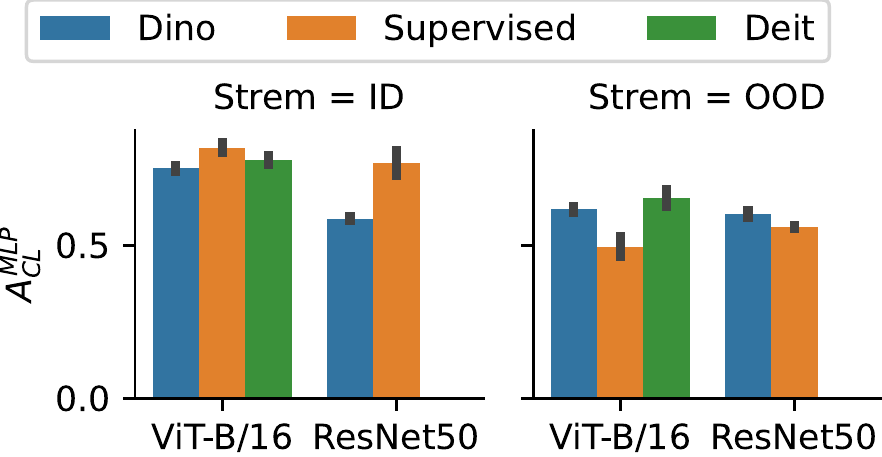}   
        \caption{Final CL accuracy of supervised (ImageNet21k), semi-supervised Dino~\citep{caron2021emerging}, and sup.+distillation DeiT~\citep{touvron2021training} pre-training (ImageNet1k). %
        ER buffer size is 50. Per model view in Fig.~\ref{ap:fig:mlp_slda_nmc_per_stream_per_model}.}
        \label{fig:sup_vs_smi_sup_pretraining}
     \end{minipage}
\end{figure}

\paragraph{Less transfer in OOD streams.} %
In this section we analyze forward transfer as a function of the input data distribution. We define transfer as the difference of final accuracy between a model trained with and without re-initialization before learning each new task (see Tab.~\ref{tab:variables}). In the presence of replay, positive transfer indicates that some information that was present in the weights of the network was not present in the replay buffer. In Fig.~\ref{fig:transfer_and_interference}b we observe that transfer in OOD streams is significantly higher fthan in ID on average. A possible explanation is that ID data leads to more orthogonality between class prototypes. In this case, few samples from the ER buffer can be more representative of the data distribution of past tasks and no additional information must be stored in the weights of the classifier. 
In the OOD regime, the replay buffer is less representative of the downstream data due to higher class and task similarity. Weights contain useful information leading to positive transfer. Counter-intuitively, we can conclude that better representation leads to less transfer.

The type of transfer we study here is between the downstream CL tasks. Thereby the tasks are produced through encoding of only four underlying raw data streams with a variety of encoders. Resulting downstream data distributions have substantially different transfer properties as evidenced by the high amount of variance in the transfer result in Fig.~\ref{fig:transfer_and_interference}b. This highlights the strong impact of data distribution properties on the downstream transfer. Studying downstream transfer at this scale in end2end ER regime can have prohibitive computational cost.
\paragraph{More transfer in small replay buffer regime.}
In Fig.~\ref{fig:transfer_and_interference}c and~\ref{fig:the_big_one} we observe that forward transfer is larger in small ER buffer regimes than in the large ER buffer regime. Intuitively, larger buffer size makes CL closer to offline training reducing the importance of weight initialization for forgetting. It is likely that similar result can be observed in the end2end ER setting, validating it would be compute intensive and we decided not to do it. %

\footnotetext{Welch's test p=1.6e-20, sample size 1330.}

\paragraph{Bigger upstream / downstream distribution shift leads to more interference.}      
Intuitively, joint learning of multiple tasks can hinder performance on each individual task due to conflicting learning signals. This effect is known as interference~\citep{kanakis2020reparameterizing,kokkinos2017ubernet}. We define \textit{total interference} as the difference between the mean task accuracy when each task is learned separately ($A_{task-iid}$) and an offline setting, in which all tasks are learned together ($A_{iid}$) (cf. Tab.~\ref{tab:variables}). Our results show that total interference is lower in the ID datasets than in the OOD datasets (Fig. \ref{fig:transfer_and_interference}a). From our earlier conclusion, we can deduce that the influence factor here is the orthogonality of encoded representations: as visualized in Fig.~\ref{fig:subspace_sim_vc_interference} the more orthogonal the representations are (as measured by subspace overlap), the lower is the total interference. In CL, interference is another factor along forgetting that can lead to  performance decrease. It gives us insight into how much the accuracy on each of the downstream tasks is sacrificed in order to remember previous tasks through replay. In Fig.~\ref{fig:transfer_and_interference}d we visualize this relationship by plotting the interference per replay buffer size. i.e. $A_{task-iid} - A_{task-cl}$. We observe that interference is significantly larger when more samples are replayed. Intuitively, more replay makes learning new tasks harder.

\subsection{Model level analysis}
\begin{wrapfigure}[23]{r}{0.38\textwidth}
    \centering             
    \includegraphics[width=0.85\linewidth]{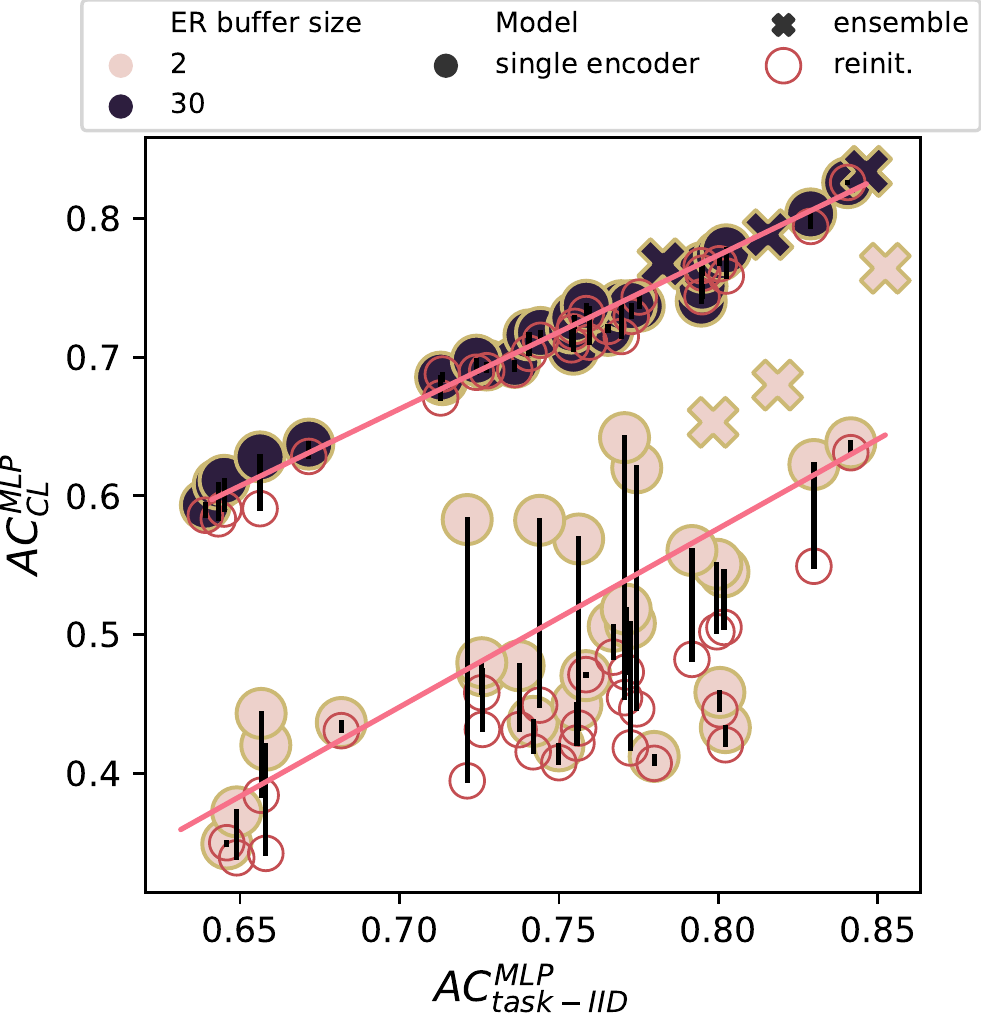}                   
    \caption{ \textbf{$A_{task-IID}^{MLP}$ vs. $A_{CL}^{MLP}$ on multi-dataset stream}.   Hollow red points -- MLP classifier with ER re-initialized before each new task. Vertical lines show forward transfer. Crosses -- ensembled representations. We consider 2 and 30 ER samples per-class on a single task ordering. Detailed view in Fig.~\ref{ap:fig:the_big_one_detailed}.}
    \label{fig:the_big_one}
\end{wrapfigure}

\paragraph{On encoder universality.}                                   
As shown in the appendix  Fig.~\ref{ap:fig:pre-training_dataset_acc_forgetting}, the best performing class of models on average are the CLIP models. This is manifested in the lower forgetting, smaller interference, and, as a result, higher overall $A_{CL}^{MLP}$. While, no encoder dominated the others across all streams, the best performing model is the CLIP ViT-L/14 (2nd largest overall), which is amongst the top-4 best models on all streams as  seen in Fig.~\ref{ap:fig:mlp_slda_nmc_per_stream_per_model}.
\paragraph{Pre-trained encoders encode complementary information.}  
Given varying efficacy of different model families on different streams, an appealing solution could be to use representation ensembling across models. This is motivated by the success of representation ensembling across layers for better OOD generalization(see Sec.~\ref{sec:head2toe}), and across  different views of the same image~\citep{ashukha2021mean} to improve ImageNet performance. Here, we concatenate the representations of several encoders and perform CL with latent ER on top. We validate the efficacy of such solution on the multi-dataset stream in Fig.~\ref{fig:the_big_one}. This scenario is especially suitable for evaluating representation ensembling as it contains each of the datasets used to create single-dataset streams, hence the best performing model on this stream should perform reasonably well on the single-dataset streams as well. As shown in Fig.~\ref{fig:the_big_one},  the best performing ensembling model outperforms the best single encoder model by 10\% points for the replay buffer size of 2, yet only by approximately 0.8\% points for replay buffer size of 30. %
It is worth mentioning that we did not investigate in detail which encoders are most suitable for ensembling of representations (see details in Fig~\ref{ap:fig:dimension_influence}) and leave this exploration for future work.
     
\footnotetext{Welch's test for (a) $p=.00$ and (b) $p=.018$ shows that the means of ID and OOD are significantly different (\# points 1300).}

\paragraph{The role of pre-training regime.} Encoders used in this paper were pre-trained in supervised, self-supervised (SSL), the mixture of both, supervised with distillation and multimodal (CLIP) regimes (see Tab~\ref{tab:models}). %
First, as shown in Fig.~\ref{fig:sup_vs_smi_sup_pretraining}, we observe that semi-supervised dino~\citep{caron2021emerging} encoders outperformed ImageNet21K pretrained supervised counterparts on the fine-grained OOD tasks \textbf{despite} having been pretrained on a much smaller ImageNet1K dataset (same holds for interference and forgetting, see Fig~\ref{ap:fig:sup_vs_elf_sup_pretraining_forgetting_interference}). This can be attributed to the intuition that supervised pre-training learns features needed to discriminate classes in the pre-training dataset (e.g. higher level classes cars vs. planes) which leads to the phenomenon of neural collapse~\citep{galanti2021role,papyan2020prevalence} -- i.e. features of same classes center around its mean feature vector. This leads to the inability of supervised models to generalize to fine-grained OOD streams. SSL regime does not suffer from model collapse and is able to learn richer features resulting in better OOD performance. Interestingly, similar observation is made for the deit encoder~\citep{touvron2021training}, which was pretrained on Imagenet1K dataset using student-teacher distillation. This result extends the observation made by~\citet{gallardo2021self} that SSL pre-training is beneficial for CL especially when number of pre-training samples is small, in that we show that it also depends on relation between the pre-training and the downstream tasks.

\paragraph{Some additional indicators} that we explored did not show clear correlation with CL performance. These include the size of latent dimension (see Fig.\ref{ap:fig:dimension_influence}), the number of parameters in the encoder (see Fig.~\ref{ap:fig:n_params_vs_accCL}). As for the the accuracy of the encoder on the ImageNet1K dataset it positively correlates with CL performance on the ID datasets and rather negatively on the OOD data streams (see Fig.\ref{ap:fig:imagenet_acc_vs_cl}).

\section{Discussion}
       
\paragraph{Actionability of variables.}
In this work, we study a number of  characteristics of pretrained models in CL. Notably, we analyze various variables (e.g. Tab~\ref{tab:variables}) that could explain the final CL performance for a given scenario and pretrained model.
Those variables have different levels of actionability in practice, i.e. not all of them can help to make informed decisions to maximize future performance in the same way.
For instance, some variables are known before accessing the downstream data, e.g. model size, latent dimension size, pre-training FLOPS and pre-training data. This a priori knowledge is highly actionable as it allows selecting models in advance. In our study we found that the influence of the pre-training data overshadows the other variables. In particular, the closeness of pre-training data with downstream data.

On the other hand, in line with~\citet{Douillard2020Insights}, we found that getting access to some information about the downstream tasks a priori can help to improve our predictions about downstream performance. We found that the data distribution shift between pre-training and downstream tasks is the most critical factor. For example, with some meta-knowledge about the downstream task such as its domain, we could select a pretrained model trained on a similar domain to minimize the distribution shift. %
If in-distribution, we found that a few-shot model is a good predictor of downstream performance. Interestingly, we found that it is possible to find if downstream data is in- or out- of distribution without any access to pre-training data using class similarity of downstream embeddings(see Fig.~\ref{fig:sim_density}). %
Without access to information before training, we can still use low compute classifiers such as NMC or SLDA on the first batches of data, to have a fast insight of the most suited encoder for a given scenario.

\paragraph{Continual Learning as a function of data.}
Instead of studying how various algorithms work on a given dataset, we fix the algorithm and changed the data distribution by changing the encoder. %
This approach allows to analyze CL while varying the input streams (data perspective ), while in the literature, CL is mostly analyzed from the algorithmic perspective~\citep{deLange2019continual,BELOUADAH202138}. In other words, instead of studying various algorithms on a normalized benchmark, we use a normalized algorithm that we study on various benchmarks. Our findings on transfer and interference show that data characteristics can be determinant for CL (cf Section \ref{sub:data_analysis}). We believe that understanding the data characteristics better could significantly improve CL performance and help develop ad-hoc approaches.

\paragraph{A new playground for CL.}
By using the pretrained encoders along with existing datasets we were able to generate easy-to-use datasets to prototype and analyze CL algorithms.
The encoded datasets are of high diversity, in part because they are created with encoders pretrained in diverse ways and on diverse datasets. Moreover, the inter-class and inter-task similarity of representation in the generated datasets varies significantly as shown in Fig.~\ref{fig:sim_density}, Fig.~\ref{ap:fig:prototype_similarity}. The variability of the CL results across encoders and classifiers (MLP, NMC, SLDA) is also a good indicator of encoded dataset diversity.
This diverse set of datasets can be used for further research in CL at a significantly lower cost than  end-to-end training on raw data. It does not mean that experimentation should not be done in an end-to-end setting. However, using encoded data is a good intermediate state between training a toy datasets such as MNIST and training on dataset that needs a huge amount of compute such as the ones we used as raw data. It is probable that most conclusions in computationally-demanding scenarios can also be made with a lower amount of compute with a diversified set of scenarios, as it was shown to be for reinforcement learning \citep{pmlr-v139-ceron21a}.
 
Besides class-incremental learning, the CL stream creation strategy used in this study can also be used for domain-incremental experimentation, for example, by creating a sequence of tasks with the same raw-dataset encoded with various encoders to simulate a representation drift \citep{caccia2021special,lesort2021continual}.

\section{Conclusion}                
This paper conducts an extensive empirical evaluation of CL on top of foundation models. We show how latent ER is more compute-efficient than end2end ER and when this compute efficiency can be achieved. Additionally, we show that pre-training on broader data results in a representation space with properties favorable for downstream CL. %
In the extreme case this can allow reduction of CL to training a non-parameteric model on top of encoded features which can have a negligible computational cost and does not require replay by design. This is important because it has the potential to side-step the problem of continual representation learning in some applications. %
We shed some light on the question of how far are we from this extreme case. It can be concluded that whether we can side-step CL of representations mainly depends on the relation between downstream and upstream data: trivially, if the downstream data distribution was covered in the pre-training, the model is more likely to produce useful features for the downstream tasks, and a non-parameteric model can perform well in the CL phase. Development of foundation models with always better generalization abilities brings us ever closer to such an extreme scenario. However, taking a more nuanced perspective, there is a risk that scaling model pre-training can become computationally unsustainable, and, as discussed by \citet{jang2021towards}, foundation models can suffer from their knowledge base becoming outdated. Therefore, an interesting future direction could be the development of algorithms for continual refinement of foundation models. Last but not least, %
this study points out several promising research directions to further improve the efficacy of latent ER in the short term. This includes ensembling of representations coming from different encoders and from the same encoder but different depth.

\newpage
         
\bibliography{continual,collas2022_conference}

\begin{thebibliography}{98}
\providecommand{\natexlab}[1]{#1}
\providecommand{\url}[1]{\texttt{#1}}
\expandafter\ifx\csname urlstyle\endcsname\relax
  \providecommand{\doi}[1]{doi: #1}\else
  \providecommand{\doi}{doi: \begingroup \urlstyle{rm}\Url}\fi

\bibitem[Aljundi et~al.(2017)Aljundi, Babiloni, Elhoseiny, Rohrbach, and
  Tuytelaars]{Aljundi17}
R.~Aljundi, F.~Babiloni, M.~Elhoseiny, M.~Rohrbach, and T.~Tuytelaars.
\newblock Memory aware synapses: Learning what (not) to forget.
\newblock \emph{CoRR}, abs/1711.09601, 2017.
\newblock URL \url{http://arxiv.org/abs/1711.09601}.

\bibitem[Aljundi et~al.(2019{\natexlab{a}})Aljundi, ~, Belilovsky, Caccia, Lin,
  Charlin, and Tuytelaars]{aljundi2019online}
R.~Aljundi, L.~, E.~Belilovsky, M.~Caccia, M.~Lin, L.~Charlin, and
  T.~Tuytelaars.
\newblock Online continual learning with maximal interfered retrieval.
\newblock In H.~Wallach, H.~Larochelle, A.~Beygelzimer, F.~d\textquotesingle
  Alch\'{e}-Buc, E.~Fox, and R.~Garnett, editors, \emph{Advances in Neural
  Information Processing Systems 32}, pages 11849--11860. Curran Associates,
  Inc., 2019{\natexlab{a}}.
\newblock URL
  \url{http://papers.nips.cc/paper/9357-online-continual-learning-with-maximal-interfered-retrieval.pdf}.

\bibitem[Aljundi et~al.(2019{\natexlab{b}})Aljundi, Lin, Goujaud, and
  Bengio]{aljundi2019gradient}
R.~Aljundi, M.~Lin, B.~Goujaud, and Y.~Bengio.
\newblock Gradient based sample selection for online continual learning.
\newblock In H.~Wallach, H.~Larochelle, A.~Beygelzimer, F.~d\textquotesingle
  Alch\'{e}-Buc, E.~Fox, and R.~Garnett, editors, \emph{Advances in Neural
  Information Processing Systems 32}, pages 11816--11825. Curran Associates,
  Inc., 2019{\natexlab{b}}.
\newblock URL
  \url{http://papers.nips.cc/paper/9354-gradient-based-sample-selection-for-online-continual-learning.pdf}.

\bibitem[Ashukha et~al.(2021)Ashukha, Atanov, and Vetrov]{ashukha2021mean}
A.~Ashukha, A.~Atanov, and D.~Vetrov.
\newblock Mean embeddings with test-time data augmentation for ensembling of
  representations.
\newblock \emph{arXiv preprint arXiv:2106.08038}, 2021.

\bibitem[Bagus and Gepperth(2021)]{bagus2021investigation}
B.~Bagus and A.~Gepperth.
\newblock An investigation of replay-based approaches for continual learning.
\newblock In \emph{2021 International Joint Conference on Neural Networks
  (IJCNN)}, pages 1--9. IEEE, 2021.
\newblock URL \url{https://arxiv.org/abs/2108.06758}.

\bibitem[Banayeeanzade et~al.(2021)Banayeeanzade, Mirzaiezadeh, Hasani, and
  Soleymani]{banayeeanzade2021generative}
M.~Banayeeanzade, R.~Mirzaiezadeh, H.~Hasani, and M.~Soleymani.
\newblock Generative vs. discriminative: Rethinking the meta-continual
  learning.
\newblock \emph{Advances in Neural Information Processing Systems}, 34, 2021.
\newblock URL
  \url{https://proceedings.neurips.cc/paper/2021/hash/b4e267d84075f66ebd967d95331fcc03-Abstract.html}.

\bibitem[Belouadah and Popescu(2018)]{Belouadah2018DeeSIL}
E.~Belouadah and A.~Popescu.
\newblock Deesil: Deep-shallow incremental learning.
\newblock In \emph{Proceedings of the European Conference on Computer Vision
  (ECCV)}, pages 0--0, 2018.

\bibitem[Belouadah et~al.(2021)Belouadah, Popescu, and
  Kanellos]{BELOUADAH202138}
E.~Belouadah, A.~Popescu, and I.~Kanellos.
\newblock A comprehensive study of class incremental learning algorithms for
  visual tasks.
\newblock \emph{Neural Networks}, 135:\penalty0 38--54, 2021.
\newblock ISSN 0893-6080.
\newblock \doi{https://doi.org/10.1016/j.neunet.2020.12.003}.
\newblock URL
  \url{https://www.sciencedirect.com/science/article/pii/S0893608020304202}.

\bibitem[Bommasani et~al.(2021)Bommasani, Hudson, Adeli, Altman, Arora, von
  Arx, Bernstein, Bohg, Bosselut, Brunskill,
  et~al.]{bommasani2021opportunities}
R.~Bommasani, D.~A. Hudson, E.~Adeli, R.~Altman, S.~Arora, S.~von Arx, M.~S.
  Bernstein, J.~Bohg, A.~Bosselut, E.~Brunskill, et~al.
\newblock On the opportunities and risks of foundation models.
\newblock \emph{arXiv preprint arXiv:2108.07258}, 2021.

\bibitem[Brown et~al.(2020)Brown, Mann, Ryder, Subbiah, Kaplan, Dhariwal,
  Neelakantan, Shyam, Sastry, Askell, et~al.]{brown2020language}
T.~B. Brown, B.~Mann, N.~Ryder, M.~Subbiah, J.~Kaplan, P.~Dhariwal,
  A.~Neelakantan, P.~Shyam, G.~Sastry, A.~Askell, et~al.
\newblock Language models are few-shot learners.
\newblock \emph{arXiv preprint arXiv:2005.14165}, 2020.

\bibitem[Caccia and Pineau(2021)]{caccia2021special}
L.~Caccia and J.~Pineau.
\newblock Special: Self-supervised pretraining for continual learning.
\newblock In \emph{IJCAI, Workshop on Continual Semi-Supervised Learning},
  2021.

\bibitem[Caccia et~al.(2020)Caccia, Rodriguez, Ostapenko, Normandin, Lin,
  Caccia, Laradji, Rish, Lacoste, Vazquez, and Charlin]{caccia2020online}
M.~Caccia, P.~Rodriguez, O.~Ostapenko, F.~Normandin, M.~Lin, L.~Caccia,
  I.~Laradji, I.~Rish, A.~Lacoste, D.~Vazquez, and L.~Charlin.
\newblock Online fast adaptation and knowledge accumulation: a new approach to
  continual learning.
\newblock \emph{NeurIPS}, 2020.
\newblock URL \url{https://arxiv.org/abs/2003.05856}.

\bibitem[Carion et~al.(2020)Carion, Massa, Synnaeve, Usunier, Kirillov, and
  Zagoruyko]{carion2020end}
N.~Carion, F.~Massa, G.~Synnaeve, N.~Usunier, A.~Kirillov, and S.~Zagoruyko.
\newblock End-to-end object detection with transformers.
\newblock In \emph{European Conference on Computer Vision}, pages 213--229.
  Springer, 2020.

\bibitem[Caron et~al.(2021)Caron, Touvron, Misra, J{\'e}gou, Mairal,
  Bojanowski, and Joulin]{caron2021emerging}
M.~Caron, H.~Touvron, I.~Misra, H.~J{\'e}gou, J.~Mairal, P.~Bojanowski, and
  A.~Joulin.
\newblock Emerging properties in self-supervised vision transformers.
\newblock \emph{arXiv preprint arXiv:2104.14294}, 2021.

\bibitem[Castro et~al.(2018)Castro, Marin-Jimenez, Guil, Schmid, and
  Alahari]{castro2018end}
F.~M. Castro, M.~J. Marin-Jimenez, N.~Guil, C.~Schmid, and K.~Alahari.
\newblock End-to-end incremental learning.
\newblock In \emph{Proceedings of the European Conference on Computer Vision
  (ECCV)}, pages 233--248, 2018.
\newblock URL \url{https://arxiv.org/abs/1807.09536}.

\bibitem[Cermelli et~al.(2020)Cermelli, Mancini, Bulò, Ricci, and
  Caputo]{cermelli2020modeling}
F.~Cermelli, M.~Mancini, S.~R. Bulò, E.~Ricci, and B.~Caputo.
\newblock Modeling the background for incremental learning in semantic
  segmentation, 2020.

\bibitem[Ceron and Castro(2021)]{pmlr-v139-ceron21a}
J.~S.~O. Ceron and P.~S. Castro.
\newblock Revisiting rainbow: Promoting more insightful and inclusive deep
  reinforcement learning research.
\newblock In M.~Meila and T.~Zhang, editors, \emph{Proceedings of the 38th
  International Conference on Machine Learning}, volume 139 of
  \emph{Proceedings of Machine Learning Research}, pages 1373--1383. PMLR,
  18--24 Jul 2021.
\newblock URL \url{https://proceedings.mlr.press/v139/ceron21a.html}.

\bibitem[Chaudhry et~al.(2018)Chaudhry, Ranzato, Rohrbach, and
  Elhoseiny]{chaudhry2018efficient}
A.~Chaudhry, M.~Ranzato, M.~Rohrbach, and M.~Elhoseiny.
\newblock Efficient lifelong learning with a-gem.
\newblock \emph{arXiv preprint arXiv:1812.00420}, 2018.

\bibitem[Chen et~al.(2017)Chen, Papandreou, Kokkinos, Murphy, and
  Yuille]{chen2017deeplab}
L.-C. Chen, G.~Papandreou, I.~Kokkinos, K.~Murphy, and A.~L. Yuille.
\newblock Deeplab: Semantic image segmentation with deep convolutional nets,
  atrous convolution, and fully connected crfs.
\newblock \emph{IEEE transactions on pattern analysis and machine
  intelligence}, 40\penalty0 (4):\penalty0 834--848, 2017.

\bibitem[Chrysakis and Moens(2020)]{chrysakis2020online}
A.~Chrysakis and M.-F. Moens.
\newblock Online continual learning from imbalanced data.
\newblock In \emph{International Conference on Machine Learning}, pages
  1952--1961. PMLR, 2020.

\bibitem[Cimpoi et~al.(2014)Cimpoi, Maji, Kokkinos, Mohamed, , and
  Vedaldi]{cimpoi14describing}
M.~Cimpoi, S.~Maji, I.~Kokkinos, S.~Mohamed, , and A.~Vedaldi.
\newblock Describing textures in the wild.
\newblock In \emph{Proceedings of the {IEEE} Conf. on Computer Vision and
  Pattern Recognition ({CVPR})}, 2014.

\bibitem[Deng et~al.(2009)Deng, Dong, Socher, Li, Li, and
  Fei-Fei]{deng2009imagenet}
J.~Deng, W.~Dong, R.~Socher, L.-J. Li, K.~Li, and L.~Fei-Fei.
\newblock Imagenet: A large-scale hierarchical image database.
\newblock In \emph{2009 IEEE conference on computer vision and pattern
  recognition}, pages 248--255. Ieee, 2009.

\bibitem[Desai et~al.(2021)Desai, Lai, Phan, and Thai]{desai2021continual}
P.~Desai, P.~Lai, N.~Phan, and M.~T. Thai.
\newblock Continual learning with differential privacy.
\newblock In \emph{International Conference on Neural Information Processing},
  pages 334--343. Springer, 2021.

\bibitem[Devlin et~al.(2018)Devlin, Chang, Lee, and Toutanova]{devlin2018bert}
J.~Devlin, M.-W. Chang, K.~Lee, and K.~Toutanova.
\newblock Bert: Pre-training of deep bidirectional transformers for language
  understanding.
\newblock \emph{arXiv preprint arXiv:1810.04805}, 2018.

\bibitem[Diethe et~al.(2018)Diethe, Borchert, Thereska, Pigem, and
  Lawrence]{diethe2018continual}
T.~Diethe, T.~Borchert, E.~Thereska, B.~d.~B. Pigem, and N.~Lawrence.
\newblock Continual learning in practice.
\newblock In \emph{NeurIPS Continual Learning Workshop}, 2018.
\newblock URL \url{https://arxiv.org/abs/1903.05202}.

\bibitem[Dosovitskiy et~al.(2020)Dosovitskiy, Beyer, Kolesnikov, Weissenborn,
  Zhai, Unterthiner, Dehghani, Minderer, Heigold, Gelly,
  et~al.]{dosovitskiy2020image}
A.~Dosovitskiy, L.~Beyer, A.~Kolesnikov, D.~Weissenborn, X.~Zhai,
  T.~Unterthiner, M.~Dehghani, M.~Minderer, G.~Heigold, S.~Gelly, et~al.
\newblock An image is worth 16x16 words: Transformers for image recognition at
  scale.
\newblock In \emph{International Conference on Learning Representations}, 2020.

\bibitem[Douillard and Lesort(2021)]{douillard2021continuum}
A.~Douillard and T.~Lesort.
\newblock Continuum: Simple management of complex continual learning scenarios,
  2021.
\newblock URL \url{https://arxiv.org/abs/2102.06253}.

\bibitem[Douillard et~al.(2020{\natexlab{a}})Douillard, Cord, Ollion, Robert,
  and Valle]{douillard2020podnet}
A.~Douillard, M.~Cord, C.~Ollion, T.~Robert, and E.~Valle.
\newblock Podnet: Pooled outputs distillation for small-tasks incremental
  learning.
\newblock In \emph{Proceedings of the IEEE European Conference on Computer
  Vision (ECCV)}, 2020{\natexlab{a}}.
\newblock URL
  \url{https://www.ecva.net/papers/eccv_2020/papers_ECCV/papers/123650086.pdf}.

\bibitem[Douillard et~al.(2020{\natexlab{b}})Douillard, Valle, Ollion, Robert,
  and Cord]{Douillard2020Insights}
A.~Douillard, E.~Valle, C.~Ollion, T.~Robert, and M.~Cord.
\newblock {Insights from the Future for Continual Learning}.
\newblock \emph{arXiv e-prints}, art. arXiv:2006.13748, June
  2020{\natexlab{b}}.

\bibitem[Douillard et~al.(2021)Douillard, Ram{\'e}, Couairon, and
  Cord]{douillard2021dytox}
A.~Douillard, A.~Ram{\'e}, G.~Couairon, and M.~Cord.
\newblock Dytox: Transformers for continual learning with dynamic token
  expansion.
\newblock \emph{arXiv preprint arXiv:2111.11326}, 2021.
\newblock URL \url{https://arxiv.org/abs/2111.11326}.

\bibitem[Evci et~al.(2022)Evci, Dumoulin, Larochelle, and
  Mozer]{evci2022head2toe}
U.~Evci, V.~Dumoulin, H.~Larochelle, and M.~C. Mozer.
\newblock Head2toe: Utilizing intermediate representations for better transfer
  learning.
\newblock \emph{arXiv preprint arXiv:2201.03529}, 2022.

\bibitem[Galanti et~al.(2021)Galanti, Gy{\"o}rgy, and Hutter]{galanti2021role}
T.~Galanti, A.~Gy{\"o}rgy, and M.~Hutter.
\newblock On the role of neural collapse in transfer learning.
\newblock \emph{arXiv preprint arXiv:2112.15121}, 2021.

\bibitem[Gallardo et~al.(2021)Gallardo, Hayes, and Kanan]{gallardo2021self}
J.~Gallardo, T.~L. Hayes, and C.~Kanan.
\newblock Self-supervised training enhances online continual learning.
\newblock \emph{BMVC}, 2021.

\bibitem[Hayes et~al.(2018)Hayes, Cahill, and Kanan]{Hayes18MemoryEfficient}
T.~L. Hayes, N.~D. Cahill, and C.~Kanan.
\newblock Memory efficient experience replay for streaming learning.
\newblock \emph{CoRR}, abs/1809.05922, 2018.
\newblock URL \url{http://arxiv.org/abs/1809.05922}.

\bibitem[Hayes et~al.(2020)Hayes, Kafle, Shrestha, Acharya, and
  Kanan]{hayes2020remind}
T.~L. Hayes, K.~Kafle, R.~Shrestha, M.~Acharya, and C.~Kanan.
\newblock Remind your neural network to prevent catastrophic forgetting.
\newblock In \emph{European Conference on Computer Vision}, pages 466--483.
  Springer, 2020.

\bibitem[He et~al.(2020)He, Mao, Shao, and Zhu]{He_2020_CVPR}
J.~He, R.~Mao, Z.~Shao, and F.~Zhu.
\newblock Incremental learning in online scenario.
\newblock In \emph{Proceedings of the IEEE/CVF Conference on Computer Vision
  and Pattern Recognition (CVPR)}, June 2020.

\bibitem[Hou et~al.(2019)Hou, Pan, Loy, Wang, and Lin]{Hou_2019_CVPR}
S.~Hou, X.~Pan, C.~C. Loy, Z.~Wang, and D.~Lin.
\newblock Learning a unified classifier incrementally via rebalancing.
\newblock In \emph{The IEEE Conference on Computer Vision and Pattern
  Recognition (CVPR)}, June 2019.

\bibitem[Hu et~al.(2022)Hu, Yan, Lu, HONG, Hu, Zhang, Li, Wang, and
  Feng]{hu2022how}
D.~Hu, S.~Yan, Q.~Lu, L.~HONG, H.~Hu, Y.~Zhang, Z.~Li, X.~Wang, and J.~Feng.
\newblock How well does self-supervised pre-training perform with streaming
  data?
\newblock In \emph{International Conference on Learning Representations}, 2022.
\newblock URL \url{https://openreview.net/forum?id=EwqEx5ipbOu}.

\bibitem[Ioffe and Szegedy(2015)]{ioffe2015batch}
S.~Ioffe and C.~Szegedy.
\newblock Batch normalization: Accelerating deep network training by reducing
  internal covariate shift.
\newblock In \emph{International conference on machine learning}, pages
  448--456. PMLR, 2015.

\bibitem[Jang et~al.(2021)Jang, Ye, Yang, Shin, Han, Kim, Choi, and
  Seo]{jang2021towards}
J.~Jang, S.~Ye, S.~Yang, J.~Shin, J.~Han, G.~Kim, S.~J. Choi, and M.~Seo.
\newblock Towards continual knowledge learning of language models.
\newblock \emph{arXiv preprint arXiv:2110.03215}, 2021.
\newblock URL \url{https://arxiv.org/abs/2110.03215}.

\bibitem[Javed and White(2019)]{javed2019meta}
K.~Javed and M.~White.
\newblock Meta-learning representations for continual learning.
\newblock In H.~Wallach, H.~Larochelle, A.~Beygelzimer, F.~d\textquotesingle
  Alch\'{e}-Buc, E.~Fox, and R.~Garnett, editors, \emph{Advances in Neural
  Information Processing Systems 32}, pages 1818--1828. Curran Associates,
  Inc., 2019.
\newblock URL
  \url{http://papers.nips.cc/paper/8458-meta-learning-representations-for-continual-learning.pdf}.

\bibitem[Ji et~al.(2020)Ji, Lee, Liang, and Poor]{ji2020convergence}
K.~Ji, J.~D. Lee, Y.~Liang, and H.~V. Poor.
\newblock Convergence of meta-learning with task-specific adaptation over
  partial parameters.
\newblock \emph{Advances in Neural Information Processing Systems},
  33:\penalty0 11490--11500, 2020.

\bibitem[Kalifou et~al.(2019)Kalifou, Caselles-Dupré, Lesort, Sun,
  Diaz-Rodriguez, and Filliat]{Kalifou19}
R.~T. Kalifou, H.~Caselles-Dupré, T.~Lesort, T.~Sun, N.~Diaz-Rodriguez, and
  D.~Filliat.
\newblock Continual reinforcement learning deployed in real-life using
  policydistillation and sim2real transfer.
\newblock In \emph{ICML Workshop on Multi-Task and Lifelong Learning}, 2019.

\bibitem[Kanakis et~al.(2020)Kanakis, Bruggemann, Saha, Georgoulis, Obukhov,
  and Van~Gool]{kanakis2020reparameterizing}
M.~Kanakis, D.~Bruggemann, S.~Saha, S.~Georgoulis, A.~Obukhov, and L.~Van~Gool.
\newblock Reparameterizing convolutions for incremental multi-task learning
  without task interference.
\newblock 2020.
\newblock URL
  \url{https://www.ecva.net/papers/eccv_2020/papers_ECCV/papers/123650681.pdf}.

\bibitem[Kaplan et~al.(2020)Kaplan, McCandlish, Henighan, Brown, Chess, Child,
  Gray, Radford, Wu, and Amodei]{kaplan2020scaling}
J.~Kaplan, S.~McCandlish, T.~Henighan, T.~B. Brown, B.~Chess, R.~Child,
  S.~Gray, A.~Radford, J.~Wu, and D.~Amodei.
\newblock Scaling laws for neural language models.
\newblock \emph{arXiv preprint arXiv:2001.08361}, 2020.

\bibitem[Kirkpatrick et~al.(2017)Kirkpatrick, Pascanu, Rabinowitz, Veness,
  Desjardins, Rusu, Milan, Quan, Ramalho, Grabska-Barwinska,
  et~al.]{kirkpatrick2017overcoming}
J.~Kirkpatrick, R.~Pascanu, N.~Rabinowitz, J.~Veness, G.~Desjardins, A.~A.
  Rusu, K.~Milan, J.~Quan, T.~Ramalho, A.~Grabska-Barwinska, et~al.
\newblock Overcoming catastrophic forgetting in neural networks.
\newblock \emph{Proc. of the national academy of sciences}, 2017.
\newblock URL \url{https://www.pnas.org/content/pnas/114/13/3521.full.pdf}.

\bibitem[Kokkinos(2017)]{kokkinos2017ubernet}
I.~Kokkinos.
\newblock Ubernet: Training a universal convolutional neural network for low-,
  mid-, and high-level vision using diverse datasets and limited memory.
\newblock In \emph{Proceedings of the IEEE conference on computer vision and
  pattern recognition}, pages 6129--6138, 2017.

\bibitem[Kolesnikov et~al.(2020)Kolesnikov, Beyer, Zhai, Puigcerver, Yung,
  Gelly, and Houlsby]{kolesnikov2020big}
A.~Kolesnikov, L.~Beyer, X.~Zhai, J.~Puigcerver, J.~Yung, S.~Gelly, and
  N.~Houlsby.
\newblock Big transfer (bit): General visual representation learning.
\newblock In \emph{Computer Vision--ECCV 2020: 16th European Conference,
  Glasgow, UK, August 23--28, 2020, Proceedings, Part V 16}, pages 491--507.
  Springer, 2020.

\bibitem[Krause et~al.(2013)Krause, Stark, Deng, and
  Fei-Fei]{KrauseStarkDengFei-Fei_3DRR2013}
J.~Krause, M.~Stark, J.~Deng, and L.~Fei-Fei.
\newblock 3d object representations for fine-grained categorization.
\newblock In \emph{4th International IEEE Workshop on 3D Representation and
  Recognition (3dRR-13)}, Sydney, Australia, 2013.

\bibitem[Krizhevsky et~al.(2009)Krizhevsky, Hinton,
  et~al.]{krizhevsky2009learning}
A.~Krizhevsky, G.~Hinton, et~al.
\newblock Learning multiple layers of features from tiny images.
\newblock 2009.

\bibitem[Lange et~al.(2019)Lange, Aljundi, Masana, Parisot, Jia, Leonardis,
  Slabaugh, and Tuytelaars]{deLange2019continual}
M.~D. Lange, R.~Aljundi, M.~Masana, S.~Parisot, X.~Jia, A.~Leonardis,
  G.~Slabaugh, and T.~Tuytelaars.
\newblock Continual learning: A comparative study on how to defy forgetting in
  classification tasks, 2019.
\newblock URL \url{https://arxiv.org/abs/1909.08383}.

\bibitem[Lee et~al.(2021)Lee, Goldt, and Saxe]{lee2021continual}
S.~Lee, S.~Goldt, and A.~Saxe.
\newblock Continual learning in the teacher-student setup: Impact of task
  similarity.
\newblock In \emph{International Conference on Machine Learning}, pages
  6109--6119. PMLR, 2021.

\bibitem[Lesort(2020)]{lesort2020continual}
T.~Lesort.
\newblock Continual learning: Tackling catastrophic forgetting in deep neural
  networks with replay processes, 2020.
\newblock URL \url{https://arxiv.org/abs/2007.00487}.

\bibitem[Lesort et~al.(2019)Lesort, Stoian, and
  Filliat]{lesort2019regularization}
T.~Lesort, A.~Stoian, and D.~Filliat.
\newblock Regularization shortcomings for continual learning.
\newblock \emph{arXiv preprint arXiv:1912.03049}, 2019.

\bibitem[Lesort et~al.(2021{\natexlab{a}})Lesort, Caccia, and
  Rish]{lesort2021understanding}
T.~Lesort, M.~Caccia, and I.~Rish.
\newblock Understanding continual learning settings with data distribution
  drift analysis.
\newblock \emph{arXiv preprint arXiv:2104.01678}, 2021{\natexlab{a}}.

\bibitem[Lesort et~al.(2021{\natexlab{b}})Lesort, George, and
  Rish]{lesort2021continual}
T.~Lesort, T.~George, and I.~Rish.
\newblock Continual learning in deep networks: an analysis of the last layer.
\newblock \emph{arXiv preprint arXiv:2106.01834}, 2021{\natexlab{b}}.
\newblock URL \url{https://arxiv.org/abs/2106.01834}.

\bibitem[Liu et~al.(2020)Liu, Wu, Menta, Herranz, Raducanu, Bagdanov, Jui, and
  de~Weijer]{liu2020generative}
X.~Liu, C.~Wu, M.~Menta, L.~Herranz, B.~Raducanu, A.~D. Bagdanov, S.~Jui, and
  J.~v. de~Weijer.
\newblock Generative feature replay for class-incremental learning.
\newblock In \emph{Proceedings of the IEEE/CVF Conference on Computer Vision
  and Pattern Recognition Workshops}, pages 226--227, 2020.

\bibitem[Lopez-Paz and Ranzato(2017)]{Lopez-Paz17}
D.~Lopez-Paz and M.-A. Ranzato.
\newblock Gradient episodic memory for continual learning.
\newblock In I.~Guyon, U.~V. Luxburg, S.~Bengio, H.~Wallach, R.~Fergus,
  S.~Vishwanathan, and R.~Garnett, editors, \emph{Advances in Neural
  Information Processing Systems 30}, pages 6467--6476. Curran Associates,
  Inc., 2017.
\newblock URL
  \url{http://papers.nips.cc/paper/7225-gradient-episodic-memory-for-continual-learning.pdf}.

\bibitem[Maji et~al.(2013)Maji, Kannala, Rahtu, Blaschko, and
  Vedaldi]{maji13fine-grained}
S.~Maji, J.~Kannala, E.~Rahtu, M.~Blaschko, and A.~Vedaldi.
\newblock Fine-grained visual classification of aircraft.
\newblock Technical report, 2013.

\bibitem[Mehta et~al.(2022)Mehta, Patil, Chandar, and Strubell]{mehta2022an}
S.~V. Mehta, D.~Patil, S.~Chandar, and E.~Strubell.
\newblock An empirical investigation of the role of pre-training in lifelong
  learning, 2022.
\newblock URL \url{https://openreview.net/forum?id=D9E8MKsfhw}.

\bibitem[Mendez and EATON(2021)]{mendez2021lifelong}
J.~A. Mendez and E.~EATON.
\newblock Lifelong learning of compositional structures.
\newblock In \emph{International Conference on Learning Representations}, 2021.
\newblock URL \url{https://openreview.net/forum?id=ADWd4TJO13G}.

\bibitem[Nguyen et~al.(2019)Nguyen, Achille, Lam, Hassner, Mahadevan, and
  Soatto]{nguyen2019toward}
C.~V. Nguyen, A.~Achille, M.~Lam, T.~Hassner, V.~Mahadevan, and S.~Soatto.
\newblock Toward understanding catastrophic forgetting in continual learning.
\newblock \emph{arXiv preprint arXiv:1908.01091}, 2019.

\bibitem[Ostapenko et~al.(2021)Ostapenko, Rodriguez, Caccia, and
  Charlin]{ostapenko2021continual}
O.~Ostapenko, P.~Rodriguez, M.~Caccia, and L.~Charlin.
\newblock Continual learning via local module composition.
\newblock In \emph{Thirty-Fifth Conference on Neural Information Processing
  Systems}, 2021.
\newblock URL
  \url{https://proceedings.neurips.cc/paper/2021/hash/fe5e7cb609bdbe6d62449d61849c38b0-Abstract.html}.

\bibitem[Papyan et~al.(2020)Papyan, Han, and Donoho]{papyan2020prevalence}
V.~Papyan, X.~Han, and D.~L. Donoho.
\newblock Prevalence of neural collapse during the terminal phase of deep
  learning training.
\newblock \emph{Proceedings of the National Academy of Sciences}, 117\penalty0
  (40):\penalty0 24652--24663, 2020.

\bibitem[Parkhi et~al.(2012)Parkhi, Vedaldi, Zisserman, and
  Jawahar]{parkhi2012cats}
O.~M. Parkhi, A.~Vedaldi, A.~Zisserman, and C.~Jawahar.
\newblock Cats and dogs.
\newblock In \emph{2012 IEEE conference on computer vision and pattern
  recognition}, pages 3498--3505. IEEE, 2012.

\bibitem[Pellegrini et~al.(2020)Pellegrini, Graffieti, Lomonaco, and
  Maltoni]{pellegrini2020latent}
L.~Pellegrini, G.~Graffieti, V.~Lomonaco, and D.~Maltoni.
\newblock Latent replay for real-time continual learning, 2020.

\bibitem[Prabhu et~al.(2020)Prabhu, Torr, and Dokania]{prabhu2020gdumb}
A.~Prabhu, P.~H. Torr, and P.~K. Dokania.
\newblock Gdumb: A simple approach that questions our progress in continual
  learning.
\newblock In \emph{European Conference on Computer Vision}, pages 524--540.
  Springer, 2020.

\bibitem[Radford et~al.(2021)Radford, Kim, Hallacy, Ramesh, Goh, Agarwal,
  Sastry, Askell, Mishkin, Clark, et~al.]{radford2021learning}
A.~Radford, J.~W. Kim, C.~Hallacy, A.~Ramesh, G.~Goh, S.~Agarwal, G.~Sastry,
  A.~Askell, P.~Mishkin, J.~Clark, et~al.
\newblock Learning transferable visual models from natural language
  supervision.
\newblock \emph{arXiv preprint arXiv:2103.00020}, 2021.

\bibitem[Ramasesh et~al.(2020)Ramasesh, Dyer, and Raghu]{ramasesh2020anatomy}
V.~V. Ramasesh, E.~Dyer, and M.~Raghu.
\newblock Anatomy of catastrophic forgetting: Hidden representations and task
  semantics.
\newblock \emph{arXiv preprint arXiv:2007.07400}, 2020.

\bibitem[Ramasesh et~al.(2021)Ramasesh, Dyer, and Raghu]{ramasesh2021anatomy}
V.~V. Ramasesh, E.~Dyer, and M.~Raghu.
\newblock Anatomy of catastrophic forgetting: Hidden representations and task
  semantics.
\newblock In \emph{International Conference on Learning Representations}, 2021.
\newblock URL \url{https://openreview.net/forum?id=LhY8QdUGSuw}.

\bibitem[Ramasesh et~al.(2022)Ramasesh, Lewkowycz, and
  Dyer]{ramasesh2022effect}
V.~V. Ramasesh, A.~Lewkowycz, and E.~Dyer.
\newblock Effect of scale on catastrophic forgetting in neural networks.
\newblock In \emph{International Conference on Learning Representations}, 2022.
\newblock URL \url{https://openreview.net/forum?id=GhVS8_yPeEa}.

\bibitem[Reddy et~al.(2021)Reddy, Basha, Hari, and Penchalaiah]{reddy2021dall}
M.~D.~M. Reddy, M.~S.~M. Basha, M.~M.~C. Hari, and M.~N. Penchalaiah.
\newblock Dall-e: Creating images from text.
\newblock 2021.

\bibitem[Ren et~al.(2015)Ren, He, Girshick, and Sun]{ren2015faster}
S.~Ren, K.~He, R.~Girshick, and J.~Sun.
\newblock Faster r-cnn: Towards real-time object detection with region proposal
  networks.
\newblock \emph{Advances in neural information processing systems},
  28:\penalty0 91--99, 2015.

\bibitem[Ridnik et~al.(2021)Ridnik, Ben-Baruch, Noy, and
  Zelnik-Manor]{ridnik2021imagenet}
T.~Ridnik, E.~Ben-Baruch, A.~Noy, and L.~Zelnik-Manor.
\newblock Imagenet-21k pretraining for the masses.
\newblock \emph{arXiv preprint arXiv:2104.10972}, 2021.

\bibitem[Sanh et~al.(2021)Sanh, Webson, Raffel, Bach, Sutawika, Alyafeai,
  Chaffin, Stiegler, Scao, Raja, et~al.]{sanh2021multitask}
V.~Sanh, A.~Webson, C.~Raffel, S.~H. Bach, L.~Sutawika, Z.~Alyafeai,
  A.~Chaffin, A.~Stiegler, T.~L. Scao, A.~Raja, et~al.
\newblock Multitask prompted training enables zero-shot task generalization.
\newblock \emph{arXiv preprint arXiv:2110.08207}, 2021.

\bibitem[Saxe et~al.(2019)Saxe, Bansal, Dapello, Advani, Kolchinsky, Tracey,
  and Cox]{saxe2019information}
A.~M. Saxe, Y.~Bansal, J.~Dapello, M.~Advani, A.~Kolchinsky, B.~D. Tracey, and
  D.~D. Cox.
\newblock On the information bottleneck theory of deep learning.
\newblock \emph{Journal of Statistical Mechanics: Theory and Experiment},
  2019\penalty0 (12):\penalty0 124020, 2019.

\bibitem[Schwarz et~al.(2018)Schwarz, Luketina, Czarnecki, Grabska-Barwinska,
  Teh, Pascanu, and Hadsell]{schwarz2018progress}
J.~Schwarz, J.~Luketina, W.~M. Czarnecki, A.~Grabska-Barwinska, Y.~W. Teh,
  R.~Pascanu, and R.~Hadsell.
\newblock Progress \& compress: A scalable framework for continual learning.
\newblock In \emph{ICML}, 2018.
\newblock URL \url{https://arxiv.org/abs/1805.06370}.

\bibitem[{Shaoning Pang} et~al.(2005){Shaoning Pang}, {Ozawa}, and
  {Kasabov}]{PangIncremental2005}
{Shaoning Pang}, S.~{Ozawa}, and N.~{Kasabov}.
\newblock Incremental linear discriminant analysis for classification of data
  streams.
\newblock \emph{IEEE Transactions on Systems, Man, and Cybernetics, Part B
  (Cybernetics)}, 35\penalty0 (5):\penalty0 905--914, 2005.
\newblock \doi{10.1109/TSMCB.2005.847744}.

\bibitem[Sun et~al.(2017)Sun, Shrivastava, Singh, and Gupta]{sun2017revisiting}
C.~Sun, A.~Shrivastava, S.~Singh, and A.~Gupta.
\newblock Revisiting unreasonable effectiveness of data in deep learning era.
\newblock In \emph{Proceedings of the IEEE international conference on computer
  vision}, pages 843--852, 2017.

\bibitem[Szegedy et~al.(2017)Szegedy, Ioffe, Vanhoucke, and
  Alemi]{szegedy2017inception}
C.~Szegedy, S.~Ioffe, V.~Vanhoucke, and A.~A. Alemi.
\newblock Inception-v4, inception-resnet and the impact of residual connections
  on learning.
\newblock In \emph{Thirty-first AAAI conference on artificial intelligence},
  2017.

\bibitem[Tan and Le(2019)]{tan2019efficientnet}
M.~Tan and Q.~Le.
\newblock Efficientnet: Rethinking model scaling for convolutional neural
  networks.
\newblock In \emph{International Conference on Machine Learning}, pages
  6105--6114. PMLR, 2019.

\bibitem[Tomasev et~al.(2022)Tomasev, Bica, McWilliams, Buesing, Pascanu,
  Blundell, and Mitrovic]{tomasev2022pushing}
N.~Tomasev, I.~Bica, B.~McWilliams, L.~Buesing, R.~Pascanu, C.~Blundell, and
  J.~Mitrovic.
\newblock Pushing the limits of self-supervised resnets: Can we outperform
  supervised learning without labels on imagenet?
\newblock \emph{arXiv preprint arXiv:2201.05119}, 2022.

\bibitem[Touvron et~al.(2021)Touvron, Cord, Douze, Massa, Sablayrolles, and
  J{\'e}gou]{touvron2021training}
H.~Touvron, M.~Cord, M.~Douze, F.~Massa, A.~Sablayrolles, and H.~J{\'e}gou.
\newblock Training data-efficient image transformers \& distillation through
  attention.
\newblock In \emph{International Conference on Machine Learning}, pages
  10347--10357. PMLR, 2021.

\bibitem[Traoré et~al.(2019)Traoré, Caselles{-}Dupré, Lesort, Sun, Cai,
  Rodr{\'{\i}}guez, and Filliat]{Traore19DisCoRL}
R.~Traoré, H.~Caselles{-}Dupré, T.~Lesort, T.~Sun, G.~Cai, N.~D.
  Rodr{\'{\i}}guez, and D.~Filliat.
\newblock Discorl: Continual reinforcement learning via policy distillation.
\newblock \emph{CoRR}, abs/1907.05855, 2019.
\newblock URL \url{http://arxiv.org/abs/1907.05855}.

\bibitem[van~de Ven and Tolias(2019)]{van2019three}
G.~M. van~de Ven and A.~S. Tolias.
\newblock Three scenarios for continual learning.
\newblock \emph{arXiv preprint arXiv:1904.07734}, 2019.
\newblock URL \url{https://arxiv.org/abs/1904.07734}.

\bibitem[van~de Ven et~al.(2020)van~de Ven, Siegelmann, and
  Tolias]{van2020brain}
G.~M. van~de Ven, H.~T. Siegelmann, and A.~S. Tolias.
\newblock Brain-inspired replay for continual learning with artificial neural
  networks.
\newblock \emph{Nature communications}, 11\penalty0 (1):\penalty0 1--14, 2020.

\bibitem[Veniat et~al.(2021)Veniat, Denoyer, and Ranzato]{veniat2021efficient}
T.~Veniat, L.~Denoyer, and M.~Ranzato.
\newblock Efficient continual learning with modular networks and task-driven
  priors.
\newblock In \emph{International Conference on Learning Representations}, 2021.
\newblock URL \url{https://openreview.net/forum?id=EKV158tSfwv}.

\bibitem[Welinder et~al.(2010)Welinder, Branson, Mita, Wah, Schroff, Belongie,
  and Perona]{WelinderEtal2010}
P.~Welinder, S.~Branson, T.~Mita, C.~Wah, F.~Schroff, S.~Belongie, and
  P.~Perona.
\newblock {Caltech-UCSD Birds 200}.
\newblock Technical Report CNS-TR-2010-001, California Institute of Technology,
  2010.

\bibitem[Wen et~al.(2022)Wen, Rios, Lekkala, and Itti]{wen2022can}
S.~Wen, A.~S. Rios, K.~Lekkala, and L.~Itti.
\newblock What can we learn from misclassified imagenet images?
\newblock \emph{arXiv preprint arXiv:2201.08098}, 2022.

\bibitem[Wu et~al.(2022)Wu, Caccia, Li, Li, Qi, and Haffari]{wu2022pretrained}
T.~Wu, M.~Caccia, Z.~Li, Y.-F. Li, G.~Qi, and G.~Haffari.
\newblock Pretrained language model in continual learning: A comparative study.
\newblock In \emph{International Conference on Learning Representations}, 2022.
\newblock URL \url{https://openreview.net/forum?id=figzpGMrdD}.

\bibitem[Wu et~al.(2019)Wu, Chen, Wang, Ye, Liu, Guo, and Fu]{wu2019large}
Y.~Wu, Y.~Chen, L.~Wang, Y.~Ye, Z.~Liu, Y.~Guo, and Y.~Fu.
\newblock Large scale incremental learning.
\newblock In \emph{Proceedings of the IEEE/CVF Conference on Computer Vision
  and Pattern Recognition}, pages 374--382, 2019.
\newblock URL \url{https://arxiv.org/abs/1905.13260}.

\bibitem[Xu et~al.(2021)Xu, Baracaldo, and Joshi]{xu2021privacy}
R.~Xu, N.~Baracaldo, and J.~Joshi.
\newblock Privacy-preserving machine learning: Methods, challenges and
  directions.
\newblock \emph{arXiv preprint arXiv:2108.04417}, 2021.

\bibitem[Yalniz et~al.(2019)Yalniz, J{\'e}gou, Chen, Paluri, and
  Mahajan]{yalniz2019billion}
I.~Z. Yalniz, H.~J{\'e}gou, K.~Chen, M.~Paluri, and D.~Mahajan.
\newblock Billion-scale semi-supervised learning for image classification.
\newblock \emph{arXiv preprint arXiv:1905.00546}, 2019.

\bibitem[Yoon et~al.(2017)Yoon, Yang, Lee, and Hwang]{yoon2017lifelong}
J.~Yoon, E.~Yang, J.~Lee, and S.~J. Hwang.
\newblock Lifelong learning with dynamically expandable networks.
\newblock \emph{arXiv preprint arXiv:1708.01547}, 2017.

\bibitem[Yosinski et~al.(2014)Yosinski, Clune, Bengio, and
  Lipson]{yosinski2014transferable}
J.~Yosinski, J.~Clune, Y.~Bengio, and H.~Lipson.
\newblock How transferable are features in deep neural networks?
\newblock \emph{Advances in Neural Information Processing Systems},
  27:\penalty0 3320--3328, 2014.

\bibitem[Yuan et~al.(2021)Yuan, Chen, Chen, Codella, Dai, Gao, Hu, Huang, Li,
  Li, et~al.]{yuan2021florence}
L.~Yuan, D.~Chen, Y.-L. Chen, N.~Codella, X.~Dai, J.~Gao, H.~Hu, X.~Huang,
  B.~Li, C.~Li, et~al.
\newblock Florence: A new foundation model for computer vision.
\newblock \emph{arXiv preprint arXiv:2111.11432}, 2021.

\bibitem[Zagoruyko and Komodakis(2016)]{zagoruyko2016wide}
S.~Zagoruyko and N.~Komodakis.
\newblock Wide residual networks.
\newblock In \emph{British Machine Vision Conference 2016}. British Machine
  Vision Association, 2016.

\bibitem[Zhai et~al.(2022)Zhai, Wang, Mustafa, Steiner, Keysers, Kolesnikov,
  and Beyer]{zhai2022lit}
X.~Zhai, X.~Wang, B.~Mustafa, A.~Steiner, D.~Keysers, A.~Kolesnikov, and
  L.~Beyer.
\newblock Lit: Zero-shot transfer with locked-image text tuning.
\newblock In \emph{Proceedings of the IEEE/CVF Conference on Computer Vision
  and Pattern Recognition}, pages 18123--18133, 2022.

\end{thebibliography}
\bibliographystyle{abbrvnat}

\section*{Acknowledgement}
Laurent Charlin holds a CIFAR AI Chair Program and acknowledges support from Samsung Electronics Co., Ldt.. We would like to thank Mila and Compute Canada for providing computational resources.
\newpage

\appendix
\onecolumn

\section*{Appendix}

\section{Detailed Motivation for Latent ER}
\label{ap:motivation_latent}

\paragraph{Potential of latent ER} 
With the fast development of foundation models and their effectiveness across tasks and domains, latent CL becomes a more and more appealing alternative to the end2end fine-tuning. This is due to (a) CL time compute advantage enabling per-task hyperparameter search and fast on-device training, (b) potential to improve privacy since storing latent representations of a privacy-preserving model~\citep{xu2021privacy} for replay may be more secure than storing raw data, as well as (c) better data efficiency since latent representations contain less task-irrelevant information such as pixel noise~\citep{saxe2019information}. (d) As we discuss in the main paper, latent CL can be reduced to training a non-parametric model on top of encoded data, which would solve several long standing problems of CL such as task inference~\citep{van2019three}, and could relegate replay as a method of choice for CL, as non-parametric models do not require replay by design~\citep{banayeeanzade2021generative}. 
In this paper we shed some light on the question of how far are we from the extreme case described above. All in all, given the high potential impact of foundation models on the field of CL we believe that the insights provided in this paper are of high relevance for the CL community.

 \paragraph{Applicability of latent ER}              
Continual learning in the latent space is akin to using the encoding procedure as a data pre-processing step. In Section~\ref{sec:computation}, we have seen that this methodology can produce representations that outperform raw data in terms of CL performance and computation. To succeed, latent ER should encode class discriminative information as good as it is encoded in the the raw data (i.e. no task-relevant information is destroyed). 
This can be true in at least two cases: either the pre-training domain is similar to the downstream CL domain (Figure~\ref{fig:compute-acc}), or the pre-training domain is extremely broad (Figure~\ref{fig:architectures}), such that the produced representations are likely to work with any domain encountered in the CL phase. An example of the first case is a situation in which the CL domain is known a priori, such as pre-training a feature encoder for self-driving applications. The latter case is addressed through the development of foundation models~\citep{radford2021learning,bommasani2021opportunities}. It is important to highlight that the goal of foundation models is specifically the creation of models effective across wide range of tasks and domains, i.e. models for which all downstream tasks will be ID. This makes latent ER an increasingly promising strategy for CL. 

\section{Latent ER vs. end2end ER}
\label{ap:compute_acc_trade}

We calculate number of flops required by the models using the fvcore   library\footnote{\url{https://github.com/facebookresearch/fvcore}}.  Thereby the computational cost of a backward path is roughly twice as large as the computation cost of the forward path.\protect\footnotemark~. Unless otherwise stated, when estimating the computational cost of latent ER models, we take into account the encoding cost, that is the cost of forward passing each sample through the encoder one time.
\footnotetext{\url{https://openai.com/blog/ai-and-compute/}}

\begin{figure*}[h]  
        \centering     
        \begin{subfigure}[b]{1\linewidth}
            \includegraphics[width=1.\textwidth]{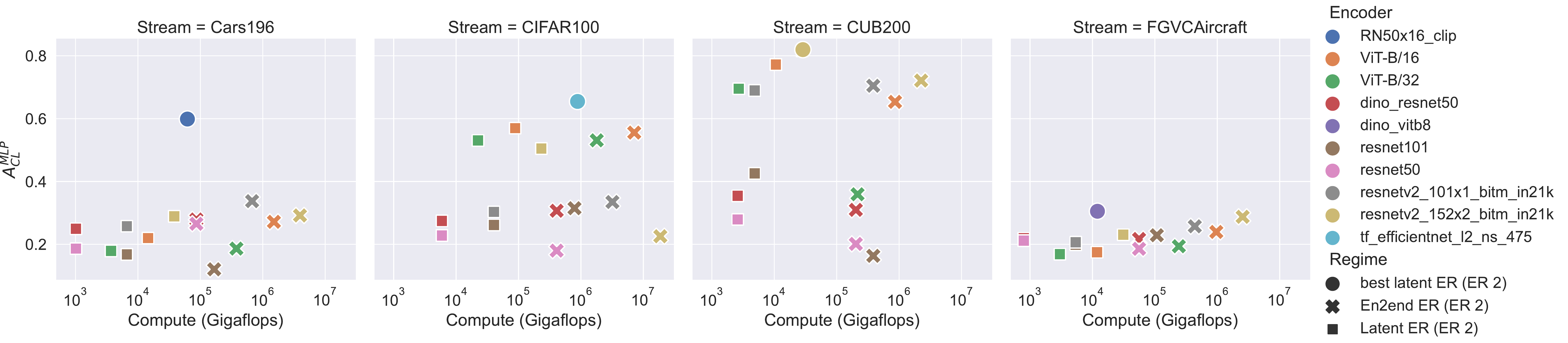}
            \caption{ER buffer size 2}
        \end{subfigure}
        \begin{subfigure}[b]{1\linewidth} 
            \includegraphics[width=1.\textwidth]{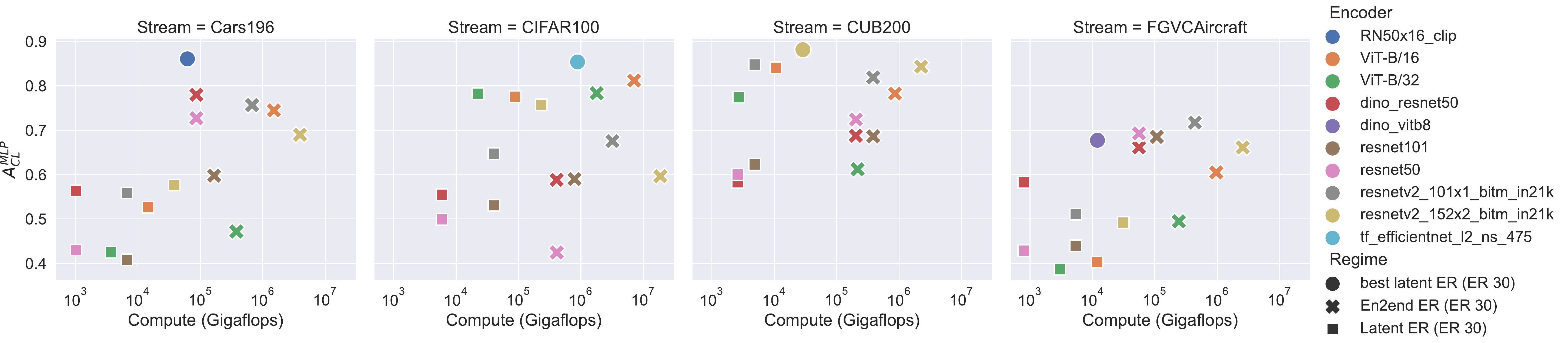} 
            \caption{ER buffer size 30}
        \end{subfigure}
\caption{ 
\textbf{Compute-accuracy trade-off} of raw data ER (end2end) vs. latent ER models. We plot 7 different models in both end2end ER and latent ER regime on 4 streams and using 2 ER buffer sizes (2 and 30 samples per class). Additionally, we plot the best latent ER model for each stream (e.g. dino\_vitb8 \protect\footnotemark model in the FGVCAircraft stream) \Message{End2end ER leads to performance improvement on the OOD streams (Cars196 and FGVCAircraft) at the expense of more compute. The best performing latent ER model outperforms or reaches comparable accuracy of the best end2end ER model on all streams.}
}
\label{ap:fig:compute-acc}
\end{figure*}

\section{Computational cost of metric based models}
\label{ap:sec:metric_based_compute}
Metric-based models used in this paper use statistics of data to classify them. Hence, the nearest mean classifier computes the mean of features of each class and looks for the closest ( $L_2$ norm ) mean to classify a new data point.
When the feature encoder is frozen the full compute cost of NMC is then: $C_{enc}*Card(\mathbb{D})+ Card(\mathbb{D})*N_z + N_z*N_c$
With $C_{enc}$ the number of flops to encode one image, $Card(\mathbb{D})$ the number of sample in the dataset/task $\mathbb{D}$, $N_z$ the latent dimension size and $N_c$. $Card(\mathbb{D})*N_z$ correspond in summing all latent vectors and $N_z*N_c$ to normalize vectors for each class.

The SLDA approach is more compute consuming since it needs to compute means and covariance matrice but also because it needs to invert the covariance matrix for inference. The covariance matrix is a $N_z \times N_z$ matrix computed such as:
\begin{equation}
    cov(z_i,z_j) =  \mathbb{E}[(z_i - \mathbb{E}[z_i])(z_j - \mathbb{E}[z_j])]
\end{equation}
With $z_i$ the dimension $i \in [0, N_z-1]$ of the latent representation. Assuming $\mathbb{E}[z_i]$ and $\mathbb{E}[z_i]$ already computed while estimating the mean of classes. The compute cost for the full matrix $\Sigma$ is:
$(3 * N_z \times \mathbb{D})^2$.
The inversion of a matrice has a complexity of $O(n^3)$, so we can roughly estimate the compute to be similar to ${N_z}^3$ flops to invert the covariance matrice.
The total compute of SLDA is then roughly: $C_{enc}*Card(\mathbb{D})+ Card(\mathbb{D})*N_z + N_z*N_c + (3 * N_z \times \mathbb{D})^2 + {N_z}^3 $.
With a large latent dimension, the biggest cost of SLDA is the covariance matrix inversion. Hence depending on the representation size SLDA might be more or less computed efficiently. For example, with our biggest representation size (8192 \textit{resnetv2-152x4-bitm}), the matrix inversion is around $5.5 \times 10^{2}$ gigaflops which is still lower than latent replay compute (cf Figure \ref{ap:fig:compute-acc}) but several order of magnitude bigger than NMC.

\footnotetext{for dino\_vitb8 we approximated the encoding compute with that of the ViT-B/16 model, both have comparable number of parameters and similar computational graph}

\section{Task and class similarity}
\label{ap:sec:similarity}

\textbf{Subspace overlap} measures the average task similarity of a sequence by comparing the orthogonal sub spaces of each task with each other task. If $n$ distinct tasks are represented by their centered latent representations $(t_1, t_2, . . . , t_n)$ in a sequence and $t_i\in R^{p\times q}$, $p$ is the number of samples and $q$ is the latent dimension:
\begin{equation}
    Overlap_{avg} = \frac{1}{T} \sum_{i=1}^{n} \sum_{j=i+1}^{n}  Overlap_{subs}(t_i, t_j),
    \label{eq:overlap_avg}
\end{equation}
Here, $T$ is the number of pairs, where each task representations are compared with the upcoming ones.
\begin{equation}
    Overlap_{subs}(t_i, t_j) =  \frac{1}{k}\| U_k^TV_k\|^2_F.
    \label{eq:overlap_subs}
\end{equation}
Here, $U_k$ be the matrix formed by eigenvectors of $t_i^Tt_i$ for top $k$ principal directions and $V_k$ be the matrix formed by eigenvectors of $t_j^Tt_j$ for top $k$ principal directions. For the subspace similarity analysis, we fix $k=20$, which captures around $56$\% of average variance of the latent representations for all the datasets and models.

To measure the \textbf{class similarities} in a sequence, we represent each class by a prototype which is computed by the mean of latent representations of the samples of that class and measure the pairwise cosine similarities. Given $c$ classes, each class is represented by prototypes $(p_1, p_2, . . . , p_c)$, where each $p_i$ is a vector in $R^{q}$. Class similarity is defined as:
\begin{equation}
    ClassSim_{avg} = \frac{1}{T} \sum_{i=1}^{c} \sum_{j=i+1}^{c}  CosSim(p_i, p_j),
    \label{eq:prototypes}
\end{equation}
Here, $T$ is the number of possible pairs of prototypes and $CosSim$ is the cosine similarity function.

\begin{figure}[h!]
    \centering
    \begin{subfigure}[b]{0.45\linewidth}
        \centering
        \includegraphics[width=\linewidth]{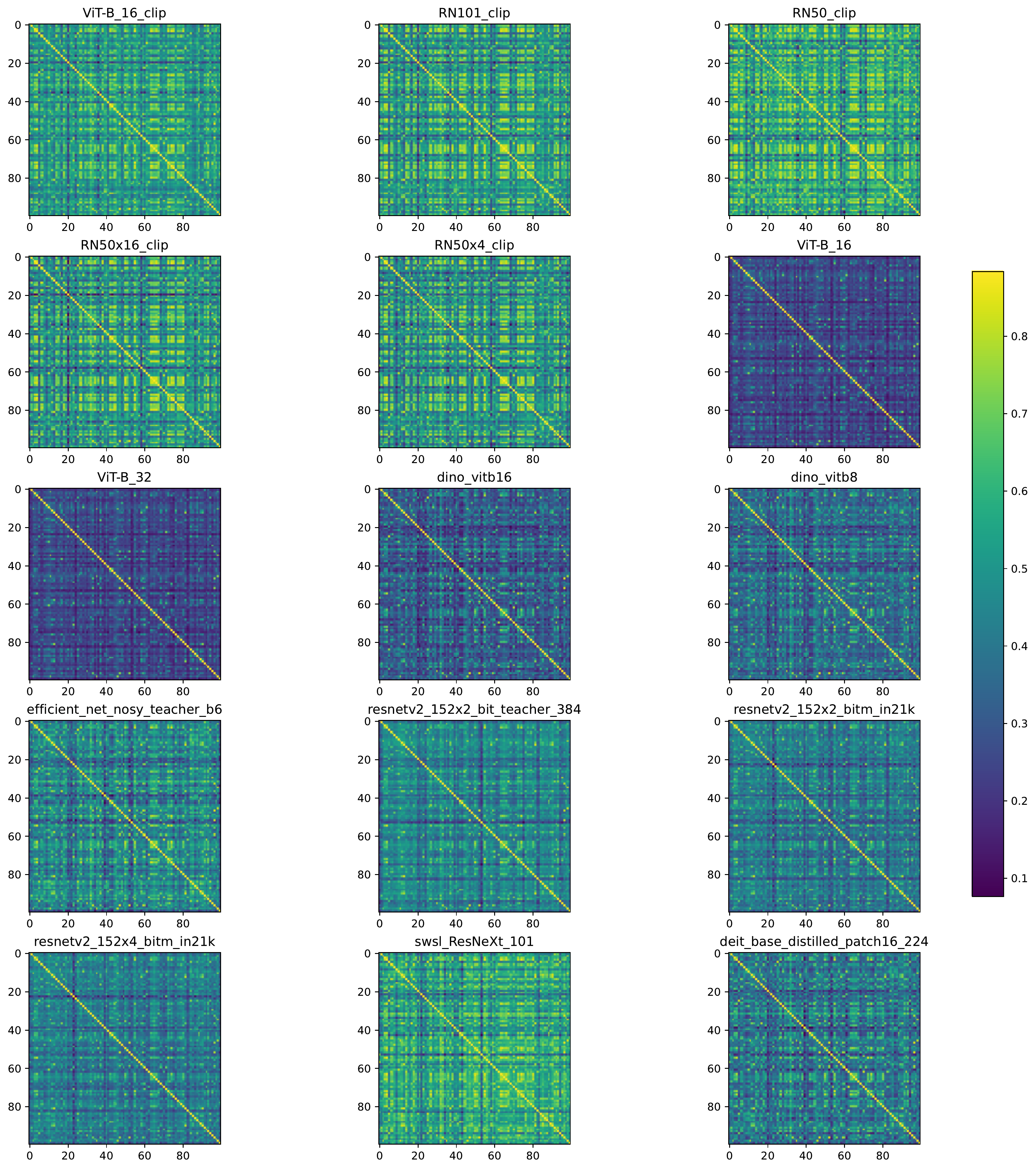}
    \caption{Classes Similarity CIFAR100}
    \end{subfigure}
    \begin{subfigure}[b]{0.45\linewidth}
        \centering
        \includegraphics[width=\linewidth]{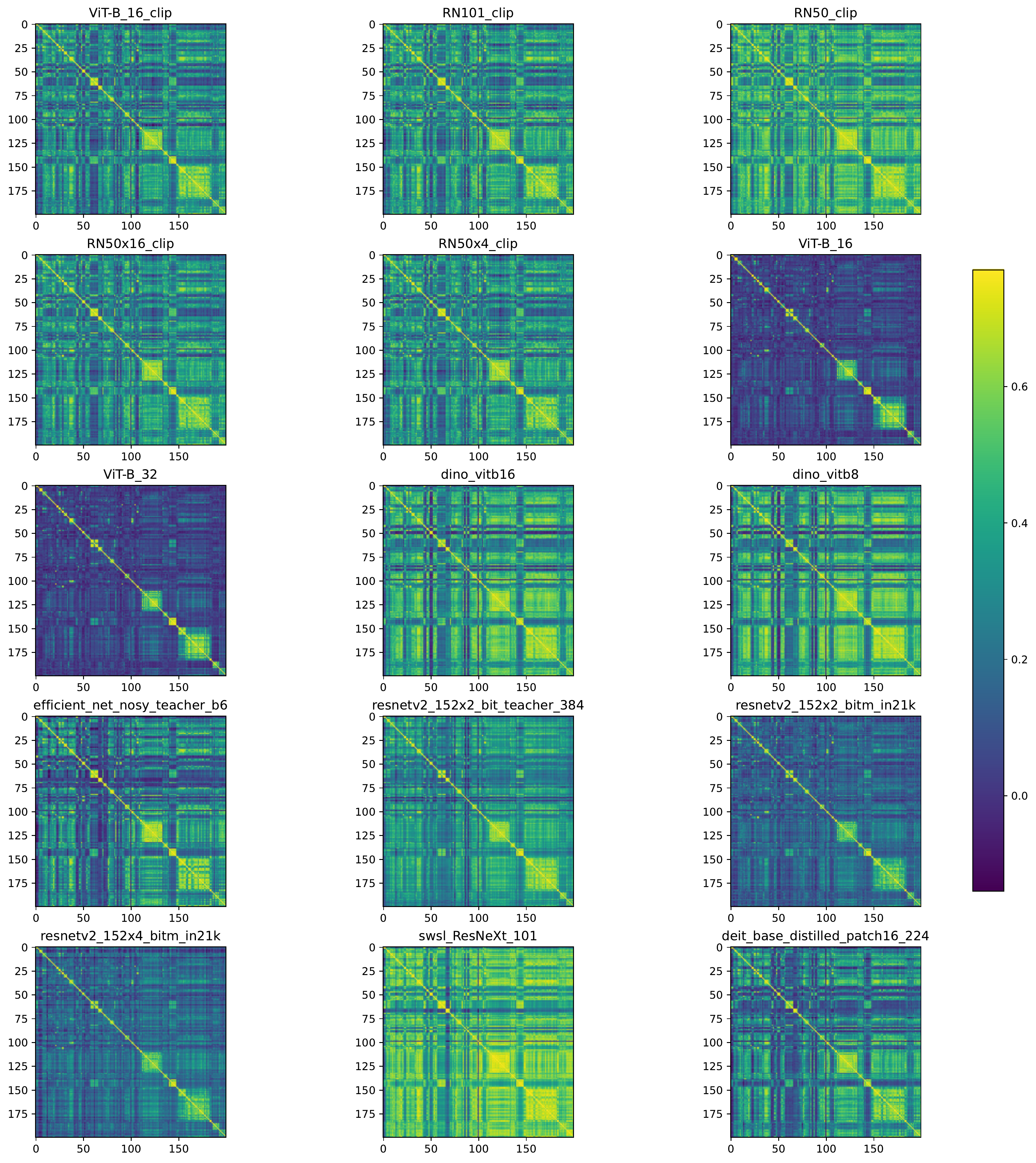}
    \caption{Classes Similarity CUB200}
    \end{subfigure}
    
    \begin{subfigure}[b]{0.45\linewidth}
        \centering
        \includegraphics[width=\linewidth]{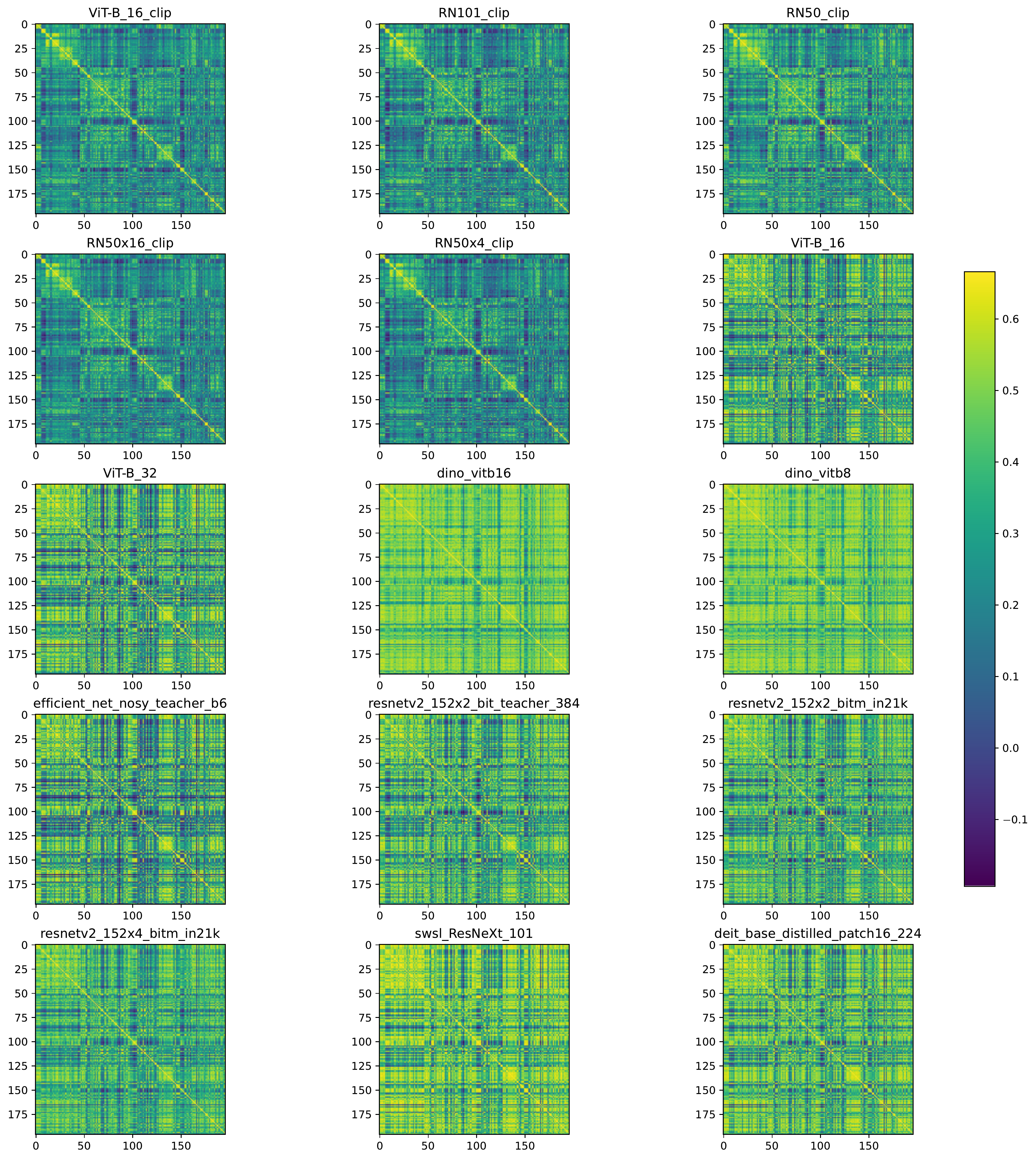}
    \caption{Classes Similarity Cars196}
    \end{subfigure}
    \begin{subfigure}[b]{0.45\linewidth}
        \centering
        \includegraphics[width=\linewidth]{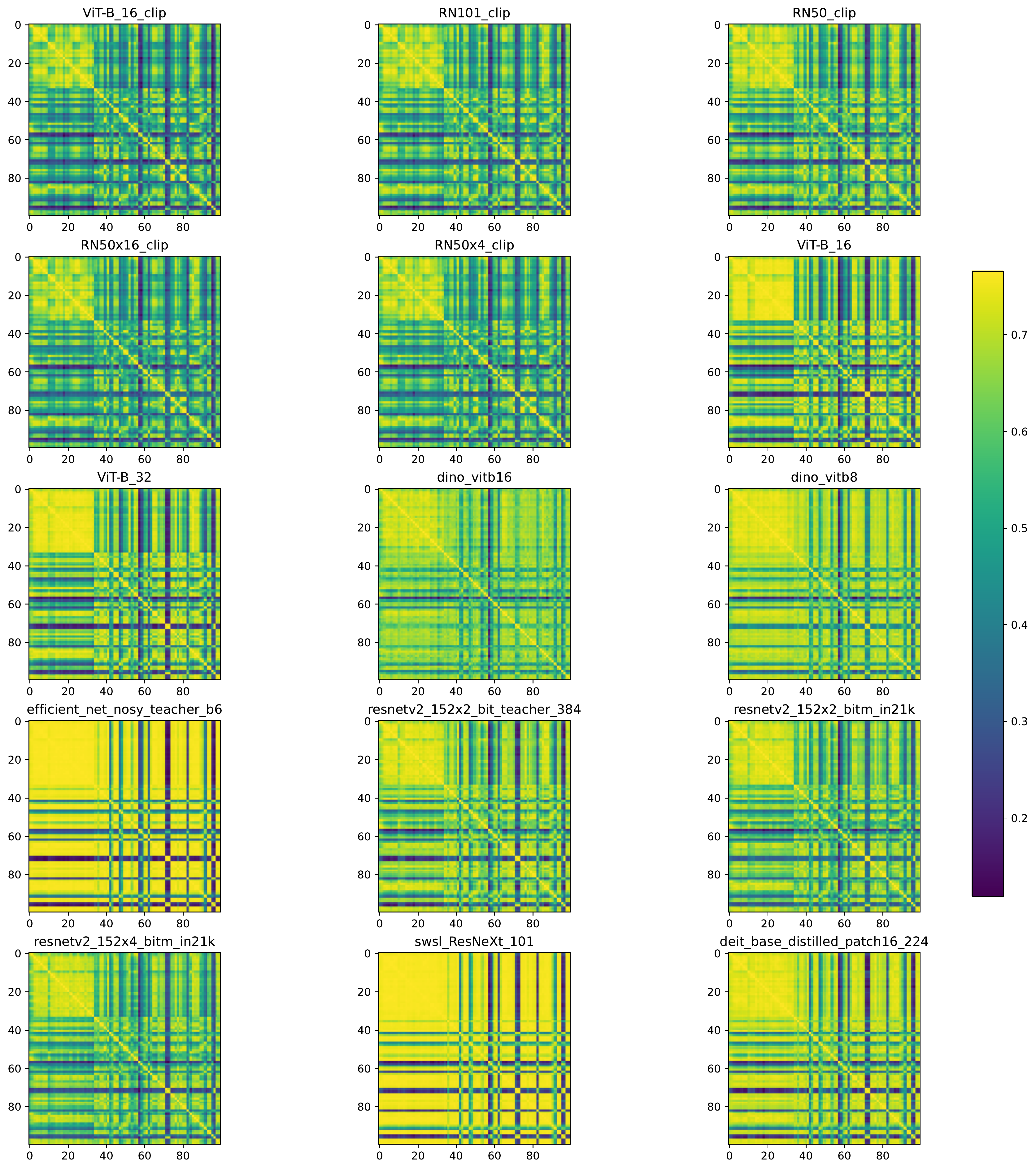}
    \caption{Classes Similarity FGVCAircraft}
    \end{subfigure}

    \caption{\textbf{Classes similarities with various encoders and datasets.} Matrices represent cosine similarities between classes. The cosine similarity is computed by computing the mean of features (the prototype) for a given model and computing the cosine similarity between prototypes. By comparing all prototypes we get the matrices here. Blue color represents low cosine similarity, i.e. vectors are more orthogonal, yellow color represents high cosine similarity, i.e. vectors are less orthogonal. In our experiments, the brighter the matrices is the more OOD the dataset is from the pretraining data. NB: the more orthogonal the vector the easier the classification task is.}
    \label{ap:fig:prototype_similarity}
\end{figure}

\clearpage
\section{Training details}
\label{ap:pretrained}

\subsection{Hyperparameters}
We perform hyperparameter selection (learning rate, weight decay, and whether to anneal the learning rate) using the validation accuracy. On single dataset splits (e.g. CIFAR100/5) we run hyperparameter selection only on the first task and keep the hyperparameters for the rest of the stream. Unless stated otherwise, we use 5 different randomly selected but fixed task orderings for all experiments. On the multi-dataset stream we perform per-task hyperparameter search. %
Unless otherwise stated, we average all results over 5 different task orders to factor out results from a particular task order.

\subsection{Models}
\label{sec:models}
Among the models used 11 are based on transformers~\citep{dosovitskiy2020image,touvron2021training} and 19 on ResNets~\citep{He_2020_CVPR, zagoruyko2016wide,kolesnikov2020big, tan2019efficientnet}. We use 13 models pre-trained on ImageNet 1k~\citep{deng2009imagenet}, 9 on ImageNet 21k~\citep{kolesnikov2020big}, 6 on CLIP~\citep{radford2021learning}, 1 on JFT~\citep{sun2017revisiting} + Imagenet 1k~\citep{tan2019efficientnet}, and 1 with 940M of unlabeled images~\citep{yalniz2019billion}. From these models, 7 are trained with self-supervision~\citep{caron2021emerging}, 7 are big-transfer models~\citep{kolesnikov2020big}, and 1 is trained via semi-weakly supervised learning~\citep{yalniz2019billion}.
Table~\ref{tab:models} contains the specifications of the models used for the empirical study.

\begin{table*}[h]
\caption{Pre-trained model details. \textit{Encoder} indicates the architecture in \texttt{timm} format, \textit{Linear} indicates whether results were obtained by fine-tuning the model (NO) or by training a linear layer on top of frozen features (YES). \textit{Pretraining data} contains the dataset used for pre-training. \textit{\#Examples} shows the size of the pre-training datasets. \textit{Supervision} indicates the type of supervision to train the encoder. \textit{Dim} is the output latent dimensionality. \textit{\#Params} is the number of parameters of the pre-trained model. \textit{Acc} is the ImageNet top-1 accuracy of the pre-trained model.}
\label{tab:models}
\setlength{\tabcolsep}{12pt}  
\resizebox{\linewidth}{!}{\begin{tabular}{@{}lccccccc@{}}
\toprule
Encoder & Linear & Pretraining data & \#Examples & Supervision & Dim & \#Params & Acc \\ \midrule
RN101\_clip                         & YES & clip dataset           & 400.0M & text                     & 512  & 119.7M & 75.70 \\
RN50\_clip                          & YES & clip dataset           & 400.0M & text                     & 1024 & 102.0M & 73.30 \\
RN50x16\_clip                       & YES & clip dataset           & 400.0M & text                     & 768  & 291.0M & 81.60 \\
RN50x4\_clip                        & YES & clip dataset           & 400.0M & text                     & 640  & 178.3M & 78.20 \\
ViT-L/14\_clip                      & YES & clip dataset           & 400.0M & text                     & 768  & 427M   & 83.90 \\
RN50x64\_clip                       & YES & clip dataset           & 400.0M & text                     & 623  & 623M   & 83.60 \\
ViT-B/16\_clip                      & YES & clip dataset           & 400.0M & text                     & 512  & 149.6M & 80.20 \\
ViT-B/32\_clip                      & YES & clip dataset           & 400.0M & text                     & 512  & 151.3M & 76.10 \\
dino\_resnet50                      & YES & Imagenet1K             & 1.3M   & dino                     & 2048 & 23.5M  & 75.30 \\
dino\_vitb16                        & YES & Imagenet1K             & 1.3M   & dino                     & 768  & 85.8M  & 78.20 \\
dino\_vitb16                        & NO  & Imagenet1K             & 1.3M   & dino                     & 768  & 85.8M  & 81.50 \\
dino\_vitb8                         & NO  & Imagenet1K             & 1.3M   & dino                     & 768  & 85.8M  & 82.80 \\
dino\_vitb8                         & YES & Imagenet1K             & 1.3M   & dino                     & 768  & 85.8M  & 80.10 \\
dino\_vits16                        & YES & Imagenet1K             & 1.3M   & dino                     & 384  & 21.7M  & 77.00 \\
dino\_vits8                         & YES & Imagenet1K             & 1.3M   & dino                     & 384  & 21.7M  & 79.70 \\
ViT-B/16                            & NO  & Imagenet21k            & 14.2M  & labels                   & 768  & 85.8M  & 84.53 \\
ViT-B/32                            & NO  & Imagenet21k            & 14.2M  & labels                   & 768  & 87.5M  & 80.72 \\
resnet101                           & NO  & Imagenet1K             & 1.3M   & labels                   & 2048 & 42.5M  & 81.93 \\
resnet152                           & NO  & Imagenet1K             & 1.3M   & labels                   & 2048 & 58.1M  & 78.66 \\
resnet50                            & NO  & Imagenet1K             & 1.3M   & labels                   & 2048 & 23.5M  & 80.37 \\
resnetv2\_50x1\_bitm\_in21k         & NO  & Imagenet21k            & 14.2M  & labels                   & 2048 & 23.5M  & 80.34 \\
resnetv2\_101x1\_bitm\_in21k        & NO  & Imagenet21k            & 14.2M  & labels                   & 2048 & 42.5M  & 82.33 \\
resnetv2\_101x3\_bitm\_in21k        & NO  & Imagenet21k            & 14.2M  & labels                   & 6144 & 381.8M & 84.44 \\
resnetv2\_152x2\_bit\_teacher\_384  & NO  & Imagenet21k            & 14.2M  & labels                   & 4096 & 232.2M & 83.84 \\
resnetv2\_152x2\_bitm\_in21k        & NO  & Imagenet21k            & 14.2M  & labels                   & 4096 & 232.2M & 84.51 \\
resnetv2\_152x4\_bitm\_in21k        & NO  & Imagenet21k            & 14.2M  & labels                   & 8192 & 928.3M & 84.92 \\
resnetv2\_50x1\_bit\_distilled      & NO  & Imagenet21k            & 14.2M  & labels                   & 2048 & 23.5M  & 82.83 \\
resnetv2\_50x1\_bitm                & NO  & Imagenet1K             & 1.3M   & labels                   & 2048 & 23.5M  & 80.34 \\
swsl\_resnext101\_32x16d            & NO  & Imagenet1K + Unlabeled & 1.3M   & labels + semi-supervised & 2048 & 192.0M & 83.36 \\
tf\_efficientnet\_l2\_ns\_475       & NO  & Imagenet1K + JFT300    & 1.3M   & labels + semi-supervised & 5504 & 474.8M & 88.23 \\
deit\_base\_distilled\_patch16\_224 & NO  & Imagenet1K             & 1.3M   & labels + distillation    & 768  & 85.8M  & 83.39 \\
efficient\_net\_nosy\_teacher\_b6   & NO  & Imagenet1K + JFT300    & 1.3M   & labels + distillation    & 2304 & 40.7M  & 86.45 \\
\bottomrule
\end{tabular}}
\end{table*}

\clearpage
\newpage
\section{Additional Figures}
\label{ap:additional_figures}
In this section we include additional figures that are referenced in the main text.

\begin{figure}[hbt!]
    \centering
    \includegraphics[width=0.7\linewidth]{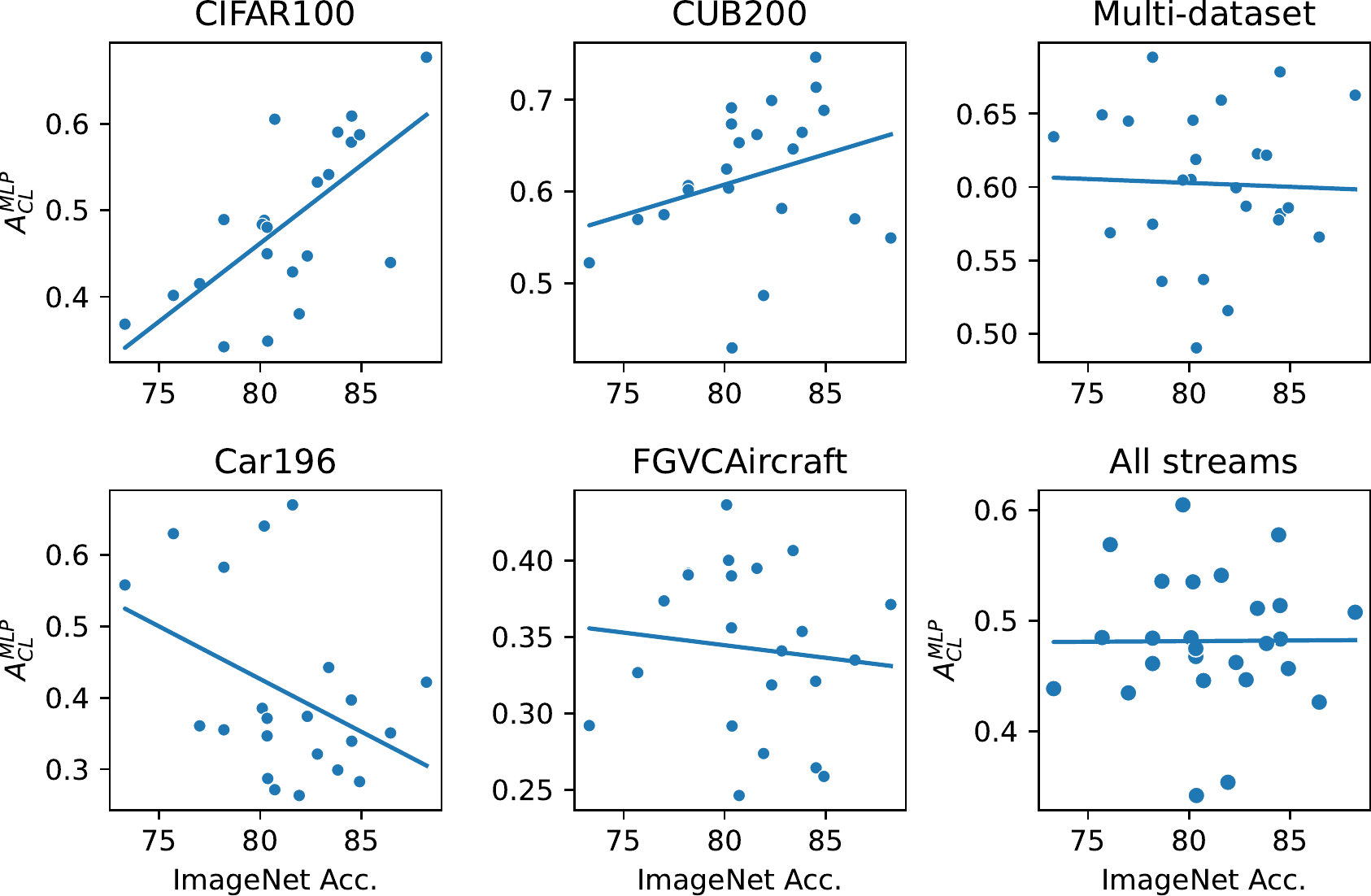}
    \caption{ImageNet accuracy of the encoder (x-axis) and downstream accuracy CL (y-axis). We find positive correlation of the ImageNet accuracy with the downstream CL accuracy for the both in-distribution datasets and slightly negative correlation for the OOD datasets. Overall, however, there is no clear correlation between the two variables. Each point corresponds to a particular, with its CL accuracy averaged over replay buffer sizes (5) and task order permutations(5).}
    \label{ap:fig:imagenet_acc_vs_cl}
\end{figure}
\begin{figure}[hbt!]
    \centering
    \includegraphics[width=0.5\linewidth]{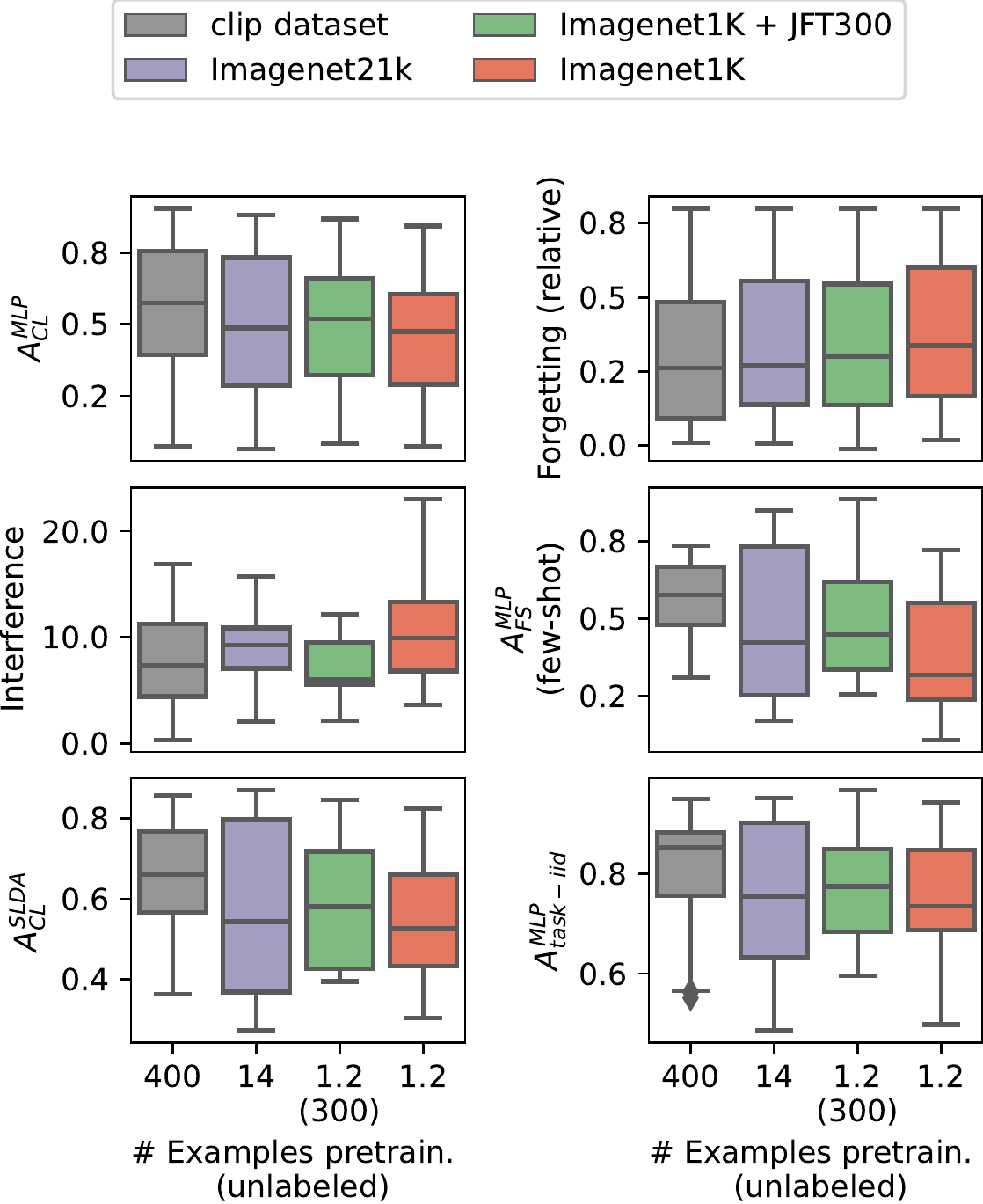}
    \caption{ \Message{Show the influence of pre-training data regime on the different variables studied.} Acc. CL ($\uparrow$), relative forgetting ($\uparrow$), Interference ($\downarrow$), Acc. few-shot, Acc. SLDA and mean task Acc. ($\uparrow$)  of models pretrained on different datasets (each boxplot is over 6 replay buffer sizes, 4 streams and 5 task order permutations).}
    \label{ap:fig:pre-training_dataset_acc_forgetting}
\end{figure}
\begin{figure}[hbt!]
\begin{minipage}[b]{0.48\textwidth}
\centering
\includegraphics[width=1\linewidth]{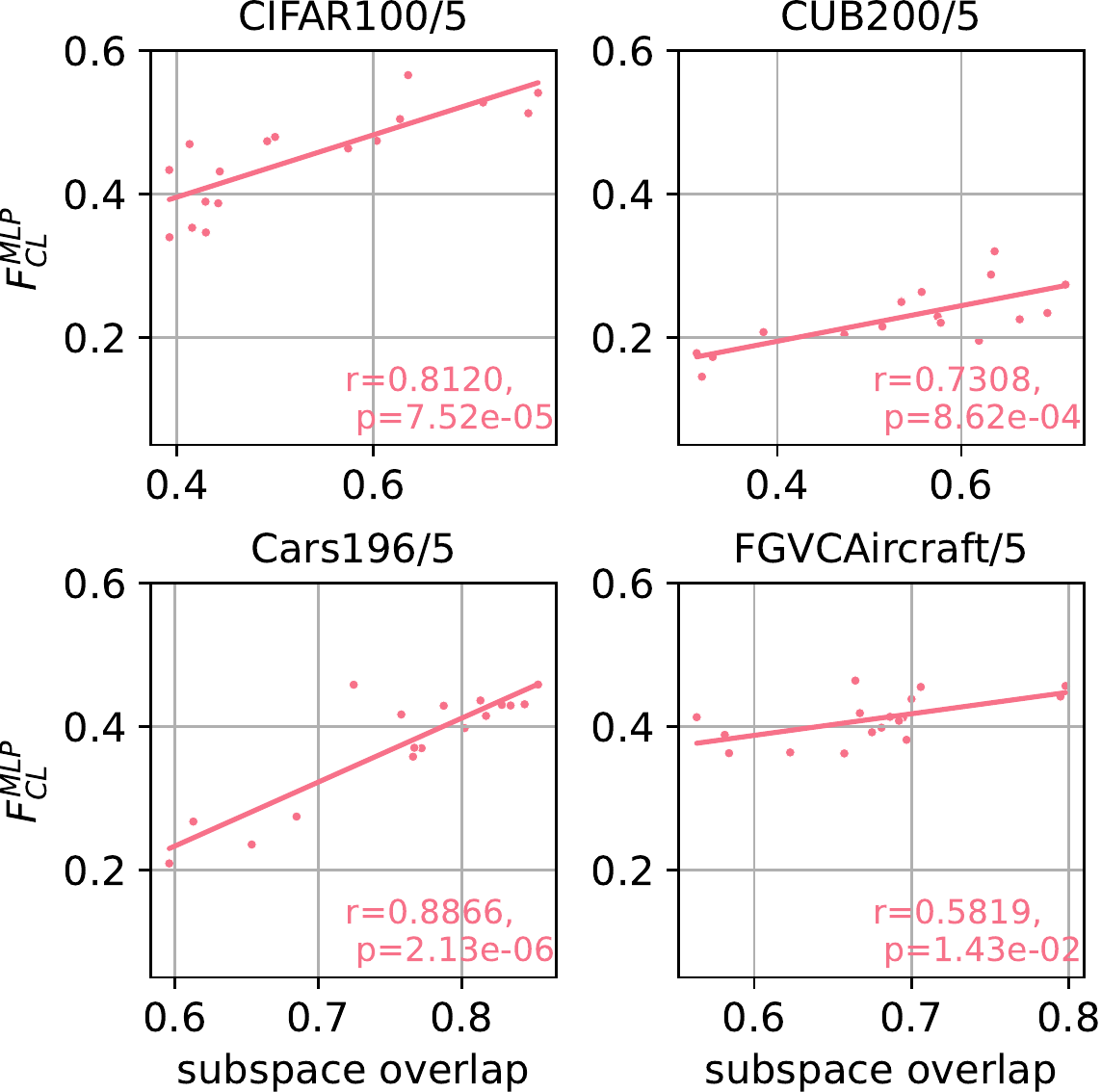}
\caption{\textbf{Average Subspace Overlap} (Similarity) vs \textbf{CL  Forgetting} for multiple datasets where each point represents a pre-trained model averaged over replay sizes and task orders. \Message{Overlap correlates negatively with the CL accuracy.}}
\label{fig:subspace_sim_vs_F_acc}
\end{minipage}
\hfill
\begin{minipage}[b]{0.48\textwidth} 
    \includegraphics[width=1\linewidth]{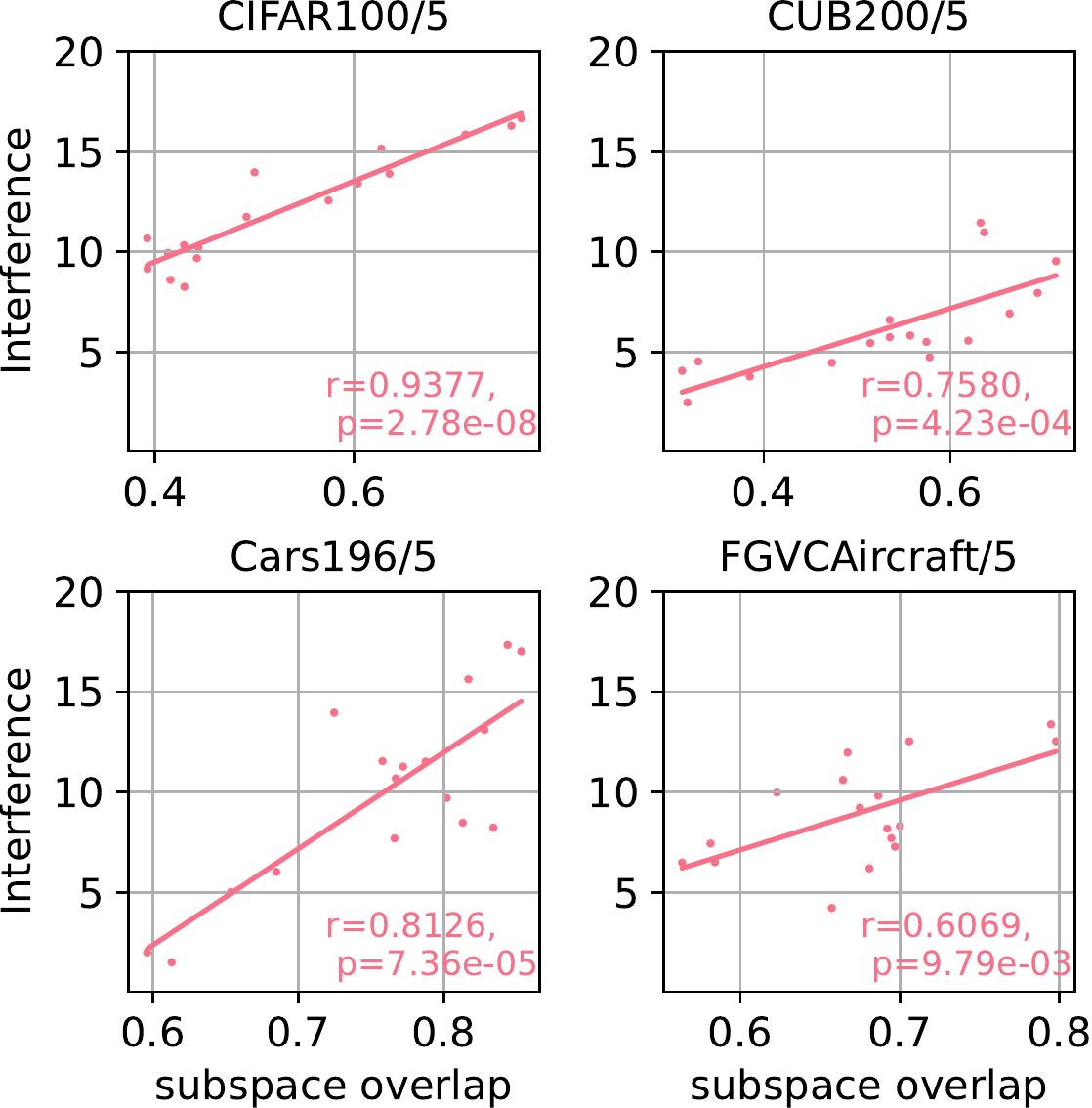}  
    \caption{\textbf{Average Subspace Overlap} (Similarity) \textbf{vs CL  interference} for multiple data-sets where each point represents a pre-trained model  averaged over replay sizes and task orders. \Message{Overlap correlates positively with the CL accuracy.}}
    \label{fig:subspace_sim_vc_interference}
\end{minipage}

\begin{minipage}[b]{0.48\textwidth}
\centering
\includegraphics[width=1\linewidth]{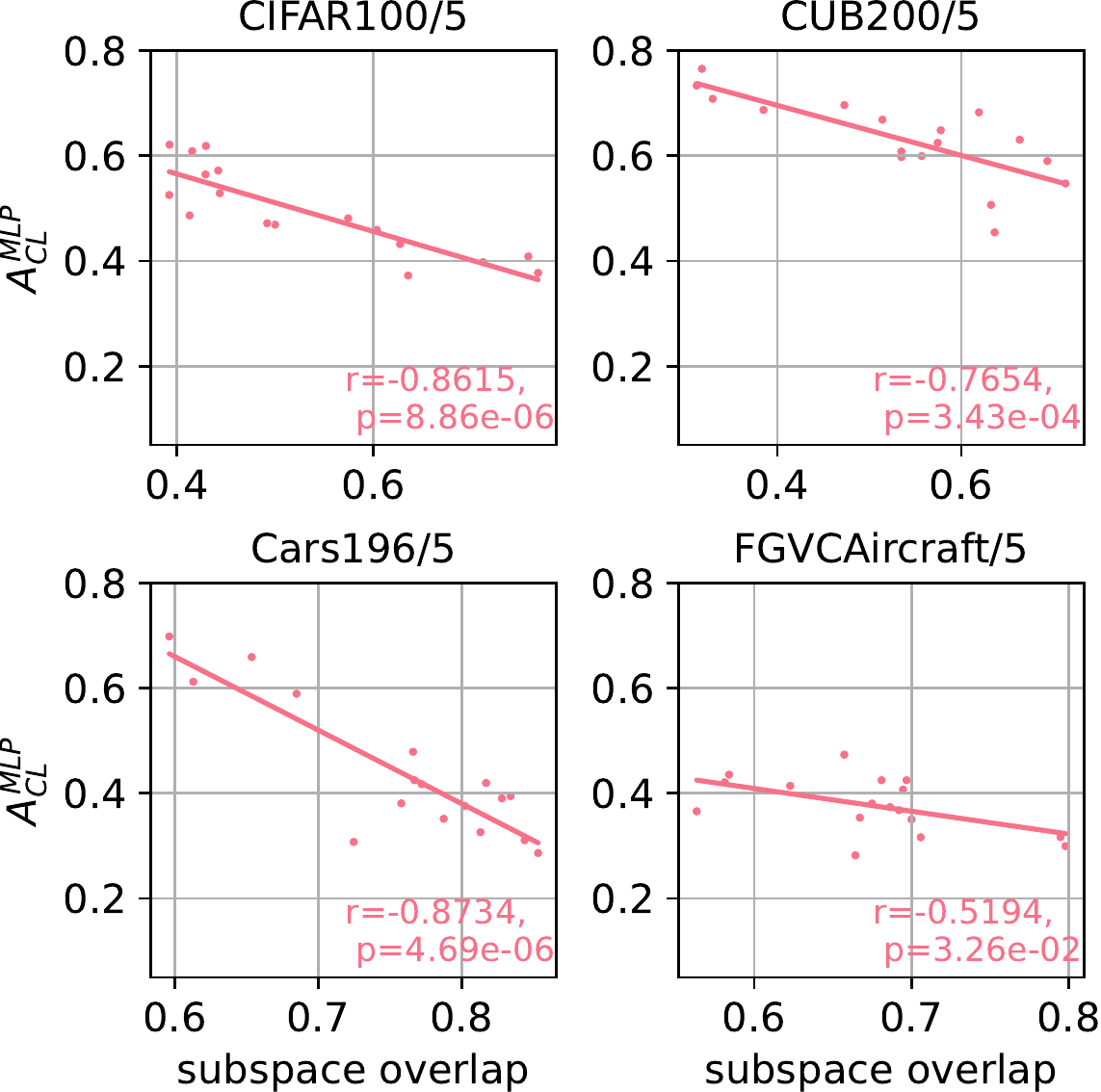}
\caption{\textbf{Average Subspace Overlap} (Similarity) vs \textbf{CL  Forgetting} for multiple datasets where each point represents a pre-trained model  averaged over replay sizes and task orders. \Message{Overlap correlates negatively with the CL accuracy.}}
\label{ap:fig:subspace_sim_vs_CL_acc}
\end{minipage}
\end{figure}

\begin{figure}[hbt!]
\centering                             
\includegraphics[width=0.3\linewidth]{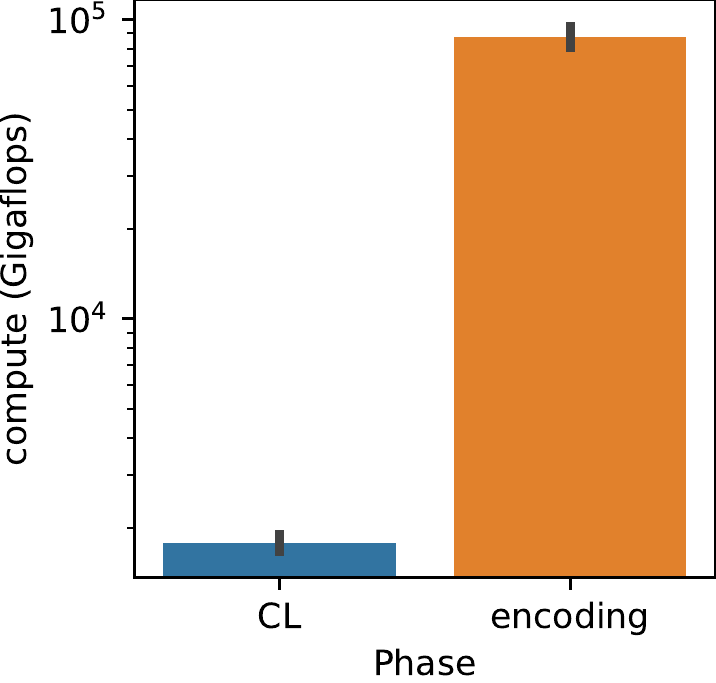}
\caption{Average computational cost per phase. Encoding requires almost 2 orders of magnitude more compute.}
\label{ap:fig:compute_encoding_vs_CL}
\end{figure}

\begin{figure}[hbt!]
\centering    
\begin{subfigure}[b]{1\linewidth}
\includegraphics[width=1\linewidth]{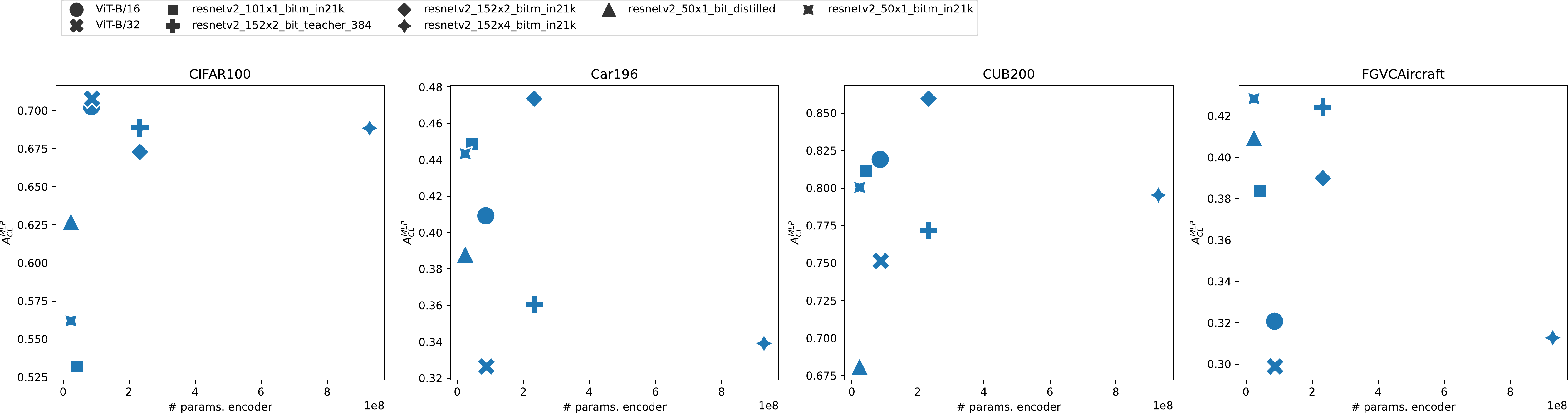}
\caption{ImageNet21k pretrained models}
\end{subfigure} 
\begin{subfigure}[b]{1\linewidth}
\includegraphics[width=1\linewidth]{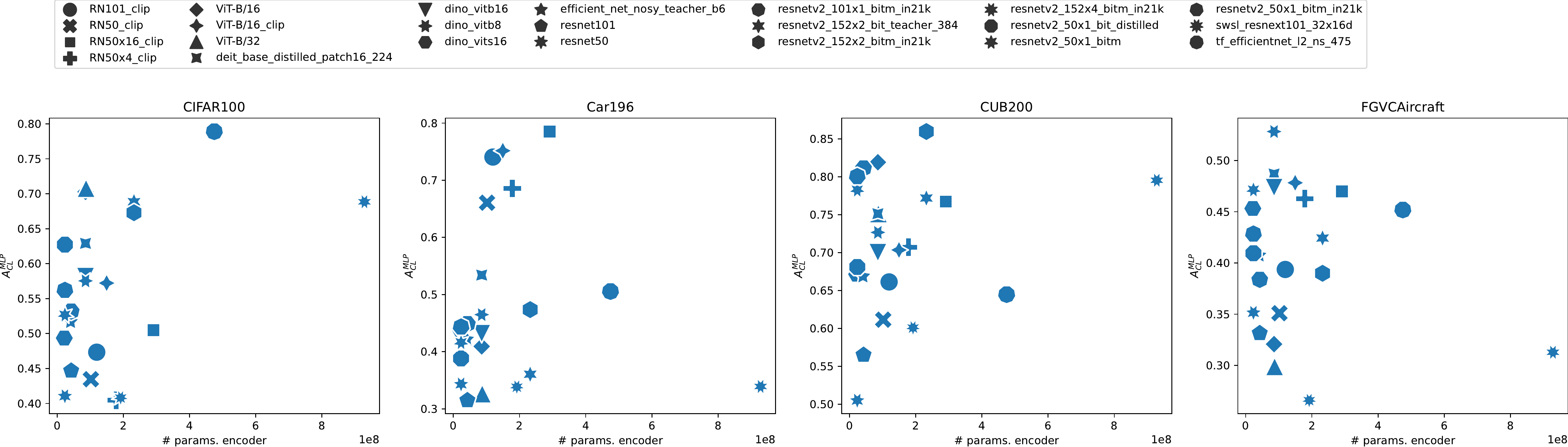}
\caption{All models}
\end{subfigure}
\caption{Number of parameters in the encoder (x-axis) vs. CL accuracy (y-axis). No clear correlation can be observed.}
\label{ap:fig:n_params_vs_accCL}
\end{figure}

\begin{figure}[hbt!]
    \centering
    \includegraphics[width=1\linewidth]{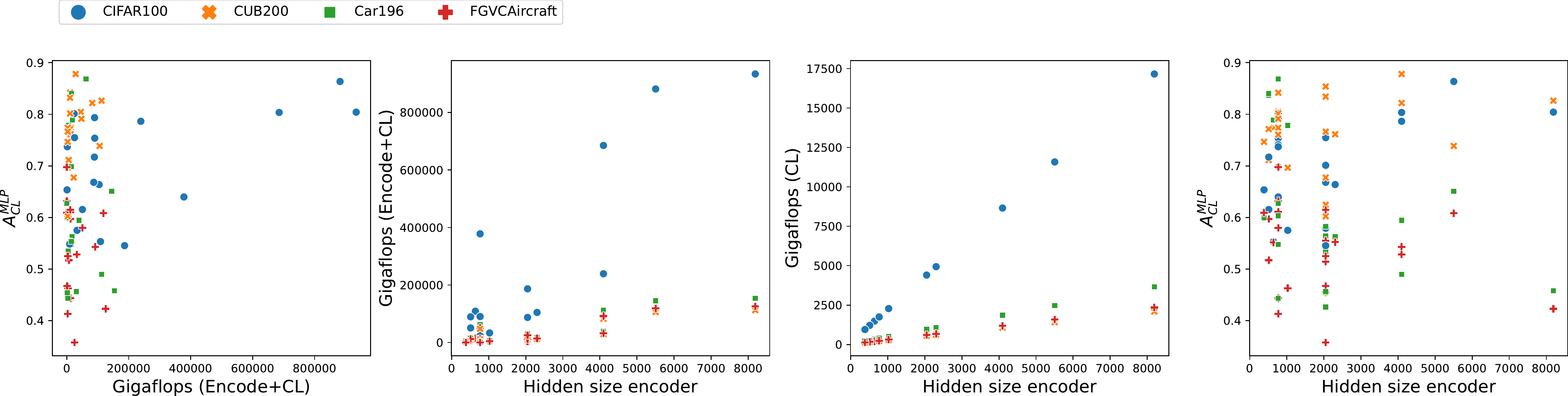}
    \caption{How compute, hidden size and latent dimension of the encoder relate to each other (ER buffer size = 50).}
    \label{ap:fig:compute_cl_hidden_size}
\end{figure}

\begin{figure}[hbt!]
\centering                             
\includegraphics[width=0.3\linewidth]{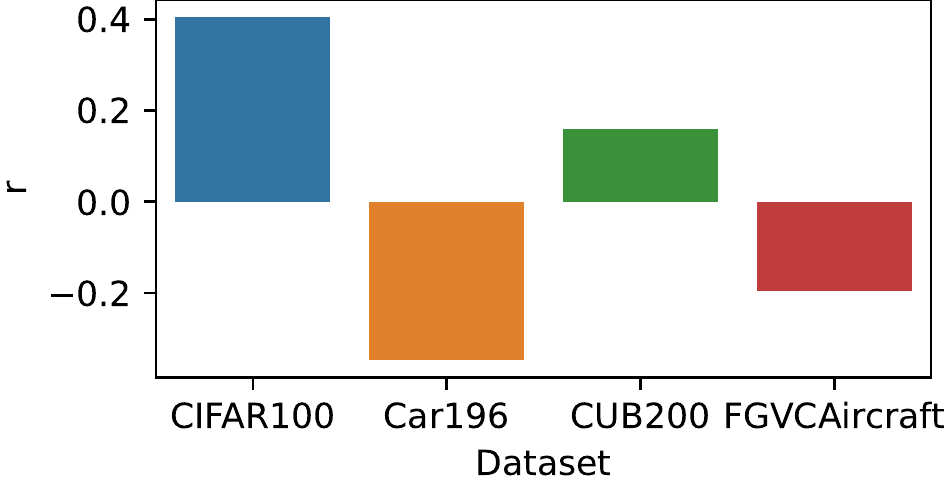}
\caption{ Correlation coefficient between latent size and forgetting (calculated separately for each replay buffer size and averaged). Refer to Fig.~\ref{ap:fig:compute_cl_hidden_size} for details. \Message{Latent dimension does not influence final performance or relative forgetting in our datasets}.}
\label{ap:fig:dimension_influence}
\end{figure}

\begin{figure}[hbt!]
    \centering    
    \includegraphics[width=1\linewidth]{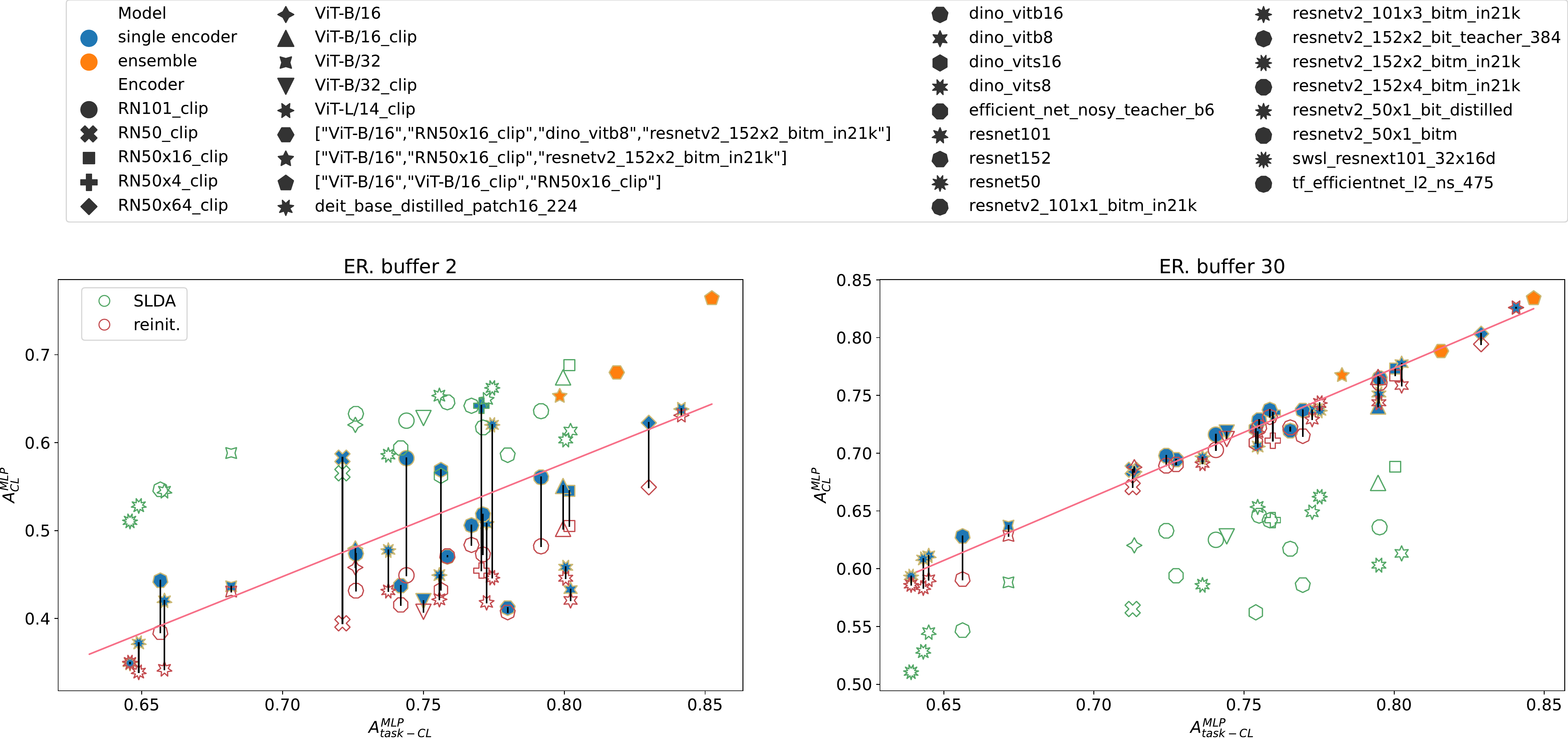}
    \caption{Detailed performance of each mopdel on the multi-dataset stream. Green hollow points correspond to the performance of the SLDA classifier $A_{CL}^{SLDA}$, hollow red -- performance of MLP trained with replay but reinitialized before each task, $A_{CL-reinit}^{MLP}$. Green points correspond to the MLP trained with ER, $A_{CL}^{MLP}$. We also include models with ensembled representations (see legend). Remarkably, the best ensembled model with replay buffer size of 2 (76.4\%) almost reaches the performance of the third best single model with the replay buffer size of 30 (``dino\_vitb8'' with 77.7\%), with the other two best ones being the largest CLIP models ('RN50x64\_clip' -- 80.3\% and ``ViT-L/14\_clip'' with 82.6\%). Due to long run-time of this experiment we only considered a single task ordering here.}
    \label{ap:fig:the_big_one_detailed}
\end{figure}

\begin{figure}[hbt!]
    \centering     
    \begin{subfigure}[b]{1\linewidth}
        \includegraphics[width=1\linewidth]{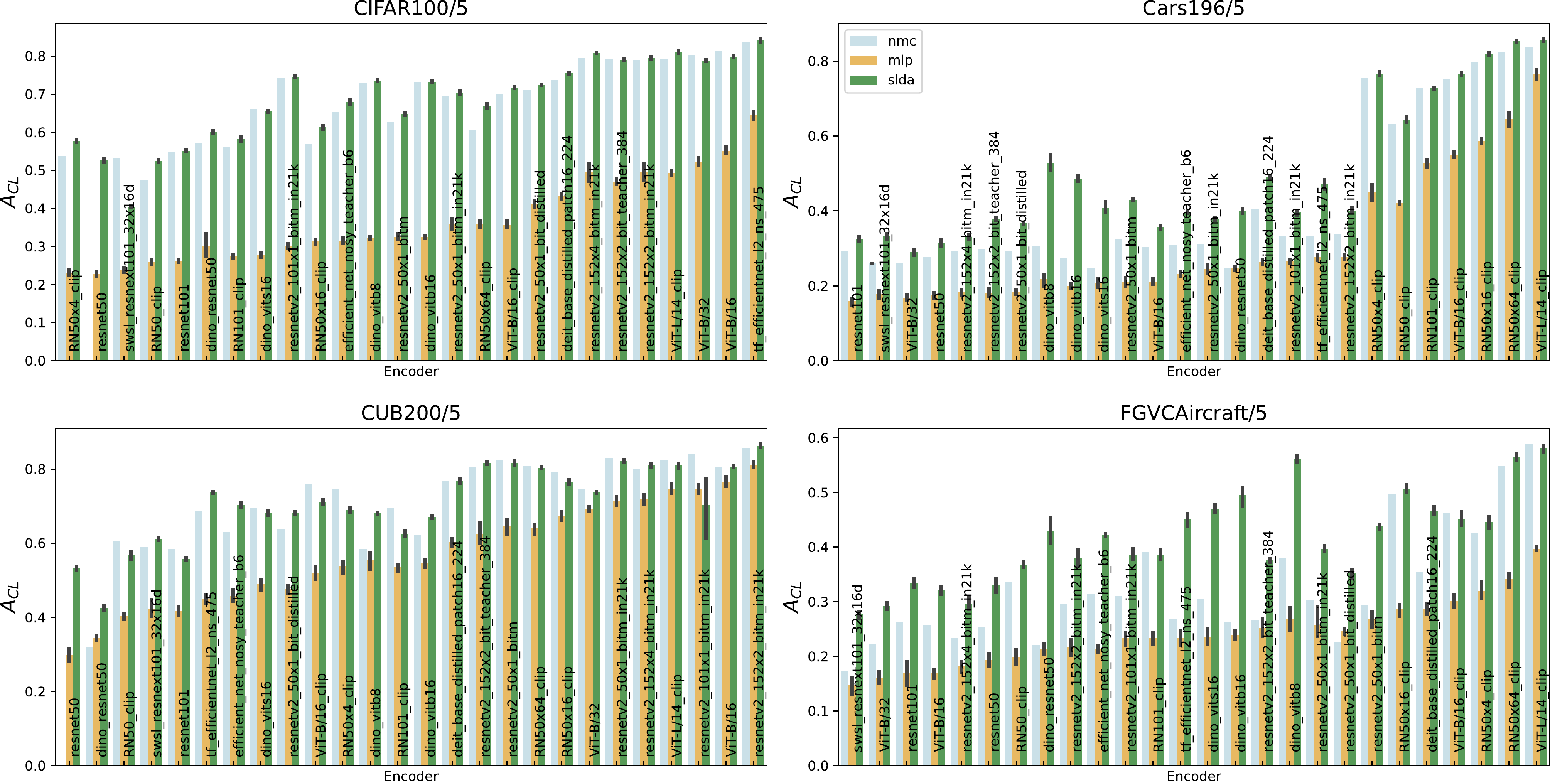}
        \caption{ER buffer 2}
    \end{subfigure}
    \begin{subfigure}[b]{1\linewidth}
        \includegraphics[width=1\linewidth]{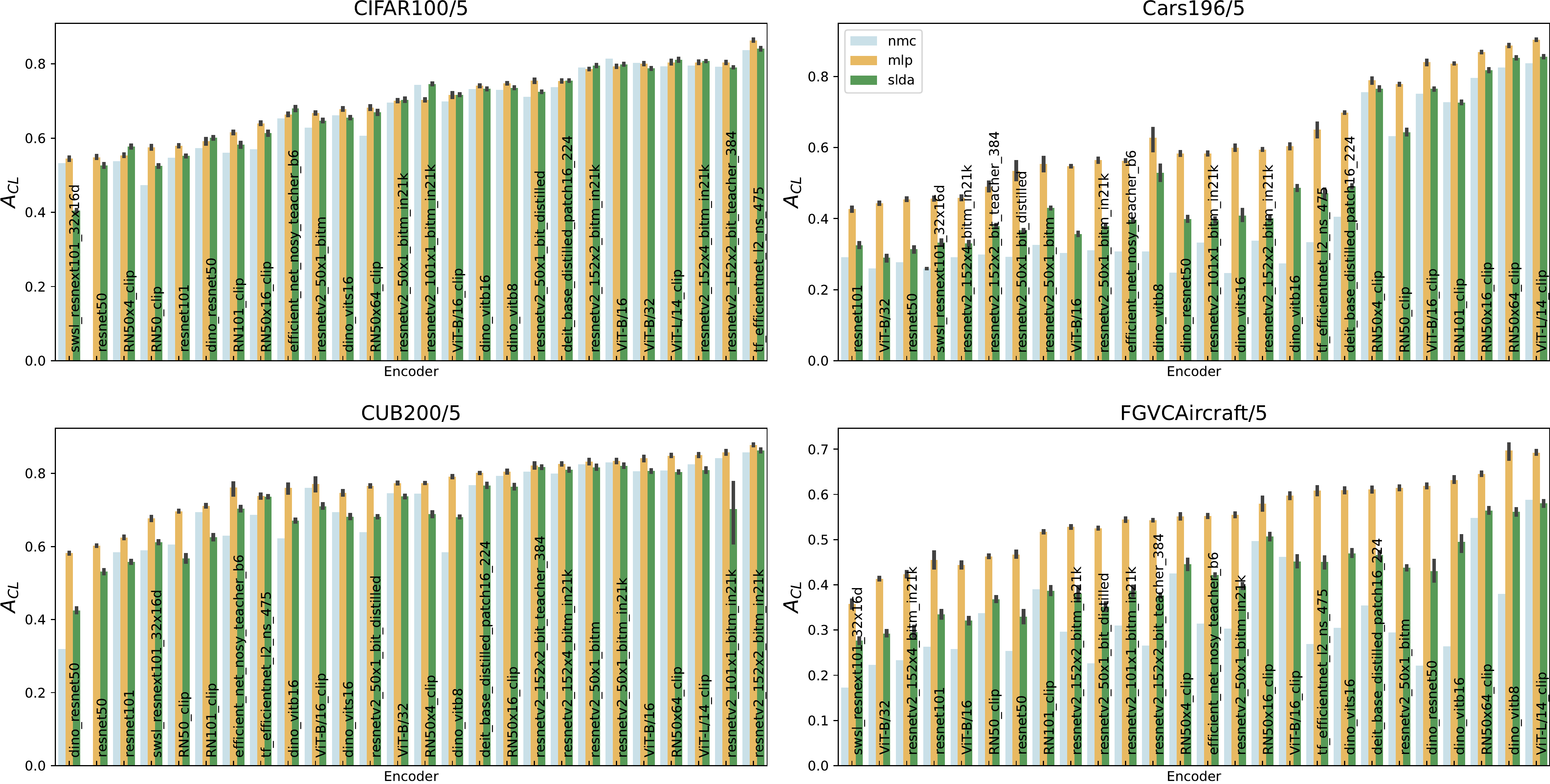}
        \caption{ER buffer 50}
    \end{subfigure}
    \caption{MLP classifier vs. metric based classifiers performance on 4 streams.}   
    \label{ap:fig:mlp_slda_nmc_per_stream_per_model}
\end{figure}

\begin{figure}[hbt!]
    \centering             
    \includegraphics[width=1\linewidth]{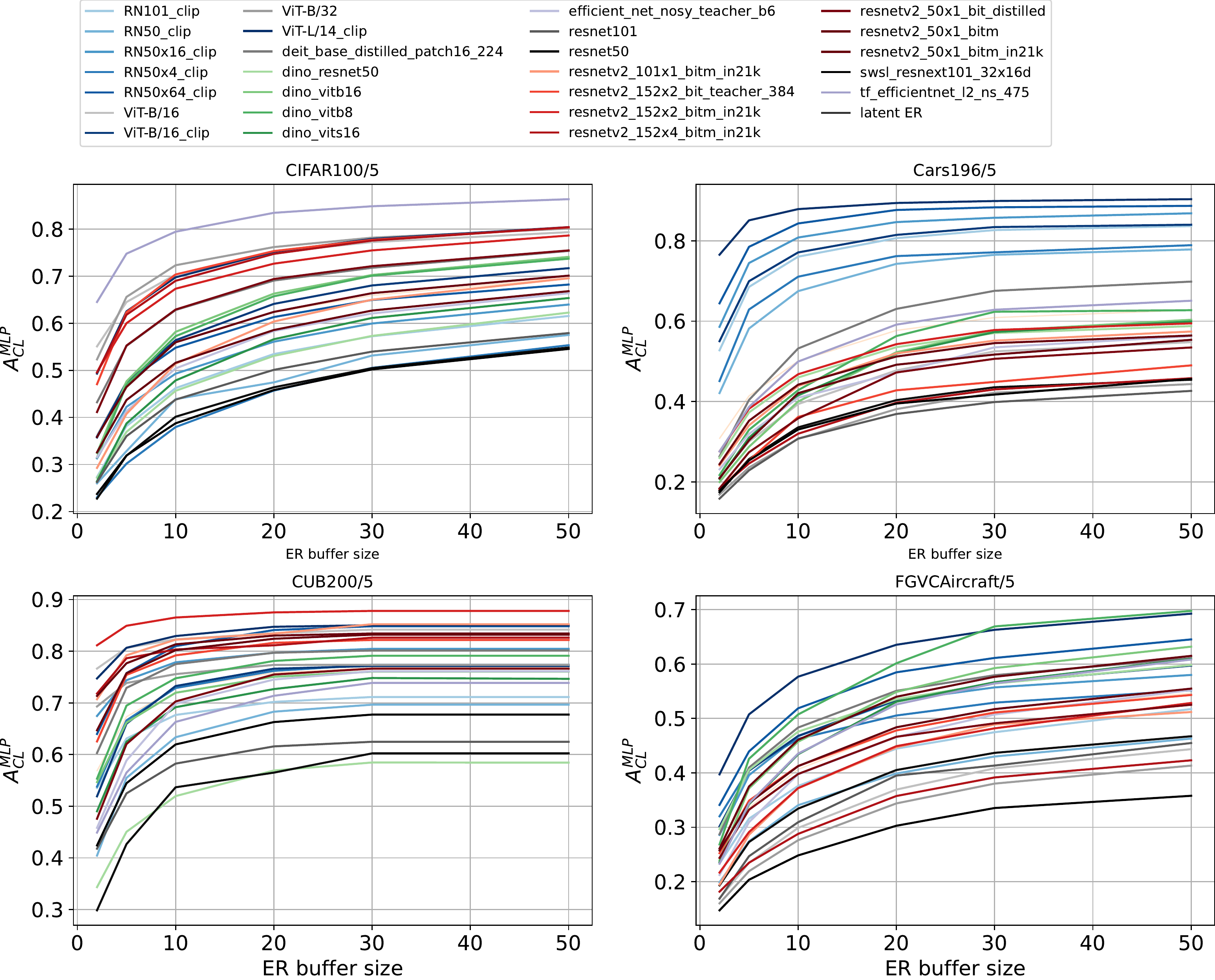}
    \caption{CL performance of various encoders (colors correspond to different encoder families, e.g. blue - CLIP, red -- BiT~\citep{kolesnikov2020big} models) for different replay buffer sizes. \Message{While all encoders' performance improves as more samples are added to the replay buffer, there is no model  universal that dominates all streams.}}
    \label{ap:fig:acc_cl_per_er_size_per_encoder}
\end{figure}

\begin{figure}[hbt!]
    \centering
    \includegraphics[width=1\linewidth]{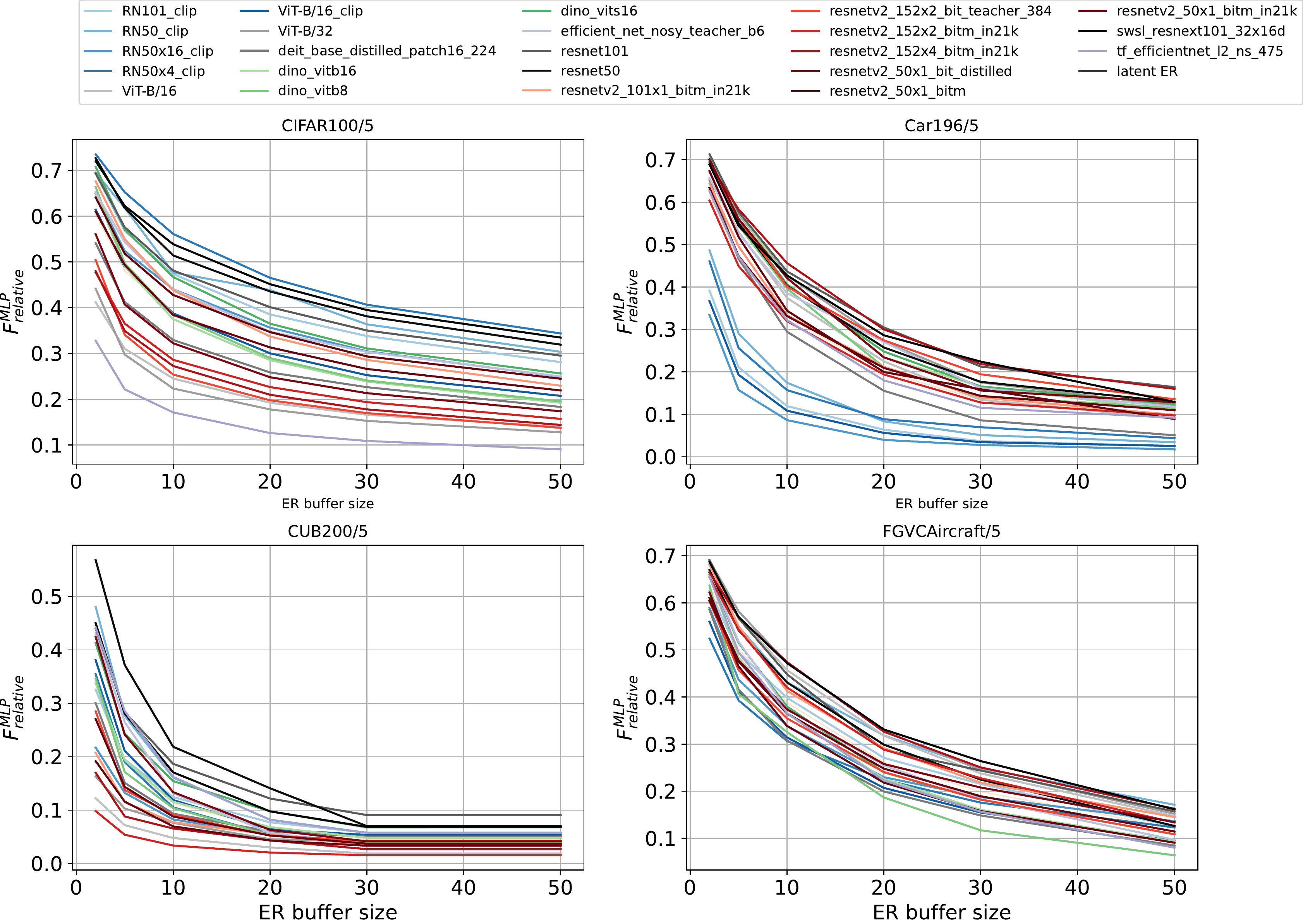}
    \caption{ER buffer efficiency per encoder. Different encoders have different amount of relative forgetting (average amount of accuracy lost after learning the whoel task sequence in \%) across the replay buffer sizes.}
    \label{ap:fig:buffer_efifciency_per_model}
\end{figure}

\begin{figure}[hbt!]
    \centering
    \includegraphics[width=1\linewidth]{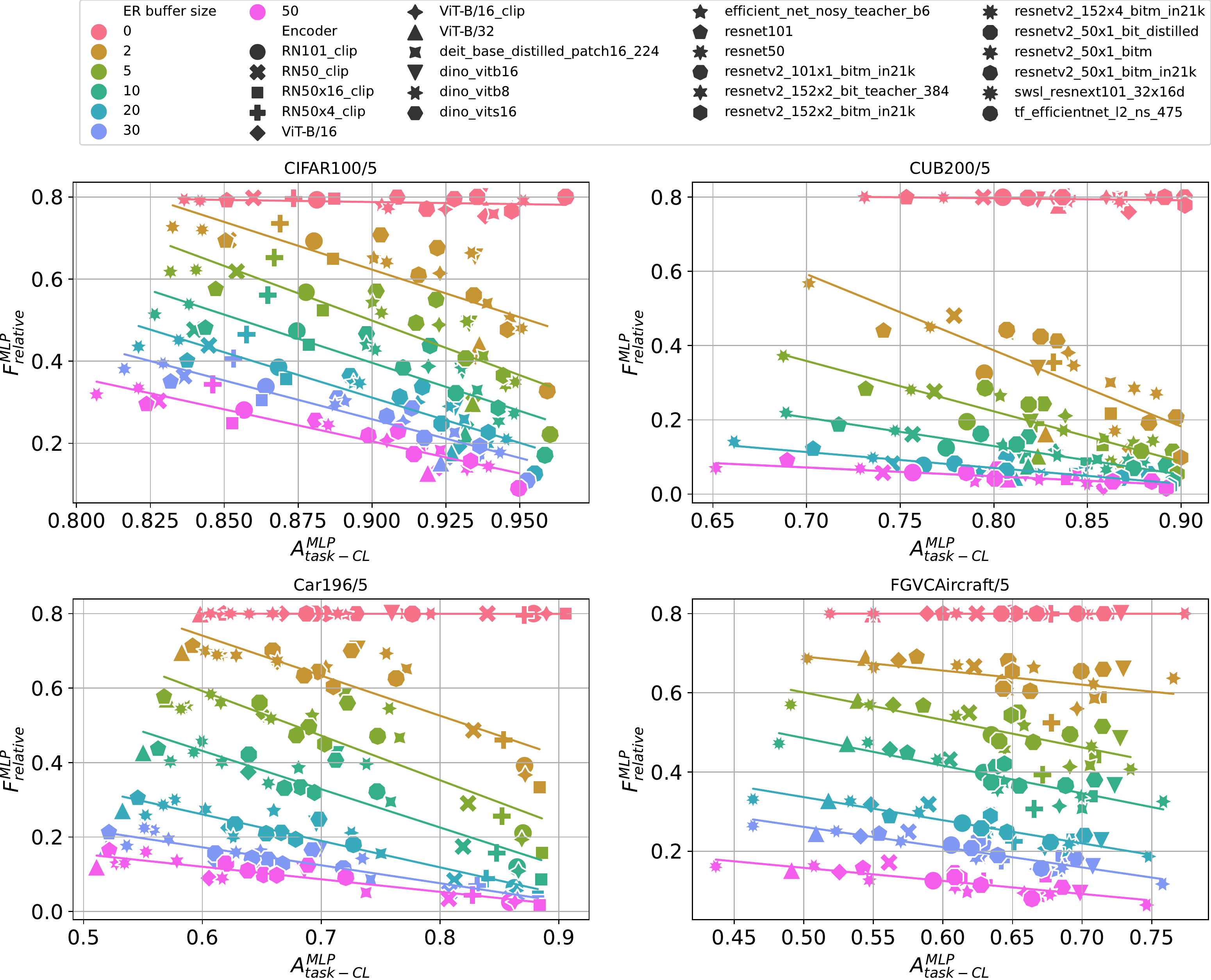}
    \caption{Mean task accuracy (x-axis) vs. relative forgetting (y-axis). Two measurements are well correlated, yet more especially for small replay buffer sizes similar mean task accuracy does not seem to explain forgetting.}
    \label{ap:fig:acc_task_cl_F}
\end{figure}

\begin{figure}[hbt!]
    \begin{subfigure}[b]{0.45\linewidth}
    \centering
    \includegraphics[width=1\linewidth]{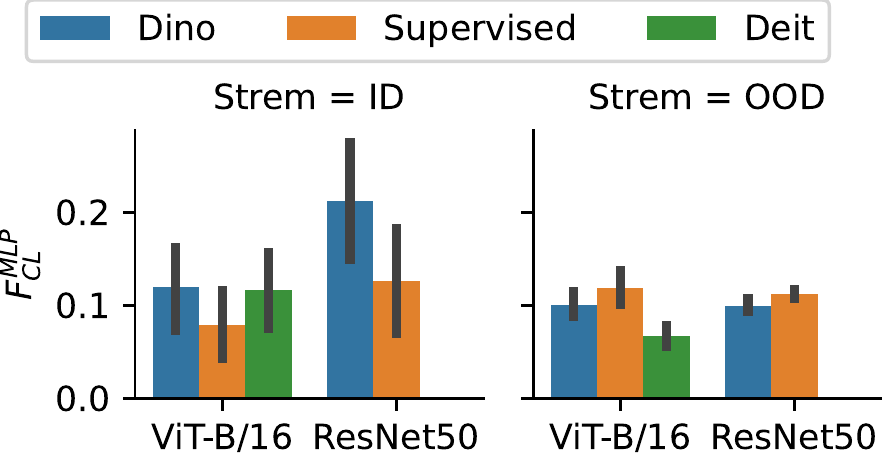}
    \caption{Forgetting}
    \label{ap:fig:supervised_vs_dino_F}
    \end{subfigure}
    \begin{subfigure}[b]{0.45\linewidth}
    \centering
    \includegraphics[width=1\linewidth]{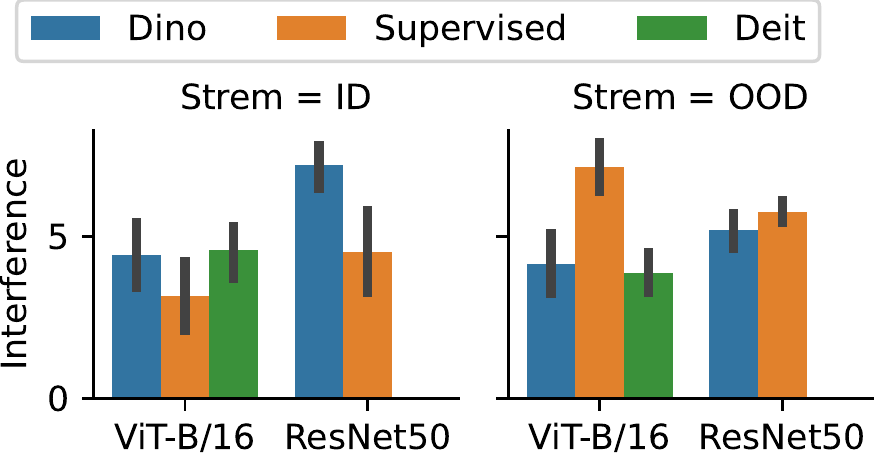}
    \caption{Interference}
    \label{ap:fig:supervised_vs_dino_I}
    \end{subfigure}
    \caption{Forgetting (a) and  interference (b) after supervised (ImageNet21K) vs. SSL (dino by \cite{caron2021emerging}, ImageNet1K) and upervised+ditillation (deit by \cite{touvron2021training} based pre-training. Here we compare dino\_vitb16, ViT-B/16 and deit\_base\_distilled\_patch16\_224, as well as dino\_resnet50 and resnetv2\_50x1\_bitm\_in21k models with ER buffer 50.}
    \label{ap:fig:sup_vs_elf_sup_pretraining_forgetting_interference}
\end{figure}

\section{Limitations}
In the following we list the limitations of this study. 

First, this study only considers one domain, namely the vision domain. We believe similar results are likely to apply in other domains such as Natural Language Processing (NLP). Due to the specifics of the NLP domain, a rigorous evaluation of modern Language Models (LM) would deserve a separate paper on its own and would exceed the scope of this study. For example, one interesting, and NLP-specific question, could be whether and when fine-tuning of LMs is necessary at all, and if simple prompting could lead to good performance on downstream tasks in certain contexts. Additionally, language offers some interesting types of distribution shifts, such as e.g. outdated factual knowledge ~\citep{jang2021towards}, which deserve a separate discussion. Finally, in the context of NLP, a similar study has been conducted recently by ~\citet{wu2022pretrained}, however this study does not consider baselines with completely fixed encoders.
  
Another limitation is that we solely focus on replay strategy and class-incremental scenario. This choice is motivated by the generality of the replay strategy as it can be applicable in most CL scenarios~\citep{Kalifou19,diethe2018continual,lesort2019regularization,prabhu2020gdumb,Traore19DisCoRL,Belouadah2018DeeSIL,wu2019large,Hou_2019_CVPR,lesort2020continual,douillard2020podnet}.

Finally, the variance (for a single experimental setup) in our study comes from different task orderings and not from varying the seed of the classifier.  initialization. Varying the seed would not affect the encoding procedure while it would significantly increase the number of experiments. Additionally, the classifier is usually not reinitialised before each task, hence changing the seed would effect only the classifier initialisation before learning the whole stream. At the same time changing the task ordering actually effects the state of the classifier throughout the stream.

\clearpage
\section{Detailed list of results}
\label{ap:results}
Tab.~\ref{tab:results_2} and Tab.~\ref{tab:results_50} contains a detailed list of results for multiple models and datasets with replay size of 2 and 50, averaged over 5 different task orders.

\begin{table*}[h!]
\centering
\caption{Baseline results on replay size of 2 averaged over 5 tasks order.}
\label{tab:results_2}
\resizebox{0.70\textwidth}{!}{
\begin{tabular}{c|c|cccccccc}
\hline 
Dataset & Encoder &  MLP & reinit & NMC & SLDA & upperbound & FSH & transfer & interference \\ 
\hline

CIFAR100 &
RN101-clip &
$27.36$\mbox{\scriptsize{$\pm0.62$}} & %
$25.38$\mbox{\scriptsize{$\pm0.55$}} & %
$56.04$\mbox{\scriptsize{$\pm0.00$}} & %
$58.20$\mbox{\scriptsize{$\pm0.71$}} & %
$72.98$\mbox{\scriptsize{$\pm0.87$}} & %
$58.26$\mbox{\scriptsize{$\pm0.20$}} & %
$1.98$\mbox{\scriptsize{$\pm0.83$}} & %
$15.15$\mbox{\scriptsize{$\pm0.98$}} \\ %
 \hline 

\rowcolor{Gray} - &
RN50-clip &
$25.99$\mbox{\scriptsize{$\pm0.74$}} & %
$25.24$\mbox{\scriptsize{$\pm1.17$}} & %
$47.30$\mbox{\scriptsize{$\pm0.00$}} & %
$52.54$\mbox{\scriptsize{$\pm0.38$}} & %
$70.13$\mbox{\scriptsize{$\pm0.19$}} & %
$49.06$\mbox{\scriptsize{$\pm0.72$}} & %
$0.75$\mbox{\scriptsize{$\pm1.46$}} & %
$15.85$\mbox{\scriptsize{$\pm0.44$}} \\ %
 \hline 

 - &
RN50x16-clip &
$31.30$\mbox{\scriptsize{$\pm0.76$}} & %
$30.54$\mbox{\scriptsize{$\pm0.89$}} & %
$57.00$\mbox{\scriptsize{$\pm0.00$}} & %
$61.33$\mbox{\scriptsize{$\pm0.61$}} & %
$75.31$\mbox{\scriptsize{$\pm0.14$}} & %
$54.12$\mbox{\scriptsize{$\pm0.29$}} & %
$0.76$\mbox{\scriptsize{$\pm0.70$}} & %
$13.40$\mbox{\scriptsize{$\pm0.21$}} \\ %
 \hline 

\rowcolor{Gray} - &
RN50x4-clip &
$23.01$\mbox{\scriptsize{$\pm0.91$}} & %
$22.44$\mbox{\scriptsize{$\pm0.70$}} & %
$53.75$\mbox{\scriptsize{$\pm0.00$}} & %
$57.76$\mbox{\scriptsize{$\pm0.44$}} & %
$73.45$\mbox{\scriptsize{$\pm0.12$}} & %
$53.17$\mbox{\scriptsize{$\pm0.82$}} & %
$0.57$\mbox{\scriptsize{$\pm0.88$}} & %
$13.90$\mbox{\scriptsize{$\pm0.12$}} \\ %
 \hline 

 - &
RN50x64-clip &
$35.97$\mbox{\scriptsize{$\pm1.14$}} & %
$29.04$\mbox{\scriptsize{$\pm1.45$}} & %
$60.68$\mbox{\scriptsize{$\pm0.00$}} & %
$66.97$\mbox{\scriptsize{$\pm0.51$}} & %
$78.58$\mbox{\scriptsize{$\pm0.09$}} & %
$53.75$\mbox{\scriptsize{$\pm0.60$}} & %
$6.93$\mbox{\scriptsize{$\pm0.64$}} & %
$12.34$\mbox{\scriptsize{$\pm0.17$}} \\
\hline

\hline

\end{tabular}
}
\end{table*}

\begin{table*}[h!]
\centering
\caption{Baseline results on replay size of 50 averaged over 5 tasks order.}
\label{tab:results_50}
\resizebox{0.70\textwidth}{!}{
 \begin{tabular}{c|c|cccccccc}
\hline 
Dataset & Encoder &  MLP & reinit & NMC & SLDA & upperbound & FSH & transfer & interference \\ 
\hline

CIFAR100 &
RN101-clip &
$61.54$\mbox{\scriptsize{$\pm0.42$}} & %
$60.88$\mbox{\scriptsize{$\pm0.55$}} & %
$56.04$\mbox{\scriptsize{$\pm0.00$}} & %
$58.20$\mbox{\scriptsize{$\pm0.71$}} & %
$72.98$\mbox{\scriptsize{$\pm0.87$}} & %
$58.26$\mbox{\scriptsize{$\pm0.20$}} & %
$0.66$\mbox{\scriptsize{$\pm0.33$}} & %
$15.15$\mbox{\scriptsize{$\pm0.98$}} \\ %
 \hline 

\rowcolor{Gray} - &
RN50-clip &
$57.51$\mbox{\scriptsize{$\pm0.56$}} & %
$56.21$\mbox{\scriptsize{$\pm2.73$}} & %
$47.30$\mbox{\scriptsize{$\pm0.00$}} & %
$52.54$\mbox{\scriptsize{$\pm0.38$}} & %
$70.13$\mbox{\scriptsize{$\pm0.19$}} & %
$49.06$\mbox{\scriptsize{$\pm0.72$}} & %
$1.30$\mbox{\scriptsize{$\pm2.69$}} & %
$15.85$\mbox{\scriptsize{$\pm0.44$}} \\ %
 \hline 

 - &
RN50x16-clip &
$63.97$\mbox{\scriptsize{$\pm0.40$}} & %
$66.62$\mbox{\scriptsize{$\pm0.26$}} & %
$57.00$\mbox{\scriptsize{$\pm0.00$}} & %
$61.33$\mbox{\scriptsize{$\pm0.61$}} & %
$75.31$\mbox{\scriptsize{$\pm0.14$}} & %
$54.12$\mbox{\scriptsize{$\pm0.29$}} & %
$-2.64$\mbox{\scriptsize{$\pm0.19$}} & %
$13.40$\mbox{\scriptsize{$\pm0.21$}} \\ %
 \hline 

\rowcolor{Gray} - &
RN50x4-clip &
$55.33$\mbox{\scriptsize{$\pm0.37$}} & %
$56.67$\mbox{\scriptsize{$\pm0.45$}} & %
$53.75$\mbox{\scriptsize{$\pm0.00$}} & %
$57.76$\mbox{\scriptsize{$\pm0.44$}} & %
$73.45$\mbox{\scriptsize{$\pm0.12$}} & %
$53.17$\mbox{\scriptsize{$\pm0.82$}} & %
$-1.34$\mbox{\scriptsize{$\pm0.59$}} & %
$13.90$\mbox{\scriptsize{$\pm0.12$}} \\ %
 \hline 

 - &
RN50x64-clip &
$68.22$\mbox{\scriptsize{$\pm0.52$}} & %
$69.00$\mbox{\scriptsize{$\pm0.23$}} & %
$60.68$\mbox{\scriptsize{$\pm0.00$}} & %
$66.97$\mbox{\scriptsize{$\pm0.51$}} & %
$78.58$\mbox{\scriptsize{$\pm0.09$}} & %
$53.75$\mbox{\scriptsize{$\pm0.60$}} & %
$-0.78$\mbox{\scriptsize{$\pm0.51$}} & %
$12.34$\mbox{\scriptsize{$\pm0.17$}} \\
\hline

\hline

\end{tabular}
}
\end{table*}

\end{document}